\documentclass[twoside,11pt]{article}

\usepackage{jmlr2e}
\usepackage[normalem]{ulem}
\usepackage{amsmath}
\usepackage{amssymb, mathtools, amsfonts}
\usepackage{tikz}
\usepackage{tikz-qtree}
\usetikzlibrary{trees}
\usepackage{algorithm}
\usepackage[noend]{algpseudocode}
\usetikzlibrary{automata,positioning}
\usepackage{multirow}
\allowdisplaybreaks
\usepackage[makeroom]{cancel}

\usepackage{caption}
\usepackage{subcaption}
\usepackage[title,toc,titletoc,page]{appendix}

\DeclareMathOperator{\pa}{pa}
\DeclareMathOperator{\ch}{ch}
\DeclareMathOperator{\de}{de}

\DeclareMathOperator{\an}{an} 

\DeclareMathOperator{\dis}{dis}
\DeclareMathOperator{\mb}{mb}
\DeclareMathOperator{\tb}{mp}
\DeclareMathOperator{\doo}{do}
\DeclareMathOperator{\primal}{\Phi}

\DeclareMathOperator{\bprimal}{\beta_{\text{primal}}}
\DeclareMathOperator{\bdual}{\beta_{\text{dual}}}

\DeclareMathOperator{\diedgeright}{\textcolor{blue}{\boldsymbol{\rightarrow}}}

\DeclareMathOperator{\biedge}{\textcolor{red}{\boldsymbol{\leftrightarrow}}}

\newcommand\ci{\perp\!\!\!\perp}

\newcommand{\red}{\textcolor{red}}

\newcommand{\blue}{\textcolor{blue}}
\newcommand{\E}{\mathbb{E}}
\newcommand{\I}{\mathbb{I}}
\newcommand{\G}{\mathcal{G}}

\usepackage{xcolor}
\usepackage[normalem]{ulem}
\newcommand{\stkout}[1]{\ifmmode\text{\sout{\ensuremath{#1}}}\else\sout{#1}\fi}

\usepackage{lastpage}
\jmlrheading{23}{2022}{1-\pageref{LastPage}}{3/20; Revised
	9/22}{10/22}{20-296}{Rohit Bhattacharya, Razieh Nabi, and Ilya Shpitser}
\ShortHeadings{Semiparametric Inference In Causal Graphical Models}{Bhattacharya, Nabi, and Shpitser}

\begin{document}

\title{
%Toward Automated Derivation of Doubly Robust and Efficient Estimators for Causal Inference
%Doubly Robust Semi-parametric Estimation Of Causal Effects In The Presence of Hidden Variables
Semiparametric Inference  For Causal Effects  In \\ Graphical Models With Hidden Variables
%Doubly Robust and Efficient Estimators for Causal Effects In Graphical Models With Hidden Variables
}

\author{\name Rohit Bhattacharya$^\dagger$ 
		\email rb17@williams.edu  \\
       	\addr Department of Computer Science \\
       	Williams College \\
       	Williamstown, MA 01267, USA
       	\AND
       	\name Razieh Nabi$^\dagger$ 
       	\email razieh.nabi@emory.edu \\
        \addr Department of Biostatistics and Bioinformatics \\
       	Emory University \\
       	Atlanta, GA 30322, USA
		\AND
		\name Ilya Shpitser
		\email ilyas@cs.jhu.edu \\
		\addr Department of Computer Science \\
		Johns Hopkins University \\
		Baltimore, MD 21218, USA
		\AND
		\name $\dagger$ Equal contribution
	}

\editor{Peter Spirtes}

\maketitle

% In order to make ulem not cause underlined emphasis
%\normalem

%%%%%%%%%%%%%%%%
% Introduction + Abstract
%%%%%%%%%%%%%%%%
\begin{abstract} 
	
	Identification theory for causal effects in causal models associated with hidden variable directed acyclic graphs (DAGs) is well studied. However, the corresponding algorithms are underused due to the complexity of estimating the identifying functionals they output.
	In this work, we bridge the gap between identification and estimation of population-level causal effects involving a single treatment and a single outcome.
	We derive influence function based estimators that exhibit double robustness for the identified effects in a large class of hidden variable DAGs where the treatment satisfies a simple graphical criterion; this class includes models yielding the adjustment and front-door functionals as special cases.
	We also provide necessary and sufficient conditions under which the statistical model of a hidden variable DAG is nonparametrically saturated and implies no equality constraints on the observed data distribution. Further, we derive an important class of hidden variable DAGs that imply observed data distributions observationally equivalent (up to equality constraints) to fully observed DAGs. In these classes of DAGs, we derive estimators that achieve the semiparametric efficiency bounds for the target of interest where the treatment satisfies our graphical criterion.
	Finally, we provide a sound and complete identification algorithm that directly yields a weight based estimation strategy for any identifiable effect in hidden variable causal models.
\end{abstract}

\begin{keywords}
	Unmeasured confounding; doubly robust estimation; nonparametric saturation; efficient influence function.  
\end{keywords}

\mbox{}
\vspace{1cm}

%%%%%%%%%%%%%%%%%%%%%%%%%%%%%%%%%%%%%%%%%%%%%%%%
\section{Introduction}
\label{sec:intro}

Causal inference is concerned with the use of observed data to reason about cause-effect relationships encoded by counterfactual parameters, such as the population-level (or average) causal effect. Since counterfactual quantities are not directly observed in the data, they must be expressed as functionals of the observed data distribution using assumptions encoded in a causal model. The ease of conveying such assumptions pictorially via a directed acyclic graph (DAG) \citep{pearl2009causality,spirtes2000causation} prompted further study of the identifiability of counterfactual quantities in causal models that factorize according to a DAG when some variables may be hidden or unobserved \citep{tian2002general}. This led to the development of a complete characterization of the identifiability of the average causal effect (ACE) of a given treatment on a given outcome in all hidden variable causal models associated with a DAG \citep{shpitser2006identification, huang2006pearl}.

Despite the sophistication of causal identification theory, estimators based on simple covariate adjustment remain the most common strategy for evaluating the ACE from data.  Estimates obtained in this way are often biased due to the presence of unmeasured confounding and/or model misspecification. A popular approach for addressing the latter issue has been to use semiparametric estimators developed using the theory of influence functions \citep{van2000asymptotic, bang2005doubly}. The most popular of these estimators is known as the \emph{augmented inverse probability weighted (AIPW)} estimator and is \emph{doubly robust} in that it gives the analyst two chances to obtain a valid estimate for the ACE -- either by specifying the correct model for the treatment assignment given observed covariates that render the treatment assignment ignorable, or by specifying the correct model for the dependence of the outcome on the treatment and these covariates. Recent work by \cite{henckel2019graphical} and \cite{rotnitzky2019efficient} yields methods for constructing statistically efficient versions of AIPW that take advantage of Markov restrictions implied on the obseved data by a fully observed causal model associated with a DAG.

If a causal model contains hidden variables, a.k.a. unmeasured confounders, causal inference becomes considerably more complicated. In the present work, we provide semiparametric estimators for the average causal effect of a single treatment variable on a single outcome variable in increasingly general scenarios, culminating in semiparametric estimators for \emph{any} hidden variable causal model of a DAG in which this effect is identifiable. The front-door model \citep{pearl1995causal} is perhaps the simplest example of a graphical model with unmeasured confounders where no valid adjustment set exists, but the effect is still identifiable. An influence function based estimator has been derived for the identified functional in the front-door model by  \cite{fulcher2020robust}. 
Weight-based estimators for a subclass of models considered in this paper, were studied in \cite{jung2020estimating}. The authors in \cite{Jung2021estimating} have a similar objective in bridging the gap between the identification and estimation theory in causal graphical models. 
Other related work includes numerical procedures for approximating the influence function proposed by \cite{frangakis2015deductive, carone2019toward}.  However, such methods are either restricted to settings where simple covariate adjustment is valid or involve numerical approximations of the function itself which may be computationally prohibitive. There also exists a rich literature on semiparametric theory with instrumental variables; see for example \cite{abadie2003semiparametric, okui2012doubly, wang2018bounded}. However, these methods impose more assumptions, such as monotonicity, exclusion restrictions, and causal relevance, than what is implied by the causal model itself represented via a DAG, possibly with hidden variables.

The paper is organized as follows. In Section~\ref{sec:prelims}, we provide a brief overview of causal graphical models. We first describe the causal and statistical models of DAGs where all variables are observed. We then move on to DAGs with hidden (or unmeasured) variables, and describe the latent projection of a hidden variable DAG into an acyclic directed mixed graph (ADMG). We discuss the district and topological factorizations of ADMGs, which are useful in deriving estimation results in Sections~\ref{sec:primal} and \ref{sec:eff_if}. Next, we introduce the general nested Markov factorization which is essential for our results in Section~\ref{sec:nested}. We close this chapter with a more detailed overview of the results and their relation to semiparametric inference theory. 

The nested Markov model of an ADMG encodes two types of equality restrictions: ordinary and generalized conditional independence constraints (a.k.a. Verma constrains). Such restrictions play an important role in deriving the tangent space of the model and the most efficient influence function based estimator for a given parameter. Hence, before diving into our estimation results for  population-level effects, we take a closer look at these constraints in Section~\ref{sec:nps}. 
We first provide a \emph{sound} and \emph{complete} procedure (Algorithm~\ref{alg:nps}) for checking whether an ADMG imposes any equality restrictions on the observed data distribution, provided the hidden variables in the corresponding hidden variable DAGs are unrestricted. In the special case where the model is nonparametric saturated, i.e., no restrictions are imposed on the tangent space of the model, the influence function corresponding to the parameter of interest is unique. Thus the corresponding estimator is the most efficient in the given class of graphical models. We then define a class of ADMGs, termed \emph{mb-shielded ADMGs}, for which the restrictions on the tangent space are all implied by simple conditional independence statements corresponding to a DAG model. Therefore, we can use  known results on deriving the tangent space of such models \citep{van2000asymptotic, rotnitzky2019efficient}, and derive estimators that achieve semiparametric efficiency bounds within this class. The results in this section are orthogonal to what the target of inference is. 

In Section~\ref{sec:primal}, we consider a class of ADMGs characterized via a simple graphical criterion for the treatment that we term \emph{primal fixability}. In this class, the average causal effect of the treatment on any choice of the outcome is always identified. We provide two alternative representations for the identified functionals that directly yield two inverse probability weighting (IPW) type estimators. These representations, called \emph{primal IPW} and \emph{dual IPW}, use variationally independent components of the natural likelihood on the observed margin of the hidden variable DAG. We further derive the nonparametric influence function -- the influence function in the nonparametric model --  
for the identified effect. The derivation is automated in the sense that in any ADMG where the treatment is primal fixable, the influence function can be mechanically derived by applying our results in Theorem~\ref{thm:eff-APIPW}. The resulting influence function based estimator can be viewed as an augmentation of the primal form. We call this \emph{augmented primal IPW (APIPW)} and show that it is doubly robust in the two sets of models involved in the primal and dual IPW estimators. We close this chapter by describing a more stringent graphical criterion leading to identification via the \emph{fixing} operation defined in \cite{richardson2017nested}.  Causal effects identification via fixing can always be reformulated as covariate adjustment, and thus leads to estimation via the semiparametric augmented IPW (AIPW) estimator.

In Section~\ref{sec:eff_if}, we focus on the class of mb-shielded ADMGs and discuss how we can exploit the constraints of such models to gain efficiency. Since mb-shielded ADMGs are equivalent to DAG models, we adapt known results for DAGs to obtain the space of all influence functions for the population-level effect of a primal fixable treatment on an outcome in this class of models. This space characterizes the \textit{regular and asymptotically linear} estimators, which are $\sqrt{n}$-consistent and asymptotically normal, for our target parameter. We further derive the most efficient estimators within this class of causal models; i.e., influence function based estimators that achieve the \emph{semiparametric efficiency bound}.

In Section~\ref{sec:nested}, we describe semiparametric estimators for general classes of functionals representing identifiable causal effects of a single treatment on a single outcome, culminating with an estimator for \emph{any} such functional. We propose the \emph{nested IPW} estimator that generalizes IPW to all hidden variable causal models where the target parameter is identified. We propose a \emph{sound} and \emph{complete} algorithm (Algorithm~\ref{alg:simplify}) that derives the corresponding nested IPW estimator when possible. One of the interesting facts about this algorithm is that when the effect is identified, it outputs a functional that only relies on the conditional densities involving the variables in the district of the treatment; see next section for a description of district and other preliminaries. 

In Section~\ref{subsec:kernel_aipw}, we discuss alternative strategies for estimating the causal effect when treatment is not primal fixable. We illustrate the key results of this paper via a series of simulation analyses in Section~\ref{sec:experiments}, followed by conclusions in Section~\ref{sec:conc}. 

%%%%%%%%%%%%%%%%%%%%%%%%%%%%%%%%%%%%%%%%%%%%%%%%

%%%%%%%%%%%%%%%%
% Preliminaries
%%%%%%%%%%%%%%%%
\section{Overview of Causal Graphical Models}
\label{sec:prelims}

The cause-effect relationship between a single treatment $T$ and an outcome $Y$ is typically established through the use of potential outcomes, a.k.a.~counterfactuals. For example, the potential outcomes $Y(1)$ and $Y(0)$ may be used to represent a hypothetical randomized controlled trial where units are randomly assigned to the treatment arm (corresponding to $T = 1$), or the control arm (corresponding to $T = 0$). The \emph{average causal effect} (ACE) is a common target that is used to compare the distribution of such counterfactual random variables on the mean difference scale. That is, $\text{ACE}\equiv \E[Y(1)] - \E[Y(0)].$ More generally, one could define a random variable $Y(t)$ corresponding to the potential outcome had treatment $T$ been assigned to some specific value $t.$ This allows for the contrast of arbitrary treatment assignments $t$ and $t'$ as $\E[Y(t)] - \E[Y(t')].$ Throughout the paper, we set our target of inference to be the mean of the counterfactual random variable $Y(t).$ That is,
\begin{align}
	\psi (t) \equiv \E[Y(t)].  \hspace{1.25cm}  {\small\textit{(target parameter)} } 
	\label{eq:target}
\end{align}

\subsection{Directed Acyclic Graphs (DAGs)}

The target parameter $\psi(t)$ cannot be expressed as a function of the observed data, or in other words, is not identified, if no assumptions are made about the data generating process \citep{pearl2009causality}. Causal graphs are a popular tool that can be used to provide an intuitive picture of substantive nonparametric assumptions made by the data analyst \citep{greenland1999causal, richardson2013single, williams2018directed, hunermund2019causal}. 

A directed acyclic graph (DAG) $\G(V)$ is defined as a set of nodes $V$ connected by directed edges such that there are no directed cycles. When the vertex set is clear from the given context, we often abbreviate $\G(V)$ as simply $\G$. 
Causal models of a DAG $\G(V)$ are defined over counterfactual random variables $V_i(\pa_i)$ for each $V_i \in V,$ where $\pa_\G(V_i)$ are the parents of $V_i$ in $\G$ and $\pa_i$ is a set of values for $\pa_{\G}(V_i).$  These counterfactuals can alternatively be viewed as being determined by a system of \emph{structural equations} $f_i(\pa_i, \epsilon_i)$ that map values $\pa_i$, as well as values of an exogenous noise term $\epsilon_i$ to values of $V_i$
\citep{pearl2009causality, malinsky2019potential,rrs2020volume_id_arxiv}. Other counterfactuals may be defined from above via recursive substitution. Specifically, for any set $A \subseteq V$, and a variable $V_i$, we have:
\[
V_i(a) \equiv V_i\big(a \cap \pa_\G(V_i), \{ V_j(a) : V_j \in \pa_\G(V_i) \setminus A \}\big),
\]
where $\{ V_j(a) : V_j \in \pa_\G(V_i) \setminus A \}$ is taken to mean the (recursively defined) set of counterfactuals associated with variables in $\pa_\G(V_i) \setminus A$, had $A$ been set to $a.$  

For any set $A \subset V$, we denote the distribution of the potential outcomes $p(\{ V_i(a) : V_i \in V \setminus A \})$ or $p(V(a))$ for short, where we assume for any $A_i \in A$, $A_i(a) = a_i$.  In other words, the potential outcome $A_i(a)$ is the constant $a_i$, the value in $a$ corresponding to $A_i$.

In a causal model of a DAG ${\G}$, $p(V(a))$ is identified by the g-formula functional \citep{robins1986new}:
\begin{align}
	p(V(a)) = \prod_{V_i \in V \setminus A} p(V_i \mid a \cap \pa_{\G}(V_i), \pa_{\G}(V_i) \setminus A). \hspace{1.25cm} {\small \textit{(g-formula)}}  
	\label{eq:g}
\end{align}
When $A$ is the empty set, we obtain the familiar DAG factorization for ${\cal G}$ meaning that the causal model of a DAG ${\cal G}$ implies the statistical model of the DAG ${\cal G}$. That is, statistical models of a DAG $\G(V)$ are sets of distributions that factorize as,
\begin{align}
	p(V) = \prod_{V_i \in V} p(V_i \mid \pa_\G(V_i)). \hspace{1.25cm}  {\small \textit{(DAG factorization)}}  
	\label{eq:dag_fact}
\end{align}

Each missing edge between pairs of variables in a DAG $\G$ imply conditional independences in $p(V).$ These can be read off directly from $\G$ via the well-known d-separation criterion \citep{pearl2009causality}. That is, for disjoint sets $X, Y, \text{ and } Z$, the following \emph{global Markov property} holds: $(X \ci_{\text{d-sep}} Y \mid Z)_\G \implies (X \ci Y \mid Z)_{p(V)}.$ When the context is clear, we simply use $X \ci Y \mid Z$ to denote conditional independence between $X$ and $Y$ given $Z.$ 

In all causal models of a DAG $\G,$ the target parameter $\psi(t)$ is identified via the back-door adjustment formula as follows,
\begin{align}
	\psi(t) = \E\big[\E[Y \mid T=t, \pa_{\G}(T)]\big]. \hspace{1.25cm}  {\small \textit{(adjustment functional)}}  
	\label{eq:bd-adj}
\end{align}%
Once the target parameter is identified, causal inference reduces to an estimation problem of the identifying functional. There exist several estimators for the adjustment functional, such as plug-in, inverse probability weighting (IPW), and augmented inverse probability weighting (AIPW) \citep{robins1994estimation, hahn1998role, robins2000marginal, van2011targeted, kennedy2017non}. An overview of these estimators can be found in Appendix~\ref{supp:DAG_inference}. 

\subsection{DAGs with Hidden Variables}

While estimation theory for fully observed causal models represented by DAGs is well developed, causal models most relevant to practical applications are sure to contain variables that are unmeasured or hidden to the data analyst. In such cases, the observed data distribution $p(V)$ can be viewed as a margin of a distribution $p(V \cup H)$ associated with a DAG $\G(V \cup H)$ where vertices in $V$ correspond to observed variables and vertices in $H$ correspond to unmeasured or hidden variables. 
Two complications arise from the presence of hidden variables. First, the target parameter $\psi(t)$ may not always be identified as a function of the observed data law, and second, parameterizations of latent variable models are generally not globally identified and may contain singularities \citep{drton2009discrete}.

A natural alternative to the latent variable model is one that places no restrictions on $p(V)$ aside from those implied by the Markov restrictions given by the factorization of $p(V \cup H)$ with respect to ${\cal G}(V \cup H)$.  It was shown by \cite{evans2018margins} that all equality constraints implied by such a factorization are captured by a nested factorization of $p(V)$ with respect to an \emph{acyclic directed mixed graph (ADMG)} ${\cal G}(V)$ derived from ${\cal G}(V \cup H)$ via the latent projection operation described by \cite{verma1990equivalence}. Such an ADMG is a smooth supermodel of infinitely many hidden variable DAGs that share the same identification theory for $\psi(t),$ and imply the same equality constraints on the margin $p(V)$ \citep{richardson2017nested, evans2019smooth}. Thus, our use of ADMGs for identification and estimation of the target $\psi(t)$ is without loss of generality.

The latent projection of a hidden variable DAG $\G(V \cup H)$ onto the observed variables $V$ is an ADMG ${\G}(V)$ with directed (\blue{$\rightarrow$}) and bidirected (\red{$\leftrightarrow$}) edges constructed as follows.  The edge $V_i \ \blue{\rightarrow} \ V_j$ exists in $\G(V)$ if there exists a directed path from $V_i$ to $V_j$ in $\G(V\cup H)$ with all intermediate vertices in $H.$ An edge $V_i \ \red{\leftrightarrow} \ V_j$ exists in $\G(V)$ if there exists a collider-free path (i.e., there are no consecutive edges of the form $\blue{\rightarrow} \circ \blue{\leftarrow}$) from $V_i$ to $V_j$ in $\G(V\cup H)$ with all intermediate vertices in $H,$ such that the first edge on the path is an incoming edge into $V_i$ and the final edge is an incoming edge into $V_j.$ An example of latent projection is provided in Appendix~\ref{app:latent_proj}. Conditional independences in the observed distribution $p(V)$ can be read off from the ADMG $\G(V)$ by a simple analogue of the d-separation criterion, known as m-separation, that generalizes the notion of a collider to include mixed edges of the form $\blue{\rightarrow} \circ \red{\leftrightarrow}, \ \red{\leftrightarrow} \circ \blue{\leftarrow},$ and $\red{\leftrightarrow} \circ \red{\leftrightarrow},$ \citep{richardson2003markov}. 

The bidirected connected components of an ADMG $\G(V)$ partition its vertices into distinct subsets known as \emph{districts}. A set $S \subseteq V$ is a district in ${\cal G}(V)$ if it forms a maximal connected component via only bidirected edges. We use $\dis_\G(V_i)$ to denote the district of $V_i$ in $\G,$ which includes $V_i$ itself, and ${\cal D}(\G)$ to denote the set of all districts in $\G$.

\subsubsection{District and Topological Factorization of ADMGs}

We first define a simple factorization of $p(V)$ relative to an ADMG in terms of its districts and objects known as kernels. A kernel $q_V(V \mid W)$ is a mapping from values of $W$ to normalized densities over $V.$ That is, $\sum_V q_V(V \mid W=w) = 1, \forall w \in W$ \citep{lauritzen96graphical}. For any set of variables $X \subseteq V,$ marginalization and conditioning in a kernel are defined in the usual way, i.e., ${q_{V\setminus X}(V\setminus X \mid W)} \equiv \sum_X q_V(V \mid W)$ and $q_V(V\setminus X \mid X, W) \equiv \frac{q_V(V \mid W)}{q_V(X \mid W)}.$ 

A distribution $p(V)$ is said to \emph{district factorize} or \emph{Tian factorize} with respect to an ADMG ${\cal G}(V)$ if
\begin{align}
	\label{eq:district_factorization}
	p(V) &= \prod_{D \in {\cal D}(\G)} q_D(D \mid \pa_\G(D)),   \hspace{1.25cm}  {\small \textit{(District ADMG factorization)}}
\end{align}%
where the parents of a set of vertices $D$ is defined as the set of parents of $D$ not already in $D,$ i.e., $\pa_{\G}(D) \equiv \left( \bigcup_{D_i \in D} \pa_{\cal G}(D_i)\right) \setminus D.$ We follow the same convention for children of a set $S,$ denoted $\ch_\G(S).$ For other standard genealogical relations defined for a single vertex $V_i,$ such as ancestors $\an_\G(V_i) \equiv \{V_j \in V \mid \exists\ V_j \blue{\rightarrow} \cdots \blue{\rightarrow} V_i \text{ in } \G \}$ and descendants $\de_\G(V_i) \equiv\{V_j \in V \mid \exists\ V_i \blue{\rightarrow} \cdots \blue{\rightarrow} V_j \text{ in } \G \}$ both of which include $V_i$ itself by convention, the extension to a set $S$ uses the disjunctive definition which also includes the set itself. For example, $\an_\G(S) = \bigcup_{S_i \in S} \an_{\cal G}(S_i).$ A list of notation and definitions used in this paper can be found in Appendix~\ref{app:glossary}.

The use of $q$ in place of $p$ in Eq.~\ref{eq:district_factorization} emphasizes the fact that these factors are not necessarily ordinary conditional distributions. Each factor $q_D(D \mid \pa_\G(D))$ can in fact be treated as a post-intervention distribution where all variables outside of $D$ are intervened on and held fixed to some constant value \citep{tian2002general}. Hence, we use $q_S(\cdot \mid \cdot)$ to denote a kernel where only variables in $S$ are random and all others are fixed. 

\cite{tian2002general} showed that each kernel $q_D(D \mid \pa_\G(D))$ appearing in Eq.~\ref{eq:district_factorization} is a function of $p(V)$ as follows. Define the \emph{Markov blanket} of a vertex $V_i$ as the district of $V_i$ and the parents of its district, excluding $V_i$ itself. That is, $\mb_\G(V_i) \equiv (\dis_\G(V_i) \cup \pa_\G(\dis_\G(V_i))) \setminus V_i$. Consider a valid topological order $\tau$ on all $k$ vertices in $V,$ that is, a sequence $(V_1, \dots, V_k)$ such that no vertex appearing later in the sequence is an ancestor of vertices earlier in the sequence. Let $\{\preceq_\tau V_i\}$ denote the set of vertices that precede $V_i$ in this sequence, including $V_i$ itself. 
Define the \emph{Markov pillow} of $V_i$, denoted by $\tb_\G(V_i)$, as its Markov blanket in a subgraph restricted to $V_i$ and its predecessors according to the topological ordering $\prec_\tau$.  We suppress the dependence of $\tb_{\G}(V_i)$ on $\prec_\tau$ for notational conciseness.
More formally, $\tb_{\G} \equiv \mb_{\G_{S}}(V_i)$ where $S = \{\preceq_\tau V_i\}$, and $\G_S$ is the subgraph of $\G$ that is restricted to vertices in $S$ and the edges between these vertices.  Then for each $D \in {\cal D}(\G)$,
\begin{align}
	\label{eq:id_tian_factor}
	q_D(D \mid \pa_\G(D)) =
	\prod_{D_i \in D} p(D_i \mid \tb_\G(D_i)). \hspace{0.75cm}  {\small \textit{(Identification of district factors)}}
\end{align}%

This leads to a factorization of the observed law as a product of simple conditional factors according to the given valid topological order,
\begin{align}
	p(V) = \prod_{V_i \in V} p(V_i \mid \tb_\G(V_i)). \hspace{1.25cm}  {\small \textit{(Topological ADMG factorization)}} \label{eq:top_factorization}
\end{align}

The above factorization (and the district factorization from which it is derived) does not always capture every equality restriction in $p(V)$ implied by the Markov property of the underlying hidden variable DAG $\G(V \cup H).$ However, it is particularly simple to work with, and under some conditions, which we derive in Section~\ref{sec:nps}, is capable of capturing all such restrictions. 

\subsubsection{Nested Markov Factorization of ADMGs}
\label{subsub:nested_factorization}

We first introduce some graphical concepts used to describe the \emph{nested Markov factorization} of an ADMG that captures all equality constraints on the observed margin $p(V).$ This factorization is defined on conditional ADMGs and kernels derived from $\G(V)$ and $p(V)$ via a fixing operation.
\emph{Conditional ADMGs} (CADMGs) $\G(V,W)$ are ADMGs whose vertices can be partitioned into random variables $V$ and fixed variables $W,$ with the restriction that only outgoing edges may be adjacent to variables in $W$ \citep{richardson2017nested}. CADMGs are often used to represent post-intervention distributions where variables in $W$ have been intervened on or \emph{fixed}. For any random variable $V_i \in V$ in a CADMG $\G(V, W)$, the usual definitions of genealogical relations and other special sets, such as parents, descendants, Markov blankets, and Markov pillows, extend naturally by allowing for the inclusion of fixed variables into these sets. In a CADMG ${\cal G}(V,W)$, districts are only defined for elements of $V$.

A vertex $V_i \in V$ is said to be \emph{fixable} in $\G(V,W)$ if $\dis_\G(V_i) \cap \de_\G(V_i) = \{V_i\}.$ In words, $V_i$ is fixable if there are no bidirected paths from $V_i$ to any of its descendants. The graphical operation of fixing $V_i,$ denoted by $\phi_{V_i}(\G),$ yields a new CADMG $\G(V \setminus V_i, W \cup V_i)$ where bidirected and directed edges into $V_i$ are removed, and $V_i$ is fixed to a particular value $v_i.$ Given a kernel $q_V(V\mid W)$ associated with the CADMG $\G(V,W),$ the corresponding probabilistic operation of fixing, denoted by $\phi_{V_i}(q_V;\G),$ yields a new kernel 

{\small
	\begin{align}
		\phi_{V_i}(q_V; \G) \equiv q_{V\setminus V_i}(V \setminus V_i \mid W \cup V_i) \equiv \frac{q_{V}(V \mid W)}{q_V(V_i \mid \mb_\G(V_i), W)}. 
		\hspace{0.5cm} \label{eq:ordinary_fixing} {\small \textit{(Probabilistic fixing operator)}}  
	\end{align}
}

The definition of fixability can be extended to a set of vertices $S$ by requiring that there exists an ordering $(S_1, \dots ,S_p)$ such that $S_1$ is fixable in $\G,$ $S_2$ is fixable in $\phi_{S_1}(\G),$ and so on. Such an ordering is said to form a valid fixing sequence for $S$. It is known that any two valid fixing sequences on $S$ yield the same CADMG, which we will denote by $\phi_S(\G(V,W)).$ Fix a CADMG $\G(V,W)$ and a corresponding kernel $q(V \mid W)$. Given a valid fixing sequence $\sigma_S$ on $S \subseteq V$ valid in $\G(V, W)$, define $\phi_{\sigma_S}(q_V; \G)$ inductively to be $q(V \mid W)$ when $S$ is empty, and $\phi_{\sigma_S\setminus S_1}(\phi_{S_1}(q_V; \G); \phi_{S_1}(\G))$ otherwise, where $\sigma_{S}\setminus S_1$ corresponds to the remainder of the sequence after removing the first element $S_1$. A concrete example demonstrating sequential applications of the graphical and probabilistic operations of fixing can be found in Appendix~\ref{supp:kernels}.

A set $D$ is called \emph{intrinsic} in $\G(V)$ if $V \setminus D$ is fixable in $\G(V)$ and $\phi_{V \setminus D}(\G(V))$ contains a single district. A distribution $p(V)$ is said to satisfy the nested Markov factorization relative to an ADMG $\G(V)$ if there exists a set of kernels $q_D(D \mid \pa_\G(D))$, one for every $D$ intrinsic in $\G(V)$, such that for every fixable set $S$ and every valid fixing sequence $\sigma_S,$
\begin{align}
	\phi_{\sigma_S}(p(V);\G) = \prod_{D \in {\cal D}(\phi_S(\G))} q_D(D \mid \pa_{\G}(D)).  \hspace{1cm} {\small \textit{(Nested Markov factorization)}}
	\label{eq:nested_Markov_model}
\end{align}%
In words, the nested Markov factorization states that every kernel that can be derived via a valid sequence of fixing satisfies the district factorization with respect to the CADMG obtained by this sequence, and each of the kernels appearing in the factorization corresponds to intrinsic sets. Given a distribution that satisfies the nested Markov factorization, for any fixable set $S$, applying any two distinct valid sequences $\sigma^1_S$, $\sigma^2_S$ to $p(V)$ and $\G(V)$ also yields the same kernel, which we can then denote as $\phi_S(p(V); \G(V))$ \citep{richardson2017nested}.

\subsection{Brief Overview of Semiparametric Inference}
\label{sec:overview_semiparam}

Assume a semiparametric model $\mathcal{M} = \{p(Z; \eta): \eta \in \Gamma\}$ where $\Gamma$ is the parameter space and $\eta$ is the parameter indexing a specific distribution. The \emph{tangent space} of a statistical model ${\cal M}$ is defined as the mean-square closure of all linear combinations of scores in corresponding parametric submodels for ${\cal M}.$ It is well-known that the tangent space in the statistical model of a DAG $\G(V)$, denoted by $\Lambda,$ can be partitioned into a direct sum of orthogonal subspaces \citep{bickel1993efficient, van2000asymptotic, tsiatis2007semiparametric}. That is, 
\begin{align}
	\Lambda \equiv \displaystyle \oplus_{V_i \in V}  \ \Lambda_i, \hspace{1.25cm}  {\small \textit{(Tangent space of statistical models of DAGs)}}
	\label{eq:DAG_tangent}
\end{align}%
where 
$\Lambda_i \equiv \big\{\alpha_i(V_i, \pa_\G(V_i)) \in \mathbb{H} \mid \E[ \alpha_i \mid \pa_\G(V_i)] = 0 \big\},$ and $\mathbb{H}$ denotes the \emph{Hilbert space} defined as the space of all mean-zero scalar functions, equipped with the inner product $\E[h_1\times h_2], \forall h_1, h_2 \in \mathbb{H}.$ If $\G$ is a complete DAG, i.e., every vertex is connected to every other vertex, then there exist no independence relations  between any sets of variables. In such scenarios, the tangent space equals the entire Hilbert space. In general, any statistical model with tangent space $\Lambda,$ where $\Lambda = \mathbb{H},$ is said to be \textit{nonparametric saturated} (NPS). 

We are often interested in a function $\psi: \eta \in \Gamma \mapsto \psi(\eta) \in \mathbb{R}$; i.e., a parameter that maps the distribution $P_\eta$ to a scalar number in $\mathbb{R}$, such as an identified average causal effect. (For brevity, we sometimes use $\psi$ instead of $\psi(\eta),$ which should be obvious from context.) %\red{Truth is denoted by $P_0$ and $\psi_0.$}

An estimator $\widehat{\psi}_n$ of a scalar parameter $\psi$ based on $n$ i.i.d copies $Z_1, \ldots, Z_n$ drawn from $p(Z; \eta),$ is \textit{asymptotically linear} if there exists a measurable random function $U_\psi(Z)$ with mean zero and finite variance such that 
\begin{align*}
	\sqrt{n} \times (\widehat{\psi}_n - \psi) = \frac{1}{\sqrt{n}} \times \sum_{i = 1}^n U_\psi(Z_i) + o_p(1), 
	%\label{eq:asym_linear} 
\end{align*}%
where $o_p(1)$ is a term that converges in probability to zero as $n$ goes to infinity. The random variable $U_\psi(Z)$ is called the \textit{influence function} (IF) of the estimator $\widehat{\psi}_n.$ The analysis is often restricted to regular and asymptotically linear (RAL) estimators to exclude super efficient estimators, such as the Hodges' estimator, whose behavior is difficult to analyze in some parts of the model space. The RAL estimator $\widehat{\psi}_n$ is  \textit{$\sqrt{n}$-consistent and asymptotically normal} (CAN), with asymptotic variance equal to the variance of its influence function $U_\psi,$ 
\begin{align*}
	\sqrt{n} \times (\widehat{\psi}_n - \psi) \ \xrightarrow[]{d} \ N\big(0, \ \text{var}(U_\psi) \big). 
\end{align*}

Influence functions in semiparametric models are derived as normalized elements of the orthogonal complement of the tangent space of the model. The orthogonal complement of the tangent space is defined as $\Lambda^{\perp} = \{h \in \mathbb{H} \mid \E[h\times h'] = 0, \forall h' \in \Lambda\};$  $\mathbb{H} = \Lambda \oplus \Lambda^\perp$, where $\oplus$ denotes the direct sum, and $\Lambda \cap \Lambda^\perp = \{0\}.$ The vector space $\Lambda^\perp$ is of particular importance because we can construct the class of all influence functions, denoted by $\cal U$, as ${\cal U} = \{U_\psi\} + \Lambda^\perp.$ In other words, upon knowing a single influence function $U_\psi$ and $\Lambda^\perp,$ we can obtain the class of all possible RAL estimators that admit the CAN property. Out of all IFs in $\cal U$, there exists a unique one which lies in the tangent space $\Lambda$ and yields the most efficient RAL estimator by recovering the \emph{semiparametric efficiency bound}. This efficient influence function can be obtained by projecting any influence function, call it $U^*_\psi$, onto the tangent space $\Lambda.$ This operation is denoted by $U^{\text{eff}}_\psi = \pi[U^*_\psi \mid \Lambda],$ where ${U^{\text{eff}}_\psi}$ denotes the efficient influence function. 
In a nonparametric saturated model (one with an unrestricted tangent space), the IF is unique and the corresponding estimator is the one that achieves the semiparametric efficiency bound. 
For a more detailed description of the concepts outlined here, see Appendix~\ref{supp:semiparam} and \citep{van2000asymptotic, tsiatis2007semiparametric}. 

%%%%%%%%%%%%%%%%%%%%%%%%%%%%%%%%%%%%%%%%%%%%%%%%

%%%%%%%%%%%%%%%%
% Tangent space and NPS
%%%%%%%%%%%%%%%%
\section{A Class of ADMGs Observationally Equivalent to DAGs}
\label{sec:nps}

Efficient semiparametric estimators must take advantage of  constraints that restrict the tangent space of the model. The nested Markov model of an ADMG encodes two types of equality constraints: ordinary conditional independences and generalized conditional independences  a.k.a Verma constraints \citep{verma1990equivalence}. For an example of the latter constraint consider the ADMG shown in Fig.~\ref{fig:nps}(a). The m-separation criterion can be used to show that the absence of an edge between $T$ and $Y$ does not correspond to any ordinary conditional independence between these variables. However, the nested Markov factorization in Eq.~\ref{eq:nested_Markov_model} implies that the kernel derived by fixing all variables except $Y$ (following any valid fixing sequence, e.g., $(L, M, T)$) district factorizes with respect to the CADMG shown in Fig.~\ref{fig:nps}(b). That is, given a distribution $p(V)$ satisfying the nested Markov factorization with respect to $\G(V)$ in Fig.~\ref{fig:nps}(a) we have,
\begin{align}
	\phi_{\{L, M, T\}}(p(V); \G)  = \sum_M p(M \mid T) \times p(Y \mid T, M, L) = q_Y(Y \mid L).
	\label{eq:verma}
\end{align}
The first equality follows from the definition of the fixing operator; the second follows from the nested Markov factorization. Eq.~\ref{eq:verma} implies that $\sum_M p(M \mid T) \times p(Y \mid T, M, L)$ is not a function of $T;$ this corresponds to a Verma constraint in $\G$. 

It is not easy to see how Verma-type restrictions can be translated into efficiency gains in estimators in general.  On the other hand, conditional independence restrictions that form Markov models associated with DAGs make efficient estimators easier to derive. This motivates our results in this section. First, in Section~\ref{subsec:nps}, we provide a sound and complete algorithm that characterizes when the nested Markov model of an ADMG $\G(V)$ is nonparametric saturated (NPS), meaning that the model imposes no equality restrictions (ordinary or generalized) on $p(V)$. In Section~\ref{subsec:mb-shielded} we describe a class of ADMGs, called \emph{mb-shielded ADMGs}, that are observationally equivalent to DAGs. In other words, when an ADMG is mb-shielded, all equality constraints in the model are implied by ordinary conditional independences according to a valid topological order. These results will allow us, in Section~\ref{sec:eff_if}, to derive efficient semiparametric estimators for certain identified counterfactual mean parameters by examining the form of the tangent space of the ADMG models in the classes we describe. However, the results in this section are general, and can be applied to any target parameter.

\subsection{Algorithm to Detect Nonparametric Saturation}
\label{subsec:nps}

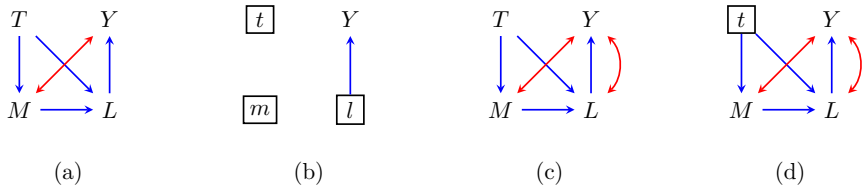
\begin{figure}[!t]
	\begin{center}
		\scalebox{0.8}{
			\begin{tikzpicture}[>=stealth, node distance=1.5cm]
				\tikzstyle{format} = [thick, circle, minimum size=1.0mm, inner sep=0pt]
				\tikzstyle{square} = [draw, thick, minimum size=4.5mm, inner sep=3pt]
				
				\begin{scope}
					\path[->, thick]
					node[] (t) {$T$}
					node[below of=t] (m) {$M$}
					node[right of=m] (l) {$L$}
					node[right of=t] (y) {$Y$}
					(t) edge[blue] (m)
					(m) edge[blue] (l)
					(l) edge[blue] (y)
					(t) edge[blue] (l)
					(m) edge[<->, red] (y)
					node[below right of=m, xshift=-0.25cm, yshift=0cm] {(a)}
					;
				\end{scope}
				
				\begin{scope}[xshift=4cm]
					\path[->, thick]
					node[square] (t) {$t$}
					node[square, below of=t] (m) {$m$}
					node[square, right of=m] (l) {$l$}
					node[right of=t] (y) {$Y$}
					(l) edge[blue] (y)
					node[below right of=m, xshift=-0.25cm, yshift=0cm] {(b)}
					;
				\end{scope}
				
				\begin{scope}[xshift=8cm]
					\path[->, thick]
					node[] (t) {$T$}
					node[below of=t] (m) {$M$}
					node[right of=m] (l) {$L$}
					node[right of=t] (y) {$Y$}
					(t) edge[blue] (m)
					(m) edge[blue] (l)
					(l) edge[blue] (y)
					(l) edge[red, <->, bend right=45] (y)
					(t) edge[blue] (l)
					(m) edge[<->, red] (y)
					node[below right of=m, xshift=-0.25cm, yshift=0cm] {(c)}
					;
				\end{scope}
				
				\begin{scope}[xshift=12cm]
					\path[->, thick]
					node[square] (t) {$t$}
					node[below of=t] (m) {$M$}
					node[right of=m] (l) {$L$}
					node[right of=t] (y) {$Y$}
					(t) edge[blue] (m)
					(m) edge[blue] (l)
					(l) edge[blue] (y)
					(l) edge[red, <->, bend right=45] (y)
					(t) edge[blue] (l)
					(m) edge[<->, red] (y)
					node[below right of=m, xshift=-0.25cm, yshift=0cm] {(d)}
					;
				\end{scope}
				
			\end{tikzpicture}
		}
	\end{center}
	\caption{ (a) Example of an ADMG where the missing edge betweem $T$ and $Y$ corresponds to a Verma constraint; (b) CADMG corresponding to the kernel $q_Y(T \mid L)$ that shows the Verma constraint in (a); (c) Example of an ADMG where the missing edge between $T$ and $Y$ does not correspond to any equality constraint; (d) The CADMG $\phi_{\neg\{Y\}}(\G)$ when checking for equality constraints in (c). }
	\label{fig:nps}
\end{figure}

The model implied by a complete DAG (a complete graph is one where all vertices are pairwise connected by a directed or bidirected edge) is nonparametric saturated, since the set of constraints from the local Markov property of the DAG model -- which states that each variable is independent of its non-descendant non-parents given its parents, and is a small list that implies all other constraints in DAG models -- is empty.  Further, any model corresponding to a DAG that is not complete is not saturated -- a missing edge in a DAG can always be translated into a statement in the local Markov property of a DAG model. 
By contrast, an ADMG that is not complete  may still represent a nonparametric saturated nested Markov model.
Consider a modification to the ADMG in Fig.~\ref{fig:nps}(a) where we add the bidirected edge $L \biedge Y;$ the resulting graph is shown in Fig.~\ref{fig:nps}(c). Though this new ADMG still lacks an edge between $T$ and $Y,$ there is no longer any Verma constraint implied by the nested Markov factorization.  In fact, it can be shown that the model is NPS.

Algorithm~\ref{alg:nps} provides a description of our procedure for checking if the nested Markov model of an ADMG $\G(V)$ is nonparametric saturated. Line~\ref{alg:nps-check} uses the notation $\phi_{\neg S}(\G)$ to denote the (unique) CADMG obtained by recursively fixing as many vertices as possible in the set $V\setminus S.$ As examples of this notation, the CADMG in Fig.~\ref{fig:nps}(b) shows $\phi_{\neg \{Y\}}(\G)$ when the original ADMG $\G$ is the one shown in Fig.~\ref{fig:nps}(a), and the CADMG in Fig.~\ref{fig:nps}(d) shows $\phi_{\neg \{Y\}}(\G)$ for the ADMG in Fig.~\ref{fig:nps}(c). 

The intuition behind the algorithm is as follows. Rather than directly examining the nested Markov model of an arbitrary ADMG $\G,$ we examine the model of its maximal arid projection \citep{shpitser2018acyclic} $\G^a.$ 
For our purposes,  $\G^a$ has the desirable property that its model is observationally equivalent to the nested Markov model of $\G$. In addition, we show (as part of the soundness proof for Algorithm~\ref{alg:nps}) that maximal arid graphs have a property similar to DAGs wrt  equality constraints: any missing edge in a maximal arid graph $\G^a$ corresponds to an (ordinary or generalized) equality constraint in the model. Algorithm~\ref{alg:nps} is designed around these facts. In particular, it returns ``NPS" when the  maximal arid projection $\G^a$ of the input ADMG $\G$ is a complete graph (one where all vertices are pairwise connected.)  Since $\G$ and $\G^a$ imply the same nested Markov model \citep{shpitser2018acyclic}, the nested Markov model of $\G$ is nonparametrically saturated by Corollary~\ref{cor:admg-complete} described in the Appendix\footnote{Briefly, any complete ADMG is mb-shielded (Theorem~\ref{thm:mb-shielded}), and hence equivalent to a complete DAG.}. The algorithm returns ``not NPS" when the maximal arid projection $\G^a$ has no edge (directed or bidirected) between at least one pair of vertices, in particular the pair $(V_i, V_j)$ for which the checks in line~\ref{alg:nps-check} succeeded. 

\begin{algorithm}[!t]
	\caption{\textproc{Check Nonparametric Saturation} $(\G)$} \label{alg:nps}
	\begin{algorithmic}[1]
		\State Let $\tau$ be a valid topological order for $V$
		\vspace{0.25em}
		\For{each distinct $(V_i, V_j) \text{ pair  in } V$}
		\State Assume wlog $V_j \prec_\tau V_i$ and let $D$ be the district of $V_i$ in $\phi_{\neg \{V_i\}}(\G)$
		\If{$V_j \not\in \pa_\G(D_i) \text{ for all } D_i \in D \text {\bf \ and }  \phi_{\neg \{V_i, V_j\}}(\G) \text{ has more than one district}$} \label{alg:nps-check}
		\State \textbf{return} not NPS \label{alg:nps-check1}
		\EndIf
		\EndFor
		\State \textbf{return} NPS \label{alg:nps-return-true}
	\end{algorithmic}
\end{algorithm}

Algorithm~\ref{alg:nps} is also computationally tractable; it runs in polynomial time with respect to the number of vertices and edges in the graph $\G.$ The complexity of the outer and inner loops is ${\cal O}(|V|^2).$ Naive implementations for computing CADMGs such as $\phi_{\neg \{V_i\}}$ and $\phi_{\neg \{V_i, V_j\}}$ also have polynomial complexity ${\cal O}(|V|^2 + |V|\times |E|)$ as it involves repeated applications of depth first search (popular algorithms for which have linear complexity ${\cal O}(|V| + |E|)$ \citep{tarjan1972depth}) in order to determine the fixability of a set of vertices.

The following theorem formalizes the soundness and completeness properties of our algorithm. That is, Algorithm~\ref{alg:nps} correctly declares the model to be NPS if it is indeed NPS; when the model is declared as not NPS, the form of an equality constraint is provided. 
\begin{theorem}[Soundness and completeness of Algorithm~\ref{alg:nps}]\ \\
	Algorithm~\ref{alg:nps} is sound and complete for determining the absence of equality constraints
	in the nested Markov model of an ADMG $\G(V)$.
	%by determining the absence of equality constraints.
	\label{thm:nps}
\end{theorem}

\subsubsection{Example: Nonparametric Saturation}

As an example, we demonstrate applying Algorithm~\ref{alg:nps} to the ADMGs in Fig.~\ref{fig:nps}(a) and (c). As all pairs of vertices besides $T$ and $Y$ are connected via a directed or bidirected edge in these ADMGs, the conditions in line~\ref{alg:nps-check} trivially evaluate to False for these pairs; we thus focus on steps executed when examining the pair $(T, Y).$ Since $T$ is an ancestor of $Y$ we have $T \prec_\tau Y.$ The algorithm first examines the CADMG $\phi_{\neg\{Y\}}(\G).$ In the case of Fig.~\ref{fig:nps}(a), this CADMG corresponds to the one shown in Fig.~\ref{fig:nps}(b), and we see $T$ is not a parent of any member of the district of $Y$ (which is just $Y$ in this case.) The CADMG $\phi_{\neg\{T, Y\}}(\G)$ is similar to the one shown in Fig.~\ref{fig:nps}(b) except $T$ remains a random vertex. In this CADMG, there are two distinct districts $\{T\}$ and $\{Y\}.$ Hence both conditions in line~\ref{alg:nps-check} are met and the algorithm returns that the model is not NPS as expected. When we apply the algorithm to the ADMG shown in Fig.~\ref{fig:nps}(c) we obtain the CADMG $\phi_{\neg\{Y\}}(\G)$ shown in Fig.~\ref{fig:nps}(d). In this case $T$ is a parent of $M$ and $L$ which are both in the district of $Y$ and so the algorithm returns that the model is NPS, once again matching the discussion at the beginning of this section.

\subsection{Mb-shielded ADMGs}
\label{subsec:mb-shielded}

In this section we describe a large class of ADMGs where all equality constraints are implied by ordinary conditional independences according to a valid topological order (resembling the local Markov property for fully observed DAG models) and derive the tangent space of such ADMGs. As mentioned earlier, deriving efficient estimators in such models is considerably easier; the derivation of the tangent space for arbitrary ADMGs is a challenging problem left for future work.

First, assume the existence of a class of ADMGs where, given a topological order $\tau$, all equality constraints implied by the ADMG $\G(V)$ can be written as ordinary conditional independence statements of the form, 
\begin{align}
	V_i \ci  \{\prec_\tau V_i\} \setminus \tb_\G(V_i) \mid \tb_\G(V_i). 
	\label{eq:mb-shielded}
\end{align} %
Such a property immediately implies that the topological factorization of the observed data distribution $p(V)$ shown in Eq.~\ref{eq:top_factorization} captures all equality constraints implied by the ADMG $\G(V).$ A sound criterion for identifying ADMGs that satisfy this property is to check that an edge between two vertices $V_i$ and $V_j$ in $\G$ is absent only if $V_i \notin \mb_\G(V_j)$ \emph{and} $V_j \not\in \mb_\G(V_i).$ We call this class of ADMGs \textit{mb-shielded ADMGs}, as pairs of vertices are always adjacent if either one is in the Markov blanket of the other. We formalize this criterion in the following theorem, and show that all equality constraints in mb-shielded ADMGs are implied by the set of ordinary conditional independence statements in Eq.~\ref{eq:mb-shielded}. 

\begin{theorem}[mb-shielded ADMGs]\ \\
	Consider a distribution $p(V)$ that district factorizes with respect to an ADMG $\G(V)$ where an edge between two vertices is absent only if $V_i \notin \mb_\G(V_j)$ \emph{and} $V_j \not\in \mb_\G(V_i).$ Then, given any valid topological order on $V,$ all equality constraints in $p(V)$ are	{implied by the set of restrictions:} $V_i \ci \{\prec V_i\} \setminus \tb_\G(V_i) \mid \tb_\G(V_i),$ {$\forall V_i \in V$}. 
	\label{thm:mb-shielded}
\end{theorem}
According to Theorem~\ref{thm:mb-shielded} and the local Markov property of DAGs, we can see that all the equality constraints in an mb-shielded ADMG are DAG-like.
One of the implications of Theorem \ref{thm:mb-shielded} is that the tangent space in an mb-shielded ADMG is identical to the one of a DAG  provided in display (\ref{eq:DAG_tangent}) by replacing each $\pa_\G(V_i)$ with $\tb_\G(V_i).$ This follows directly from Lemma 1.6 in  \cite{van2003unified}, but we reiterate these results below for the sake of completeness.
\begin{lemma}[$\Lambda$ and $\Lambda^{\perp}$ in mb-shielded ADMGs]\ \\
	Consider the statistical model $\mathcal{M}(\G)$ where $\G(V)$ is an mb-shielded ADMG. The tangent space of $\mathcal{M}(\G)$ is given by a direct sum of mutually orthogonal spaces: $\Lambda = \oplus_{V_i \in V} \Lambda_{i},$ where 
	\begin{align*}
		\Lambda_{i} 
		&= \big\{\alpha_i(V_i, \tb_\G(V_i)) \in \mathbb{H} \text{ s.t. }   \E[\alpha_i \mid \tb_\G(V_i)] = 0 \big\}\\
		&= \big\{ \alpha_i(V_i, \tb_\G(V_i)) -  \E[\alpha_i \mid \tb_\G(V_i)], \ \forall \alpha_i(V_i, \tb_\G(V_i)) \in \mathbb{H}   \big\}.
	\end{align*}%
	
	\noindent In addition, the projection of an element $h(V) \in \mathbb{H}$ onto $\Lambda_{i},$ denoted by $h_i,$ is given by {\small $h_i \equiv \Pi[h(V) \mid \Lambda_{i}] = \E\big[h(V) \mid V_i, \tb_\G(V_i)\big] - \E\big[h(V) \mid \tb_\G(V_i) \big].$} Consequently, the orthogonal complement of the tangent space $\Lambda^{\perp}$ is given as follows,
	{\small 
		\begin{align*}
			\Lambda^{\perp}  
			&= \bigg\{ \sum_{V_i \in V} \ \alpha_i(V_1, \dots, V_i) - \E\Big[\alpha_i(V_1, \dots, V_i) \ \Big| \ V_i, \tb_\G(V_i)\Big]\bigg\},
		\end{align*}
	}%
	where $\alpha_i(V_1, \dots, V_i)$ is any function of $V_1$ through $V_i$ in $\mathbb{H},$ such that $\E[\alpha_i \mid V_1, \dots, V_{i-1}] = 0.$ 
	\label{lem:Lambda} \\ 
\end{lemma} 

In the following section, we provide new IPW and influence function based estimators for our counterfactual mean target $\psi(t),$ in ADMGs that satisfy a simple graphical criterion that we term \emph{primal fixability}. In Section~\ref{sec:eff_if}, we derive estimators that achieve the semiparametric efficiency bound in mb-shielded ADMGs that satisfy primal fixability. 

%%%%%%%%%%%%%%%%%%%%%%%%%%%%%%%%%%%%%%%%%%%%%%%%

%%%%%%%%%%%%%%%%
% Doubly robust IF
%%%%%%%%%%%%%%%%
\section{Average Causal Effects: Primal Fixability of Treatment T}
\label{sec:primal}

\begin{figure}[t]
	\begin{center}
		\scalebox{0.8}{
			\begin{tikzpicture}[>=stealth, node distance=2cm]
				\tikzstyle{format} = [thick, circle, minimum size=1.0mm, inner sep=0pt]
				\tikzstyle{square} = [draw, thick, minimum size=1.0mm, inner sep=3pt]
				
				\begin{scope}
					\path[->, thick]
					node[] (a) {$T$}
					node[right of=a] (b) {$M$}
					node[right of=b] (c) {$L$}
					node[right of=c] (y) {$Y$}
					node[above right of=b, xshift=-0.5cm] (l) {$C$}
					(a) edge[blue] (b)
					(b) edge[blue] (c)
					(c) edge[blue] (y)
					(b) edge[blue, bend left] (y)
					(a) edge[<->, red, bend right=20] (c)
					(a) edge[<->, red, bend right=30] (y)
					(l) edge[blue, bend right] (a)
					(l) edge[blue, bend right] (b)
					(l) edge[blue, bend left] (c)
					(l) edge[blue, bend left] (y)
					node[below right of=b, xshift=-0.35cm] {(a)}
					;
				\end{scope}
				
				\begin{scope}[xshift=8.5cm]
					\path[->, thick]
					node[] (a) {$T$}
					node[right of=a] (b) {$M$}
					node[right of=b] (c) {$L$}
					node[right of=c] (y) {$Y$}
					node[above right of=b, xshift=-0.5cm] (l) {$C$}
					(a) edge[blue] (b)
					(b) edge[blue] (c)
					(c) edge[blue] (y)
					(a) edge[blue, bend left] (y)
					(a) edge[<->, red, bend right] (c)
					(b) edge[<->, red, bend right] (y)
					(l) edge[blue, bend right] (a)
					(l) edge[blue, bend right] (b)
					(l) edge[blue, bend left] (c)
					(l) edge[blue, bend left] (y)
					node[below right of=b, xshift=-0.35cm] {(b)}
					;
				\end{scope}
			\end{tikzpicture}
		}
	\end{center}
	\caption{ Examples of acyclic directed mixed graphs where $T$ is primal fixable. }
	\label{fig:motiv}
\end{figure}
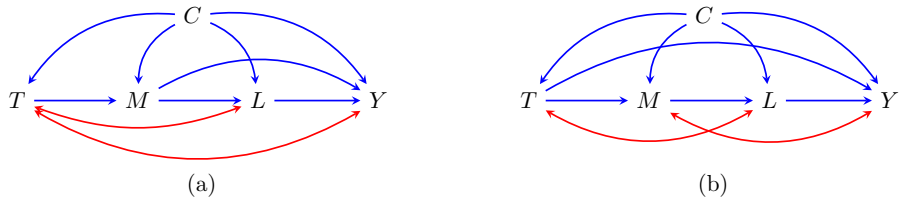

Unmeasured confounding (bidirected arrows in ADMGs) complicates identification, and hence, estimation of causal effects. Consider the ADMGs shown in Fig.~\ref{fig:motiv}. It is easy to confirm that in both ADMGs there exists no valid adjustment set\footnote{A set $Z$ is said to be a valid adjustment set wrt to $T$ and $Y$ if $p(Y(t)) = \sum_Z p(Y \mid T=t, Z) \times p(Z)$.} to identify the causal effect of $T$ on $Y.$ However, this effect is indeed identified in both graphs via more complicated functionals. The defining characteristic of these ADMGs that permits identification of the target $\psi(t)$ is that the district of $T$ does not intersect with any of its children.

In this section we consider the class of ADMGs where $\dis_\G(T) \cap \ch_\G(T) = \emptyset.$ 
This criterion encompasses many popular models in the literature, including those that satisfy the back-door and front-door criteria \citep{pearl2009causality, pearl1995causal} as special cases. We name this criterion primal fixability or \emph{p-fixability} for short, due to its generalization of the fixing criterion introduced in the definition of the nested Markov model. In what follows we discuss several identification and estimation methods for the effect of a p-fixable treatment $T$ on outcome $Y$. 

Assume $p(V)$ factorizes with respect to an ADMG $\G(V)$ where $T$ is primal fixable, and  assume, without loss of generality, that $Y$ has no descendants in $\G$.\footnote{As discussed in Section~\ref{sec:eff_if} the efficient IF based estimator is not a function of descendants of $Y$.} 
For the remainder of the paper, we assume a fixed valid topological ordering $\tau$ where the treatment $T$  appears later than all of its non-descendants i.e., $T \succ_\tau V \setminus \de_\G(T),$ and the outcome $Y$ is the final element in the topological ordering. 
This allows for easier exposition by fixing the definition of pre-treatment covariates as being any variable that appears earlier than $T$ under the ordering $\tau.$ 

We partition the set of all variables in $V$ into three disjoint sets: (i) all pre-treatment variables, (ii) all post-treatment variables that are in the same district as $T,$ and (iii) all post-treatment variables that are \emph{not} in the district of $T.$ Let $D_T$ denote the district of $T.$\footnote{The special notation for the district of $T$ as $D_T$ is due to its frequent occurrence in subsequent results.} 
Then $V$ is partitioned as follows: $V = \{ \mathbb{C}, \mathbb{L}, \mathbb{M} \}$ where
\begin{align}
	\mathbb{C} &= \{ C_i \in V \mid C_i \prec T\},  \nonumber \\
	\mathbb{L} &= \{L_i \in V \mid L_i \in D_T, L_i \succeq T\},  \nonumber \\
	\mathbb{M} &= \{ M_i \in V \mid M_i \not\in \mathbb{C} \cup \mathbb{L} \}.
	\label{eq:decomposed_V}
\end{align} 

Primal fixability is known to be a necessary and sufficient condition for the identifiability of the causal effect of $T$ on all other variables $V \setminus T$ \citep{tian2002general}. In observed data distributions $p(V)$ that district factorize according to an ADMG $\G(V)$ where $T$ is primal fixable, the resulting identifying functional for the target is 

{\small
	\begin{align}
		\label{eq:tian2002rearrange}
		\psi(t) =  \sum_{V \setminus T} \ Y \times \prod_{M_i \in  \mathbb{M}} \ p(M_i \mid \tb_\G(M_i)) \Big\vert_{T=t}  \times \sum_T \prod_{L_i \in \mathbb{L}} p(L_i \mid \tb_\G(L_i)) \times  p(\mathbb{C}),
	\end{align}
}
where $\Big\vert_{T=t}$ denotes the evaluation of $p(M_i \mid \tb_\G(M_i))$ at $T = t,$ for all $M_i \in \mathbb{M}.$

We now discuss four different estimators for the above identifying functional that rely on parametric models for a subset of pieces in the observed data distribution while allowing the other pieces to remain unrestricted. The four estimators are: (i) plug-in maximum likelihood, (ii) primal IPW, (iii) dual IPW, and (iv) augmented primal IPW; estimators (ii)--(iv) are novel contributions of the present work.
Under regularity conditions (e.g., \cite{robins1992estimating}),  all four estimators are consistent and asymptotically normal under the corresponding assumed model, discussed below. We will show that the primal and dual IPW estimators rely on distinct and variationally independent pieces of a natural parameterization for the observed data likelihood, and that augmented primal IPW is doubly robust under a semiparametric union model thereby allowing for robustness to partial model misspecification.

\subsection{Semiparametric Estimation}

The first of the four estimators is the plug-in maximum likelihood estimator (MLE) in a semiparametric model where conditional densities for variables in $\mathbb{M} \cup \mathbb{L}$ are assumed parametric forms and the law of $\mathbb{C}$ is unrestricted. 
We denote this statistical model by $\mathcal{M}_{\mathbb{M} \cup \mathbb{L}} \coloneqq \{p(V; \eta) = p(\mathbb{C}) \times \prod_{V_i \in V \setminus \mathbb{C}} p(V_i \mid \tb_{\G}(V_i); \eta_{v_i}) :  p(\mathbb{C}) \text{ is unrestricted and } \eta_{v_i} \in \Gamma_i \}$, with $\eta_{v_i}$ indexing a finite set of parameters in the parameter space $\Gamma_i$.

The parametric factors could in principle be made as flexible as allowed by sample size. By the plug-in principles, the MLE when $p(V) \in \mathcal{M}_{\mathbb{M} \cup \mathbb{L}},$ is obtained via 

{\small
	\begin{align}
		\label{eq:mle-primal}
		\widehat{\psi(t)}_{\text{mle}} =  \mathbb{P}_n \Bigg[ \sum_{V \setminus {T, \mathbb{C}}} \ Y \times \prod_{M_i \in  \mathbb{M}} \ p(M_i \mid \tb_\G(M_i); \widehat{\eta}_{m_i}) \Big\vert_{T=t}  \times \sum_T \prod_{L_i \in \mathbb{L}} p(L_i \mid \tb_\G(L_i); \widehat{\eta}_{l_i}) \Bigg] , 
	\end{align}
}%
where $\mathbb{P}_n[.] \coloneqq 1/n \sum_{j=1}^n (.)$ and $\widehat{\eta}_{v_i}$ denotes the MLE of $\eta_{v_i}.$ We can simplify the above estimator based on how $Y$ is related to $T:$ If $Y \in D_T,$ then $Y \in \mathbb{L}$ and $\sum_Y Y \times p(Y \mid \tb_\G(Y)) = \E[Y \mid \tb_\G(Y)]$ evaluated at the observed value of $T;$ if $Y \not\in D_T,$ then $Y \in \mathbb{M}$ and $\sum_Y Y \times p(Y \mid T =t, \tb_\G(Y) \setminus T) = \E[Y \mid T = t, \tb_\G(Y) \setminus T].$ The estimator in (\ref{eq:mle-primal}) is only consistent under the correct specification of the required models for variables in $\mathbb{M} \cup \mathbb{L},$ i.e., when $p(V) \in \mathcal{M}_{\mathbb{M} \cup \mathbb{L}}.$ 

We can further simplify the estimator in (\ref{eq:mle-primal}), by grouping $T$ and $\mathbb{C}$ together, and evaluating the joint distribution $p(\mathbb{C}, T)$ empirically. We refer to the corresponding statistical model by $\mathcal{M}_{\mathbb{M} \cup \mathbb{L} \setminus T}$, which is a supermodel of $\mathcal{M}_{\mathbb{M} \cup \mathbb{L}}$ where $p(T \mid \mathbb{C})$ is also left unrestricted. The MLE when $p(V) \in \mathcal{M}_{\mathbb{M} \cup \mathbb{L} \setminus T}$ is given by 

{\small
	\begin{align*}
		%		\label{eq:mle-primal-2}
		\widehat{\psi(t)}_{\text{mle,2}} =  \mathbb{P}_n \Bigg[ \sum_{V \setminus {T, \mathbb{C}}} \ Y \times \prod_{M_i \in  \mathbb{M}} \ p(M_i \mid \tb_\G(M_i); \widehat{\eta}_{m_i}) \Big\vert_{T=t}  \times  \prod_{L_i \in \mathbb{L} \setminus T} p(L_i \mid \tb_\G(L_i); \widehat{\eta}_{l_i}) \Bigg].  
	\end{align*}
}%

\subsection{Primal and Dual IPW Estimators}
\label{subsec:p-fixable_IPW}

In this section we introduce new IPW estimators whose consistency rely on correct specification of only particular subsets of models.  
Since these  estimators provide different perspectives on estimating the same target, we draw inspiration from the optimization literature \citep{dantzig1956primal, boyd2004convex} in naming them primal and dual IPW. We derived these  estimators by rethinking the identifying functional given in Eq.~\ref{eq:tian2002rearrange}. We also generalized the original definition of the fixing operator -- introduced in Section~\ref{sec:prelims} -- to a primal fixing operator, where the new kernel is now derived using the weights from primal IPW. We defer a description of the primal fixing operator to Appendix~\ref{app:primal-fixing-operator}. We will first formalize our results, and then discuss some intuition and examples.

\begin{lemma}[Primal IPW formulation]\ \\ 
	Given a distribution $p(V)$ that district factorizes with respect to an ADMG $\G(V)$ where $T$ is primal fixable, $\psi(t) = \E[\beta(t)_\text{primal}]$ where
	
	\begin{align}
		\beta(t)_{\text{primal}} \ &\equiv\  \frac{\I(T=t)}{q_{D_T}(T \mid \mb_\G(T))} \times Y %\nonumber  \\
		%&
		=\  \I(T=t) \times \frac{\displaystyle \sum_T\ \prod_{V_i \in \mathbb{L}} \ p(V_i \mid \tb_\G(V_i))}{\displaystyle \prod_{V_i \in \mathbb{L}} \ p(V_i \mid \tb_\G(V_i))} \times Y. \label{eq:primal_ipw}
	\end{align}
	\label{lem:primal_ipw}
\end{lemma}

The kernel $q_{D_T}(T \mid \mb_\G(T))$ in Lemma~\ref{lem:primal_ipw} may be viewed as a \emph{nested} propensity score derived from the post-intervention distribution $q_{D_T}(D_T \mid \pa_\G(D_T))$ where all variables outside of $D_T$ are intervened on and held fixed to some constant value. Recall that the kernel $q_{D_T}(D_T \mid \pa_\G(D_T))$ is identified as $\prod_{V_i \in D_T} p(V_i \mid \tb_\G(V_i))$ as in Eq.~\ref{eq:id_tian_factor}. Consequently, $q_{D_T}(T \mid \mb_\G(T))$ is identified by the definition of conditioning on all elements in $D_T$ outside of $T$ in the kernel $q_{D_T}(D_T \mid \pa_\G(D_T))$ as,

{\small
	\begin{align*}
		q_{D_T}(T \mid \mb_\G(T)) &=  q_{D_T}(T \mid D_T \cup \pa_{\G}(D_T) \setminus T) = \frac{q_{D_T}(D_T \mid \pa_{\G}(D_T))}{q_{D_T}(D_T \setminus T \mid \pa_\G(D_T))} \\
		&= \frac{q_{D_T}(D_T \mid \pa_\G(D_T))}{\sum_T q_{D_T}(D_T \mid \pa_\G(D_T))} = \frac{\prod_{V_i \in D_T} p(V_i \mid \tb_\G(V_i))}{\sum_T \prod_{V_i \in D_T} p(V_i \mid \tb_\G(V_i))}.\\
	\end{align*}
}%
The final expression simplifies further by noticing that all vertices appearing prior to $T$ under the topological order $\tau,$ do not contain $T$ in their Markov pillows. Consequently, $p(V_i \mid \tb_\G(V_i))$ is not a function of $T$ if $V_i \prec T.$ Thus, these terms may be pulled out of the summation in the denominator, and cancel with the corresponding term in the numerator. This reduces $V_i \in D_T$ to simply $V_i \in \mathbb{L},$ where $\mathbb{L}$ is defined in display~(\ref{eq:decomposed_V}), and yields the resulting primal IPW formulation in Eq.~\ref{eq:primal_ipw}.

We now introduce the dual formulation of $\psi(t)$ in Eq.~\ref{eq:tian2002rearrange}. Define the \emph{inverse Markov pillow} of $T$ as the set of variables outside the district of $T$ that have $T$ in their Markov pillow. Given the definition of $\mathbb{M}$ in display~(\ref{eq:decomposed_V}), the inverse Markov pillow is going to be a subset of $\mathbb{M}.$ We denote this subset by $\mathbb{M}^*$ and define it formally as follows: 
\begin{align}
	\mathbb{M}^* = \{V_i \in \mathbb{M} \mid T \in \tb_\G(V_i) \}. 
	\label{eq:inverse-markov-pillow}
\end{align}

\begin{lemma}[Dual IPW formulation]\ \\
	Given a distribution $p(V)$ that district factorizes with respect to an ADMG $\G(V)$ where $T$ is primal fixable, $\psi(t) =  \E[\beta(t)_\text{dual}]$ where
	
	\begin{align}
		\beta(t)_{\text{dual}} 
		&= \frac{\displaystyle \prod_{V_i \in \mathbb{M}^*} \ p(V_i \mid \tb_\G(V_i))  \Big \vert_{T=t}}{\displaystyle \prod_{V_i \in \mathbb{M}^*} \ p(V_i \mid \tb_\G(V_i))}\times Y. 
		\label{eq:dual-ipw} 
	\end{align}
	\label{lem:dual_ipw}
\end{lemma} 

The reason why the products in Eq.~\ref{eq:dual-ipw} are over only a subset of variables in $\mathbb{M}$ is straightforward: if there exists $V_j \in \mathbb{M} \setminus \mathbb{M}^*$ -- in other words $T \not\in \tb_\G(V_j)$ -- then $p(V_j \mid \tb_\G(V_j))\vert_{T =t}$ in the numerator cancels out with $p(V_j \mid \tb_\G(V_j))$ in the denominator. 

The representation of $\psi(t)$ as $\E[\beta(t)_{\text{primal}}]$ and $\E[\beta(t)_{\text{dual}}]$ in Lemmas~\ref{lem:primal_ipw} and \ref{lem:dual_ipw} immediately yields the corresponding primal and dual IPW estimators. Consider the following two semiparametric models: 

\begin{itemize}
	\item[(i)]   $\mathcal{M}_{\mathbb{L}}:$ where the conditional densities of $p(L_i \mid \tb_\G(L_i)),$ $\forall L_i \in \mathbb{L},$ assume parametric forms, and everything else in the observed data distribution is unrestricted.
	
	\item[(ii)] $\mathcal{M}_{\mathbb{M}}:$ where the conditional densities of $p(M_i \mid \tb_\G(M_i)),$ $\forall M_i \in \mathbb{M},$ assume parametric forms, and everything else in the observed data distribution is unrestricted. 
\end{itemize}

The primal IPW estimator $\widehat{\psi(t)}_\text{primal}$ and the dual IPW estimator $\widehat{\psi(t)}_\text{dual}$ are obtained as follows:
{\small
	\begin{align}
		\widehat{\psi(t)}_\text{primal} &= \mathbb{P}_n \Bigg[ \ 
		\I(T=t) \times \frac{ \displaystyle \sum_T\ \prod_{V_i  \in \mathbb{L}}\ p(V_i \mid \tb_\G(V_i); \widehat{\eta}_{v_i})}{\displaystyle \prod_{V_{i} \in \mathbb{L}} \ p(V_i \mid \tb_\G(V_i); \widehat{\eta}_{v_i})} \times Y \Bigg],
		\label{eq:ipw-primal} \\
		\widehat{\psi(t)}_\text{dual} &= \mathbb{P}_n \Bigg[ \ 
		\frac{\displaystyle \prod_{V_{i} \in \mathbb{M}^*} \ p(V_i \mid \tb_\G(V_i); \widehat{\eta}_{v_i}) \Big\vert_{T=t}}{\displaystyle \prod_{V_{i} \in \mathbb{M}^*} \ p(V_i \mid \tb_\G(V_i); \widehat{\eta}_{v_i})}\times Y \Bigg]. 
		\label{eq:ipw-dual}
	\end{align}
}%

\begin{theorem}[Primal and Dual IPW estimators]\ \\ 
	Under standard regularity conditions and positivity assumptions, $\widehat{\psi(t)}_\text{primal}$ and $\widehat{\psi(t)}_\text{dual}$ are consistent and asymptotically normal under models $\mathcal{M}_{\mathbb{L}}$ and  $\mathcal{M}_{\mathbb{M}},$ respectively. 
	\label{lem:ipw_consistency}
\end{theorem}

Note that when $Y \in \mathbb{L}$, we can use a weaker set of assumptions for  $\widehat{\psi(t)}_\text{primal}$  by replacing $p(Y \mid \tb_{\G}(Y); \widehat{\eta}_y) \times Y$ in the numerator with $\E[Y \mid \tb_{\G}(Y); \widehat{\eta}_y]$ and deleting the factor $p(Y \mid \tb_{\G}(Y); \widehat{\eta}_y)$ in the denominator. With this replacement, the consistency of the estimator requires correct specification of the conditional mean of $Y$ rather than its entire distribution. Likewise, when $Y \in \mathbb{M}^*$, we can use weaker assumptions for $\widehat{\psi(t)}_\text{dual}$ with a similar move; see Section~\ref{subsec:ex_primalIPW} for examples. These moves are general and can be applied anytime $Y\in \mathbb{L}$ or $Y \in \mathbb{M}^*$.

The sets of nuisance models in the primal and dual IPW estimators form variationally independent components of natural parameterizations of the observed data distribution $p(V)$, as formalized  in the following theorem. 

\begin{theorem}[Variational independence of primal IPW and dual IPW]\ \\
	Given a distribution $p(V)$ that district factorizes with respect to an ADMG $\G(V)$ where $T$ is primal fixable, the IPW estimators $\psi_{primal}$ and $\psi_{\text{dual}}$ proposed in Lemmas~\ref{lem:primal_ipw} and \ref{lem:dual_ipw} respectively, use variationally independent components of the observed distribution $p(V),$ under any parameterization of $p(V)$ that is decomposable with respect to districts in $\G(V)$.
	\label{thm:disjoint}
\end{theorem}

Primal IPW can be viewed as a generalization of the truncated factorization in DAGs to \textit{truncated kernel factorization} in ADMGs. The g-computation algorithm for a DAG model involves truncation of the DAG factorization, namely dropping a simple conditional factor of the treatment given its parents, i.e., $p(V(t)) = \{p(V)/p(T = t \mid \pa_\G(T))\} \vert_{T=t}.$ 
Similarly, the primal formulation can be viewed as truncation of the district factorization in Eq.~\ref{eq:tian2002rearrange}, where the nested conditional factor for the treatment given its Markov blanket, which is a part of one of the district terms in Eq.~\ref{eq:tian2002rearrange}, is dropped from the observed joint distribution, i.e., $p(V(t)) = \{p(V)/q_{D_T}(T \mid \mb_\G(T))\}\vert_{T=t}.$
The intuition for the dual IPW can be gained by viewing it as a probabilistic formalization of the node splitting operation in single world intervention graphs (SWIGs) described in \cite{richardson2013single}. To provide more concrete intuition on the primal and dual IPW estimators, we discuss their application to the ADMGs shown in Fig.~\ref{fig:motiv}. 

\subsubsection{Examples: primal and dual IPW estimators}
\label{subsec:ex_primalIPW}

Consider the ADMG in Fig.~\ref{fig:motiv}(a). $T$ is primal fixable as there is no bidirected path from $T$ to any of its children, namely $M.$ The inverse Markov pillow of $T$ in Fig.~\ref{fig:motiv}(a) is just $M.$ Per Lemmas~\ref{lem:primal_ipw} and \ref{lem:dual_ipw}, the primal  and dual IPW formulations for the identifiable functional of the target parameter $\psi(t)$ in Fig.~\ref{fig:motiv}(a) are given by,

{\scriptsize
	\begin{align*}
		(\text{Fig.~\ref{fig:motiv}a}) 
		\qquad \psi(t)_{\text{primal}} &=
		\E \left[\ \I(T=t) \times \frac{\sum_T\  p(T\mid C)\times p(L\mid T,M,C)\times p(Y\mid T,M,L,C)}{p(T\mid C)\times p(L\mid T,M,C)\times p(Y\mid T,M,L,C)}\times Y \right], \text{ and}
		\\
		\qquad \psi(t)_{\text{dual}} &=
		\E\left[\frac{p(M \mid T = t,C)}{p(M \mid T,C)}\times Y \right].
	\end{align*}
}

To estimate $\psi(t)$ we proceed as follows. In case of  primal IPW, we fit conditional densities {\small $p(T \mid C), p(L \mid T, M, C),$} and {\small $p(Y \mid T, M, L, C),$} either parametrically (using generalized linear models for instance), or via more flexible models (like generalized additive models or nonparametric kernel regression methods) as long as sample size allows. The target parameter is then obtained by empirically evaluating the outer expectation using the fitted models. We can also avoid modeling the conditional density of $Y$, as the outcome regression {\small $\E[Y \mid T, M, L, C]$} suffices to estimate $\psi(t),$ i.e., $\psi_{\text{primal}}$ can be expressed equivalently as

{\scriptsize
	\[
	\E \left[\ \I(T=t) \times \frac{\sum_T\  p(T\mid C)\times p(L\mid T,M,C) \times \E[Y\mid T,M,L,C] }{p(T\mid C)\times p(L\mid T,M,C)} \ \right].
	\] 
} %

\noindent A simple procedure to estimate the dual IPW involves modeling the conditional density {\small $p(M \mid T, C).$} However, a more sophisticated procedure may take advantage of modeling the density ratio directly as suggested by \cite{sugiyama2010dimensionality}.

We now turn our attention to the ADMG in Fig.~\ref{fig:motiv}(b). The inverse Markov pillow of $T$ in Fig.~\ref{fig:motiv}(b) is $\{M, Y\}.$ The corresponding primal and dual IPW formulations are given by,

{\scriptsize
	\begin{align*}
		(\text{Fig.~\ref{fig:motiv}b}) 
		\hspace{1.5cm} \psi_{\text{primal}} &=
		\E \left[ \I(T=t) \times \frac{\sum_T\  p(T\mid C)\times p(L\mid T,M,C)}{p(T\mid C)\times p(L\mid T,M,C)}\times Y \right], \text{ and}
		\\
		\qquad \psi_{\text{dual}} &=
		\E\left[ \frac{p(M \mid T = t,C)}{p(M \mid T,C)} \times \frac{p(Y \mid T = t,M,L,C)}{p(Y \mid T,M,L,C)}\times Y\right].
	\end{align*}
}%

\noindent Similar strategies as n the previous case can be used to estimate $\psi(t)$. The conditional density of $Y$ in $\psi_{\text{dual}}$ can be replaced vby the outcome regression {\scriptsize $\E[Y \mid T = t, M, L, C],$} i.e., $\psi_{\text{dual}}$ can be expressed equivalently as

{\scriptsize
	\[
	\E\Big[ \frac{p(M \mid T = t,C)}{p(M \mid T,C)} \times \E[Y \mid T = t, M, L, C] \Big].
	\]
}

\subsection{Augmented Primal IPW Estimators}
\label{sec:p-fixable_apipw}

In the previous section we have shown the existence of two estimators for the target $\psi(t)$ that use variationally independent portions of the likelihood when $T$ is p-fixable. The question naturally arises if it is possible to combine these estimators to yield a single estimator that exhibits double robustness in the sets of models used in each one. 
In the following theorem, we derive the efficient influence function for $\psi(t)$ in the nonparametric model, $\mathcal{M}_{\text{np}},$ with no restriction on the observed data distribution. 
We then prove that the estimator obtained via this influence function is doubly robust. 
This influence function can be viewed as augmenting the primal IPW with pieces from the dual IPW. For readability, we use $\prod_{L_i \prec M_i}$ as shorthand for $\prod_{L_i \in \mathbb{L} \mid L_i \prec M_i}.$ 
The sets $\mathbb{C, L, M}$ are defined in display~(\ref{eq:decomposed_V}).

\begin{theorem}[The efficient influence function of $\psi(t)$ in $\mathcal{M}_{\text{np}}$]\ \\
	Given a distribution $p(V)$ that district factorizes with respect to an ADMG $\G(V)$ where $T$ is primal fixable, the efficient influence function for the target parameter $\psi(t)$ in the nonparametric model $\mathcal{M}_{\text{np}}$ is as follows.
	
	{\small
		\begin{align}
			U_{\psi_t} = 
			&\sum_{M_i \in \mathbb{M}} \Bigg\{ \ \frac{\I(T=t)}{\prod_{L_i \prec M_i} p(L_i \mid \tb_\G(L_i))} \times \bigg(\  \sum_{ \substack{T \cup \{\succ M_i\}}} \ Y \times \ \prod_{ \substack{V_i \in \ \mathbb{L} \ \cup \\ \{\succ M_i\} }} \ p(V_i \mid \tb_\G(V_i))\Big\vert_{T=t\text{ if } V_i \in \mathbb{M}}\ \nonumber \\
			&\hspace{4cm} - \ \sum_{ \substack{T\cup \{\succeq M_i\}}} \ Y \times \ \prod_{ \substack{V_i \in \ \mathbb{L} \ \cup \\ \{\succeq M_i\}}} \ p(V_i \mid \tb_\G(V_i)) \Big\vert_{T=t\text{ if } V_i \in \mathbb{M}} \bigg)\   \Bigg\} \nonumber 
			\\
			&+ \sum_{L_i \in \mathbb{L} \setminus T} \ \Bigg\{ \ \frac{\prod_{M_i \prec L_i} p(M_i \mid \tb_\G(M_i))\Big\vert_{T=t} }{\prod_{M_i \prec L_i} p(M_i \mid \tb_\G(M_i))} \times \bigg(\  \sum_{\{\succ L_i\}} Y \times \prod_{V_i \succ L_i} p(V_i \mid \tb_\G(V_i))\Big\vert_{T=t\text{ if } V_i \in \mathbb{M}} \nonumber \\
			&\hspace{5cm} - \sum_{\{\succeq L_i\}} \ Y \times  \prod_{V_i \succeq L_i} p(V_i \mid \tb_\G(V_i))\Big\vert_{T=t\text{ if } V_i \in \mathbb{M}}  \bigg)\ \Bigg\}	\nonumber
			\\
			&+  \sum_{V \setminus \{T, \mathbb{C}\}} \ Y \times \prod_{M_i \in \mathbb{M}} p(M_i \mid \tb_\G(M_i))\Big\vert_{T=t}  \times \prod_{L_i \in \mathbb{L} \setminus T} p(L_i \mid \tb_\G(L_i)) -  \psi(t), \label{eq:IF_childless}
		\end{align}
	}%
	where the sets $\mathbb{C, L, M}$ are defined in display~(\ref{eq:decomposed_V}). The asymptotic efficiency bound is given by the variance of $U_{\psi_t}.$
	\label{thm:IF_childless} 
\end{theorem}

The efficient influence function contains three terms.  The first may be viewed as a centered version of the primal IPW estimator (\ref{eq:primal_ipw}), the second as a centered version of the dual IPW estimator (\ref{eq:dual-ipw}), and the third as a centered version of the plug-in estimator (\ref{eq:tian2002rearrange}).

The influence function $U_{\psi_t}$ in Theorem~\ref{thm:IF_childless} depends on unknown conditional densities (a.k.a. nuisance parameters). Let $\widehat{U}_{\psi_t}$ denote the influence function where the unknown nuisance parameters are replaced with their corresponding estimators. Thus, the estimating equation $\mathbb{P}_n[\widehat{U}_{\psi_t}] = 0$ yields an estimator for $\psi(t)$ that we call  \emph{augmented primal IPW} (APIPW). In the following theorem, we show that this estimator exhibits a double robustness behavior with respect to models involving variables in $\mathbb{M}$ and $\mathbb{L}$.

\begin{theorem}[Double robustness of Augmented Primal IPW]\ \\
	Under standard regularity conditions and positivity assumptions, the estimator obtained by solving the estimating equation $\mathbb{P}_n[\widehat{U}_{\psi_t}]=0,$ where $U_{\psi_t}$ is given in Theorem~\ref{thm:IF_childless}, is consistent and asymptotically normal if all models in either  $\{ p(M_i \mid \tb_\G(M_i)), \ \forall M_i \in \mathbb{M} \}$ or $\{ p(L_i \mid \tb_\G(L_i)), \ \forall L_i \in \mathbb{L} \}$ are correctly specified. Further, the estimator is locally semiparametric efficient for the functional $\psi(t)$ under the union model $\mathcal{M}_{\mathbb{M}} \cup \mathcal{M}_{\mathbb{L}},$ at the intersection submodel $\mathcal{M}_{\mathbb{M}} \cap \mathcal{M}_{\mathbb{L}}$.    
	\label{lem:dr}
\end{theorem}

According to Theorem~\ref{lem:dr}, the APIPW estimator is a \textit{doubly robust} estimator. This allows us to perform consistent inference for the target parameter $\psi(t)$ even in settings where a large part of the model likelihood is arbitrarily misspecified, provided that the appropriate conditional models for variables in either $\mathbb{M}$ or $\mathbb{L}$ are specified correctly. The double robustness of the APIPW estimator stems from the fact that its bias  has a product form which allows parametric $(\sqrt{n})$ convergence rates for $\psi(t)$ to be obtained even if flexible machine learning models with slower than parametric convergence rates (but faster than $n^{-1/4}$) are used to fit the nuisance models, that is the conditional factors involving variables in $\mathbb{M}$ and $\mathbb{L}$; see \cite{chernozhukov2018double} for more details and \cite{robins2008higher,rotnitzky2020characterization} for further discussions. 

\subsubsection{Cancellation of Terms in the IF}

Given a post treatment variable $V_i$ and its conditional density $p(V_i \mid \tb_\G(V_i))$ in the identified functional for $\psi(t)$ in Eq.~\ref{eq:tian2002rearrange}, there is a corresponding term in the influence function $U_{\psi_t}$ in Theorem~\ref{thm:IF_childless} of the form 

{\small
	\begin{align}
		f_1(\prec V_i) \times \Big( f_2(\preceq V_i) - \sum_{V_i} f_2(\preceq V_i) \times p\big(V_i \mid \tb_\G(V_i)\big) \Big),
		\label{eq:intuit_IF}
	\end{align}
}%
where $f_1(\prec V_i)$ denotes a function of variables that precedes $V_i$ in the topological order. Similarly, $f_2(\preceq V_i)$ is a function of $\{\prec V_i\}$ and $V_i$ itself. Sometimes, these terms in the influence function $U_{\psi_t}$ may cancel each other out. For instance, assume there are two consecutive variables $V_i, V_{i+1} \in \mathbb{L}$ (or $\in \mathbb{M}$) such that $\tb_\G(V_{i+1}) \setminus V_i \subseteq \tb_\G(V_i)$. The corresponding terms in the influence function share some common terms: First, the two share the same weight terms, i.e., $f_1(\prec V_{i+1}) = f_1(\prec V_{i}),$ and second $ f_2(\preceq V_{i}) =  \sum_{V_{i+1}} f_2(\preceq V_{i+1}) \times p\big(V_{i+1} \mid \tb_\G(V_{i+1})\big).$ Therefore, through simple algebra, we note that $V_i$ and $V_{i+1}$ can be viewed as contributing a single term to the influence function of the form shown in Eq.~\ref{eq:intuit_IF}, and that is

{\small
	\begin{align*}
		f_1(\prec V_{i+1}) \times \Big( f_2(\preceq V_{i+1}) - \sum_{V_i, V_{i+1}} f_2(\preceq V_{i+1}) \times p\big(V_{i+1}, V_{i} \mid \tb_\G(V_{i})\big) \Big).
	\end{align*}
}%

This cancellation occurs regardless of whether $V_i \in \mathbb{L}$ or $V_i \in \mathbb{M}.$ However it is important that both $V_i$ and $V_{i+1}$ be in the same set (since they need a common $f_1$ term to be factored out.) Such cancellations may be applied recursively to consecutive variables in $\mathbb{L}$ or $\mathbb{M}$.   

Another possible cancellation of terms may occur in the weights that correspond to ``dual weights'' in Eq.~\ref{eq:IF_childless}. The factors in the numerator and the denominator are exactly the same except for the fact that the numerator is evaluated at $T = t$. However, if there exists $M_i \in \mathbb{M}$ such that $T$ is not in its Markov pillow, i.e., $M_i \ci T \mid \tb_\G(M_i),$ then $p(M_i \mid \tb_\G(M_i))\vert_{T=t} / p(M_i\mid \tb_\G(M_i)) = 1.$ Note that such cancellations only involves variables in $\mathbb{M}.$ In fact, the set of conditional densities that stay in these weight terms correspond to the variables in 
% $\tb_\G^{-1}(T),$
the inverse Markov pillow of $T, $ previously denoted by $\mathbb{M}^*.$ Therefore, we have the following general simplification to the influence function $U_{\psi_t}$ in Theorem~\ref{thm:IF_childless},

{\small
	\begin{align*}
		\ \frac{\displaystyle \prod_{M_i \prec L_i} p(M_i \mid \tb_\G(M_i)) \Big\vert_{T=t} }{\displaystyle \prod_{M_i \prec L_i} \ p(M_i \mid \tb_\G(M_i))} 
		= 
		\frac{\displaystyle \prod_{M_i \in \  \mathbb{M}^* \ \cap \ \{\prec L_i\}} p(M_i \mid \tb_\G(M_i)) \Big\vert_{T=t} }{\displaystyle \prod_{M_i \in \ \mathbb{M}^* \ \cap \ \{\prec L_i\}} \ p(M_i \mid \tb_\G(M_i))}.
	\end{align*}
}

More intuition on the nonparametric IF is provided in Appendix~\ref{supp:intuition_primal_IF}. An implication of the two aforementioned forms of cancellation is that the robustness statement in Theorem~\ref{lem:dr} is somewhat conservative. In other words, it may not be necessary to model all the conditional terms mentioned in the doubly robust statement of Theorem~\ref{lem:dr}. It is sometimes possible to prune vertices from the ADMG and still achieve a doubly robust estimator that requires fitting less models as demonstrated via an example in the next subsection.

\subsubsection{Reformulation of the IF}

An alternative representation of the influence function in Theorem~\ref{thm:IF_childless} can be expressed solely in terms of the primal and dual IPW statements in Lemmas~\ref{lem:primal_ipw} and \ref{lem:dual_ipw}. This formulation serves as a helpful analytic tool for deriving efficient IFs in Section~\ref{sec:eff_if}, and can have practical implications in terms of practical usage. 

\begin{lemma}[Reformulation of the IF for augmented primal IPW]\ \\
	Under the same conditions stated in Theorem~\ref{thm:IF_childless}, the efficient influence function for the target parameter $\psi(t)$ in the nonparametric model $\mathcal{M}_{\text{np}}$ can be re-expressed as follows. 
	\begin{align*}
		U_{\psi_t} 
		= & \sum_{M_i \in \mathbb{M}} \E[\bprimal \mid \{\preceq M_i\}] - \E[\bprimal \mid \{\prec M_i\}] \ \\
		& + \sum_{L_i \in \mathbb{L}} \E[\bdual \mid \{\preceq L_i\}] - \E[\bdual \mid \{\prec L_i\}] \\
		&+ \E[\beta_{\text{primal/dual}} \mid \mathbb{C}] - \psi(t),
	\end{align*}
	where $\bprimal$ and $\bdual$ are defined via Lemmas~\ref{lem:primal_ipw} and \ref{lem:dual_ipw} respectively, and $\beta_{\text{primal/dual}}$ means that we may use either $\bprimal$ or $\bdual.$ 
	\label{thm:reform_IF}
\end{lemma}

According to the above lemma, the portion of the IF that relates to elements in $\mathbb{C},$ may be recovered using either $\bprimal$ or $\bdual.$ That is, {\small$\E[\bprimal \mid \mathbb{C}] = \E[\bdual \mid \mathbb{C}] $} and {\small $\E[\bprimal] = \E[\bdual] = \psi(t).$} Reformulation of the IF offers the advantage of restricting the modeling of conditional densities to only those involved in $\bprimal$ and $\bdual.$ The analyst may then rely on flexible regression methods in order to model each $\E[ \cdot \mid \cdot]$ above. The downside of such a formulation is that we may no longer be able to take advantage of the double robustness properties of the APIPW estimators, as stated in Theorem~\ref{lem:dr}. However, such a reformulation is quite useful for deriving the form of efficient IFs by performing projections onto the tangent space of the model; we provide further comments on estimation in Section~\ref{sec:eff_if} after deriving efficient IFs in mb-shielded ADMGs. From a practical stand-point, it is easy to translate back and forth between the reformulation and the original IF using Theorem~\ref{thm:IF_childless} and Lemma~\ref{thm:reform_IF} depending on the needs of the data analyst.

\subsubsection{Examples: Augmented Primal IPW}
\label{subsec:ex_APlIPW}

We now revisit the ADMGs in Fig.~\ref{fig:motiv} and derive the corresponding efficient influence functions in the nonparametric model $\mathcal{M}_{\text{np}}.$ Consider the ADMG in Fig.~\ref{fig:motiv}(a). The sets in display~(\ref{eq:decomposed_V}) are as follows, {\small $\mathbb{C} = \{C\}, \mathbb{L}=\{T, L, Y\},$} and {\small $\mathbb{M} = \{M\}.$} Applying Theorem~\ref{thm:IF_childless} to this graph, yields the influence function:

{\scriptsize
	\begin{align*}
		(\text{Fig.~\ref{fig:motiv}a}) 
		\hspace{1cm} U_{\psi_t} =\  &\frac{\I(T=t)}{p(T \mid C)} \times \bigg( \sum_{T,L} \ p(T \mid C) \times p(L \mid T, M, C) \times \E[Y \mid T,M,L,C]\   \\
		&\hspace{2cm} -  \sum_{T, L, M} p(M \mid T=t, C) \times p(T \mid C) \times p(L \mid T, M, C) \times \E[Y \mid T,M,L,C] \bigg)\  \\
		&+ \frac{p(M \mid T=t, C)}{p(M \mid T, C)}\times \bigg(Y - \E[Y \mid T, M, L, C] \bigg) \  \\
		&+ \frac{p(M \mid T=t, C)}{p(M \mid T, C)}\times \bigg(\E[Y \mid T, M, L, C]\ - \sum_L \ p(L \mid T, M, C) \times \E[Y \mid T, M, L, C] \bigg)\   \\
		&+\sum_{M, L} \ p(M \mid T=t, C)\times p(L \mid T, M, C) \times \E[Y \mid T, M, L, C] - \psi(t).
	\end{align*}
}

Note that in the above influence function, the term {\small $\frac{p(M \mid T=t, C)}{p(M \mid T, C)}\times \E[Y \mid T, M, L, C]$} appears twice with opposite signs. This is an example of the kind of cancellation mentioned in the previous section, where $Y$ and $L$ are consecutive elements in the set $\mathbb{L}$ that share essentially the same Markov pillow, i.e., $\tb_\G(Y)\setminus L = \tb_\G(L).$ In fact, this observation allows us to simplify the influence function even further by deriving the influence function in the ADMG $\G(V \setminus L)$ where $L$ is treated as latent; projecting out $L$ in this example, corresponds to removing all the edges into and out of $L.$ This ADMG is simply the front-door graph with baseline confounding. Given Theorem~\ref{thm:IF_childless}, the IF is as follows. 

{\scriptsize
	\begin{align*}
		(\text{Fig.~\ref{fig:motiv}a}) 
		\hspace{0.4cm} U_{\psi_t} =\  &\frac{\I(T=t)}{p(T \mid C)} \times \bigg( \sum_{T} \ p(T \mid C) \times \E[Y \mid T,M,C]\  - \sum_{T, M} p(M \mid T=t, C) \times p(T \mid C)  \times \E[Y \mid T,M,C] \bigg)\  \\
		&+ \frac{p(M \mid T=t, C)}{p(M \mid T, C)}\times \bigg(Y - \E[Y \mid T, M,C] \bigg) \ 
		+ \sum_{M}p(M \mid T=t, C) \times \E[Y \mid T, M, C] - \psi(t). 
		%	\label{eq:apipw_fig3a}
	\end{align*}
}

Now consider the ADMG in Fig.~\ref{fig:motiv}(b) where no simplification of the IF is possible. The sets in display~(\ref{eq:decomposed_V}) are {\small $\mathbb{C} = \{C\}, \mathbb{L} = \{T, L\},$} and {\small $\mathbb{M} = \{M, Y\}.$} The influence function per Theorem~\ref{thm:IF_childless} is,

{\scriptsize
	\begin{align}
		(\text{Fig.~\ref{fig:motiv}b}) 
		\hspace{1cm} U_{\psi_t} =\  &\frac{\I(T=t)}{p(T \mid C) \times p(L \mid T, M, C)} \times \bigg( \sum_T \ p(T \mid C) \times p(L \mid T, M, C) \times Y \  \nonumber \\
		&\hspace{1cm} - \sum_T \ p(T \mid C) \times p(L \mid T, M, C) \times \E[Y \mid T=t, M, L, C]\bigg)\  \nonumber \\
		&+ \frac{\I(T=t)}{p(T \mid C)} \times \bigg(\sum_{T,L} \ p(T \mid C) \times p(L \mid T, M, C) \times \E[Y \mid T=t, M, L, C]\  \nonumber  \\
		&\hspace{1cm} - \sum_{T,M,L} \ p(T \mid C) \times p(M \mid T=t, C) \times p(L \mid T, M, C) \times \E[Y \mid T=t, M, L, C] \bigg)\ \nonumber  \\
		&+ \frac{p(M \mid T=t, C)}{p(M \mid T, C)} \times \bigg( \E[Y\mid T=t, M, L, C]\  
		- \sum_{L} \ p(L \mid T, M, C) \times \E[Y\mid T=t, M, L, C] \bigg)\  \nonumber \\
		&+ \sum_{M, L} \ p(M \mid T=t, C)\times p(L \mid T, M, C) \times \E[Y \mid T=t, M, L, C] - \psi(t). 
		\label{eq:apipw_fig3b}
	\end{align}
}

We briefly describe estimation strategies for estimators resulting from the Theorem~\ref{thm:IF_childless} using the influence function in Eq.~\ref{eq:apipw_fig3b} as an example. An estimator for the target $\psi(t)$ is obtained by solving the estimating equation $\mathbb{P}_n[U_{\psi_t}] = 0.$ In the resulting estimator, conditional densities for {\small $p(T \mid C), p(M \mid T, C), p(L \mid T, M, C)$} and the outcome regression {\small $\E[Y \mid T, M, L, C]$} can be fit either parametrically or using flexible machine models (as long as the rates of convergence are fast enough to achieve $\sqrt{n}$-consistency). 
The outer expectation is then evaluated empirically using the fitted models in order to yield the target parameter. Per Theorem~\ref{lem:dr}, the estimator for $\psi(t)$ is consistent as long as one of the sets {\small $\{p(T \mid C), p(L \mid T, M, C)\}$ or $\{p(M \mid T, C), \E[Y \mid T, M, L, C]\}$} is correctly specified while allowing for arbitrary misspecification of the other. 

Another estimation strategy that is computationally simpler stems from the usage of Theorem~\ref{thm:reform_IF} to the ADMG in Fig.~\ref{fig:motiv}(b). With the simplification that {\small $\E[\bprimal \mid Y, T, M, L, C] = \bprimal,$} the target can be written as follows. 

{\small 
	\begin{align}
		(\text{Fig.~\ref{fig:motiv}b}) 
		\hspace{0.4cm} {\psi(t)}_{\text{reform}} =\ \E \big[
		& \bprimal - \E[\bprimal \mid T, M, L, C]\  \nonumber  \\
		&+ \E[\bprimal \mid M, T, C] - \E[\bprimal\mid T, C]\ \nonumber \\
		&+ \E[\bdual \mid L, T, M, C] - \E[\bdual \mid T, M, C]\  \nonumber \\
		&+ \E[\bdual \mid T, C]\  \big]. 
		\label{eq:apipw_fig3b_reform}
	\end{align}
}%
The above can be estimated from finite samples by first obtaining estimates for $\bprimal$ and $\bdual$ for each row in our data 
and then fitting flexible regressions for each $\E[\cdot \mid \cdot]$ shown in Eq.~\ref{eq:apipw_fig3b_reform} using these estimates as pseudo outcomes. The outer expectation is then evaluated empirically using these fitted models, yielding an estimate for the target parameter $\psi(t).$ 

The two estimation strategies described above come with trade-offs. The former approach requires modeling conditional densities and computing sums, but preserves the double robustness property and does not face issues of model compatibility. The latter approach trades model compatibility and double robustness for computational efficiency. 

\subsection{Special Case of Simplification When the Treatment Is Fixable} 
\label{subsec:a-fixable}

Consider the class of ADMGs where in addition to being primal fixable, the treatment $T$ has no bidirected path to any of its descendants, i.e., $\dis_\G(T) \cap \de_\G(T) = \{T\}.$ This coincides with the original criterion of fixing used in the definition of the nested Markov model in Section~\ref{sec:prelims}. We now show that when this condition holds that the identification functional in Eq.~\ref{eq:tian2002rearrange} simplifies to covariate adjustment using the Markov pillow of the treatment, and the corresponding influence function then simplifies to standard augmented IPW. Recall that we are assuming a fixed valid topological ordering $\tau$ where the treatment $T$ occurs after all its non-descendants. When the treatment is fixable, it is easy to show that this implies the Markov pillow of $T$ and Markov blanket of $T$ are the same,  $\tb_\G(T) = \mb_\G(T).$ 
The following result follows by definition of fixing, m-separation, and the backdoor adjustment criterion \citep{pearl1995causal}. 

\begin{lemma}[Identifying functional when $T$ is fixable]\ \\
	Given a distribution $p(V)$ that district factorizes with respect to an ADMG $\G(V)$ in which $T$ is fixable, $\psi(t)$ is identified as $\psi(t) = \E[\E[Y \mid T = t, \tb_\G(T)]].$
	\label{thm:a-fix_ID}
\end{lemma}

The identifiability of the target in this manner, immediately yields that the efficient influence function of Theorem~\ref{thm:IF_childless} simplifies to the one corresponding to the AIPW estimator except the conditioning set is now extended to include members of the district of $T$ and parents of this district. While the above result is not an exhaustive criterion for when such simplifications occur, it provides a simple link between the fixing operator used to define nested Markov models and the validity of covariate adjustment. 
For general criteria regarding covariate adjustment in the presence of hidden variables we refer the reader to \cite{shpitser2010validity, perkovic2015complete}.

%%%%%%%%%%%%%%%%%%%%%%%%%%%%%%%%%%%%%%%%%%%%%%%%

%%%%%%%%%%%%%%%%
% Semiparam Efficiency bound
%%%%%%%%%%%%%%%%

\section{Semiparametric Efficiency Bounds}
\label{sec:eff_if}

In Section~\ref{sec:nps}, we provided Algorithm~\ref{alg:nps} as a means of checking whether the model implied by an ADMG $\G(V)$ is nonparametrically saturated. In an NPS model with a p-fixable treatment, the augmented primal IPW estimator 
is not only doubly robust but also the most efficient estimator. On the other hand, constraints in a semiparametric model shrink the tangent space of the model. Hence, we no longer have a \emph{unique} influence function. (the class of all influence functions is $\{U_\psi + \Lambda^\perp\}$). In this section, we discuss efficiency results for the class of mb-shielded ADMGs that was proposed in Theorem~\ref{thm:mb-shielded}. The general form of the efficient IF in an arbitrary mb-shielded ADMG where $T$ is p-fixable is provided in the following theorem.  

\begin{theorem}[Efficient augmented primal IPW in mb-shielded ADMGs]\ \\
	Given a distribution $p(V)$ that district factorizes with respect to an mb-shielded ADMG $\G(V)$ where $T$ is primal fixable, the efficient influence function for the target parameter $\psi(t)$ is given as follows,
	\begin{align}
		U^{\text{eff}}_{\psi_t} 
		&= \sum_{M_i \in \mathbb{M}} \E[\bprimal \mid M_i, \tb_\G(M_i)] - \E[\bprimal \mid \tb_\G(M_i)] \  \nonumber \\
		&+ \sum_{L_i \in \mathbb{L}} \E[\bdual \mid L_i, \tb_\G(L_i)] - \E[\bdual\mid \tb_\G(L_i)] \nonumber \\
		&+ \sum_{C_i \in \mathbb{C}} \E[\beta_{\text{primal/dual}} \mid C_i, \tb_\G(C_i)] - \E[\beta_{\text{primal/dual}} \mid \tb_\G(C_i)]  
		\label{eq:p-fix_EIF}
	\end{align} %
	where $\mathbb{C, L, M}$ are defined in display~(\ref{eq:decomposed_V}), and $\bprimal$ and $\bdual$ are obtained as in Lemmas~\ref{lem:primal_ipw} and \ref{lem:dual_ipw} respectively. $\beta_{\text{primal/dual}}$ means that we can either use $\bprimal$ or $\bdual.$
	\label{thm:eff-APIPW}
\end{theorem}

The primal and dual IPWs comprise the fundamental elements of the efficient influence function in the setting where $T$ is primal fixable. Simplified symbolic representations of the efficient IF in terms of the conditional densities that appear in the topological factorization can be obtained by plugging in the expression from Theorem~\ref{thm:IF_childless} into computer algebra systems, such as \cite{tikka2017simplifying} and \cite{maxima2020maxima}. Further details on the use of these tools is outside of the scope of the paper, and we will work with the above representation for brevity.

A simple estimation procedure based on Eq.~\ref{eq:p-fix_EIF} proceeds as follows. First, fit nuisance models for $\bprimal$ and $\bdual$ and obtain estimates of these for each row in the data. Next, fit flexible regressions for each $\E[\cdot \mid \cdot]$ appearing in Eq.~\ref{eq:p-fix_EIF}  using the estimates of $\bprimal$ and $\bdual$ as pseudo outcomes. The estimating equation $\mathbb{P}_n[\widehat{U}^{\text{eff}}_{\psi_t}]=0$  yields an estimator for the target $\psi(t).$

In Section~\ref{subsec:a-fixable}, we discussed when treatment is not just primal fixable but fixable, the identifying functional in Eq.~\ref{eq:tian2002rearrange} simplified to the adjustment functional and the augmented primal IPW estimator simplified to augmented IPW. Similarly, the efficient IF in Eq.~\ref{eq:p-fix_EIF} simplies when $T$ is fixable. In the following lemma, we show that the efficient IF in mb-shielded ADMGs where $T$ is fixable is a special case of the efficient IF in Theorem~\ref{thm:eff-APIPW}, and uses only terms that involve $\beta_{\text{primal}}.$ This result can also be directly obtained from \cite{rotnitzky2019efficient} (since constraints in mb-shielded ADMGs are ordinary conditional independencies.) 
The conditional independences below rely on a slight abuse of notation where $A \ci B \mid C$ when $B \cap C \not= \emptyset$ is taken to mean $A \ci B\setminus \{B \cap C\} \mid C$.

\begin{lemma}[Efficient augmented IPW in mb-shielded ADMGs]\ \\
	Given a distribution $p(V)$ that district factorizes with respect to an mb-shielded ADMG $\G(V)$ where $T$ is fixable, the efficient influence function for the target parameter $\psi(t)$ is given as follows,
	
	{\small
		\begin{align*}
			U^{\text{eff}}_{\psi_t} = \sum_{V_i \in V^*} \E\Big[ \ \frac{\mathbb{I}(T = t)}{p(T \mid \tb_\G(T))}\times Y  \ \Big| \ V_i, \tb_\G(V_i)  \Big] - \E\Big[ \ \frac{\mathbb{I}(T = t)}{p(T \mid \tb_\G(T))}\times Y  \ \Big| \ \tb_\G(V_i) \Big],
		\end{align*}
	}
	where $V^* = V \setminus (T \cup Z \cup D)$ and 
	\label{thm:a-fix_EIF}
	{\small
		\begin{align*}
			Z &= \{Z_i \in V \mid Z_i \ci Y \mid \tb_\G(Z_i) \text{ in } \G_{V\setminus T} \text { and } Z_i \not\ci T \mid \tb_\G(Z_i)\}, \\
			D &= \{D_i \in V \mid D_i \ci T, \tb_\G(T), Y \mid \tb_\G(D_i) \}. \\
		\end{align*}
	}
\end{lemma}

Some interesting facts follow from the form of the efficient influence function shown in Lemma~\ref{thm:a-fix_EIF}. First, the efficient influence function can be obtained by simply projecting the (primal) IPW portion of the AIPW influence function. Second, the set $V \setminus V^*$ enumerates several vertices that do not affect the efficiency of estimating the target parameter $\psi(t).$ These include vertices $Z_i$ that meet the criteria for a \emph{conditional instrumental variable} (conditional on their Markov pillow) as defined in \cite{pearl2009causality, van2015efficiently}. Further, no efficiency is lost by disregarding other vertices $D_i$ that include descendants of $Y,$ and irrelevant non-descendants of $Y,$ as given by definitions of the set $D$ in Lemma~\ref{thm:a-fix_EIF}.
For further, extensive discussion of connections between efficient IFs and graphical features in fully observed DAGs see \cite{rotnitzky2019efficient}.

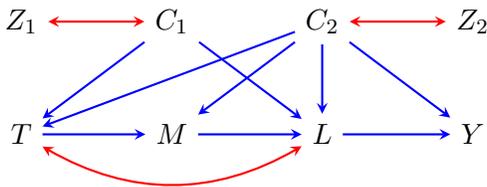
\begin{figure}[t]
	\begin{center}
		\scalebox{1}{
			\begin{tikzpicture}[>=stealth, node distance=2cm]
				\tikzstyle{format} = [thick, circle, minimum size=1.0mm, inner sep=0pt]
				\tikzstyle{square} = [draw, thick, minimum size=1.0mm, inner sep=3pt]
				
				\begin{scope}
					\path[->, thick]
					node[] (t) {$T$}
					node[right of=t] (m) {$M$}
					node[right of=m] (l) {$L$}
					node[above of=m, yshift=-0.5cm] (c1) {$C_1$}
					node[above of=l, yshift=-0.5cm] (c2) {$C_2$}
					node[right of=l] (y) {$Y$}
					node[above of=t, yshift=-0.5cm, ] (z1) {$Z_1$}
					node[above of=y, yshift=-0.5cm] (z2) {$Z_2$}
					(c1) edge[blue, bend right=0] (t)
					(c1) edge[blue] (l)
					(c2) edge[blue] (l)
					(c2) edge[blue] (t)
					(t) edge[blue] (m)
					(m) edge[blue] (l)
					(l) edge[blue] (y)
					(t) edge[red, bend right, <->] (l)
					(z1) edge[red, <->] (c1)
					(z2) edge[red, <->] (c2)
					(c2) edge[blue] (m)
					(c2) edge[blue] (y)
					;
				\end{scope}
			\end{tikzpicture}
		}
	\end{center}
	\caption{ An mb-shielded ADMG that is not NPS and where $T$ is primal fixable. }
	\label{fig:p-fix_EIF}
\end{figure}

\subsubsection*{5.1. Example: Efficient APIPW}
%\label{subsec:ex_APlPW-2}

Consider the ADMG shown in Fig.~\ref{fig:p-fix_EIF}. The conditional independencies implied by the graphs are: $C_1 \ci C_2$ and $M \ci C_1, Z_1, Z_2 \mid T, C_2.$ As this model is no longer NPS, the IF obtained via Theorem~\ref{thm:IF_childless} is not the most efficient.  However, it is easy to see that this ADMG is mb-shielded and therefore the efficient IF is given by Theorem~\ref{thm:eff-APIPW}. Fix a valid topological order $(C_1, C_2, Z_1, Z_2, T, M, L, Y).$ We have:

{\small
	\begin{align}
		(\text{Fig.~\ref{fig:p-fix_EIF}}) 
		\hspace{1cm}
		\beta_{\text{primal}} &= \I(T=t) \times \frac{\sum_T \ p(T \mid C_1, C_2)\times p(L \mid T, M,  C_1, C_2)}{p(T \mid C_1, C_2)\times p(L \mid T, M,  C_1, C_2)} \times Y, \nonumber \\
		\beta_{\text{dual}} &= \frac{p(M \mid T = t, C_2)}{p(M \mid T, C_2)} \times Y. \label{eq:bprimal_dual_fig6}
	\end{align}
}%
Define the sets $\mathbb{M} = \{M, Y\},$ $\mathbb{L}=\{T, L\},$ and $\mathbb{C} = \{C_1, C_2\}.$ Note that we have dropped terms involving the vertices $Z_1$ and $Z_2$ as it is easy to check that $\E[\beta_{\text{dual}} \mid Z_i, \tb_{\G}(Z_i)] = \E[\beta_{\text{dual}} \mid \tb_{\G}(Z_i)],$ resulting in a cancellation of these terms. By applying Theorem~\ref{thm:eff-APIPW} to Fig.~\ref{fig:p-fix_EIF}, we get the following form of the efficient APIPW. 

{\small
	\begin{align}
		(\text{Fig.~\ref{fig:p-fix_EIF}}) \hspace{1cm}
		{\psi(t)}_{\text{eff-apipw}} 
		= \E\big[     
		&\E[ \bprimal \mid Y, L, C_2 ] - \E[\bprimal \mid L, C_2] \nonumber \\
		&+ \E[ \bprimal \mid M, T, C_2 ] - \E[\bprimal \mid T, C_2] \nonumber \\
		&+ \E[\bdual \mid L, M, T, C_1, C_2] - \E[\bdual\mid M, T, C_1, C_2] \nonumber \\
		&+ \E[\bdual\mid T, C_1, C_2] - \E[\bdual \mid C_1, C_2] \nonumber \\
		&+ \E[\bdual \mid C_2]  +  \E[\bdual \mid C_1] -  \E[\bdual]    
		\big]
	\end{align}
}%
The estimation strategy for the above functional is very similar to the one used for Eq.~\ref{eq:apipw_fig3b_reform} and elaborated upon earlier in the section.

\begin{figure}[t]
	\begin{center}
		\scalebox{1}{
			\begin{tikzpicture}[>=stealth, node distance=2cm]
				\tikzstyle{format} = [thick, circle, minimum size=1.0mm, inner sep=0pt]
				\tikzstyle{square} = [draw, thick, minimum size=1.0mm, inner sep=3pt]
				
				\begin{scope}
					\path[->, thick]
					node[] (t) {$T$}
					node[right of=t] (m) {$M$}
					node[right of=m] (y) {$Y$}
					node[left of=t] (z2) {$Z_2$}
					node[left of=z2] (z1) {$Z_1$}
					node[above of=t, yshift=-0.5cm] (c1) {$C_1$}
					node[above of=m, yshift=-0.5cm] (c2) {$C_2$}
					node[above of=y, yshift=-0.5cm] (d1) {$D_1$}
					node[right of=y] (d2) {$D_2$}
					(t) edge[blue] (m)
					(z2) edge[red, <->, bend left=27] (c1)
					(z1) edge[blue] (z2)
					(z2) edge[blue] (t)
					(m) edge[blue] (y)
					(c1) edge[blue] (t)
					(c1) edge[blue, bend left=10] (m)
					(z1) edge[red, <->, bend left=30] (t)
					(c1) edge[blue, bend right=20] (z1)
					(c2) edge[blue, bend right=10] (t)
					(c2) edge[<->, red, bend left=7] (y)
					(c2) edge[blue] (m)
					(c2) edge[blue, bend right=3] (z1)
					(c2) edge[blue] (d1)
					(d1) edge[<->, red] (y)
					(y) edge[blue] (d2)
					(d1) edge[blue, bend left=10] (d2)
					(m) edge[blue, bend left=7] (d1)
					;
				\end{scope}
			\end{tikzpicture}
		}
	\end{center}
	\caption{An mb-shielded ADMG that is not NPS and where $T$ is fixable. }
	\label{fig:aipw}
\end{figure}
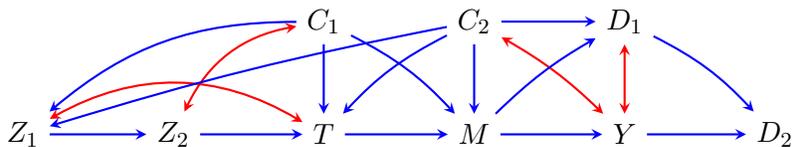

\subsubsection*{5.2 Example: Efficient AIPW}

As a concrete example of simplification of the efficient IF under fixability consider the mb-shielded ADMG in Fig.~\ref{fig:aipw}. Fix the topological order $\tau = \{C_1, C_2, Z_1, Z_2, T, M, Y, D_1, D_2\}.$ One can check that the vertices labeled $Z_1$ and $Z_2$ meet the criteria for conditional instruments and the vertices $D_1$ and $D_2$ meet the criteria of $D_i \ci T, \tb_\G(T), Y \mid \tb_\G(D_i)$ and thus do not appear in the terms of the efficient influence function given in Lemma~\ref{thm:a-fix_EIF}. When $T$ is fixable,  it is always true that $\bprimal=\{ {\I(T=t)}/{p(T \mid \tb_\G(T))} \} \times Y.$ Consequently, by applying Lemma~\ref{thm:a-fix_EIF}, we obtain the following form of the efficient AIPW for $\psi(t).$

{\small 
	\begin{align}
		(\text{Fig.~\ref{fig:aipw}}) 
		\hspace{1cm} {\psi(t)}_{\text{eff-aipw}} =\ \E & \Big[\  \E[\bprimal \mid Y, M, C_2] - \E[\bprimal \mid M, C_2] \nonumber \\
		&+ \E[\bprimal \mid M, T, C_1, C_2] - \E[\bprimal \mid T, C_1, C_2 ] \nonumber \\
		&+\E[\bprimal \mid C_2] + \E[\bprimal \mid C_1]- \E[\bprimal]\  \Big], \label{eq:aipw_eff_fig2}
	\end{align}
}%
where $\bprimal =\{ {\I(T=t)}/{p(T \mid Z_1, Z_2, C_1, C_2)}\} \times Y.$ The above functional can be estimated by following a similar strategy discussed for the functional in Eq.~\ref{eq:apipw_fig3b_reform}.

In this section, we focused on providing a representation of the efficient influence function for $\psi(t)$ when the treatment is primal fixable and the model is an mb-shielded ADMG, and  propose a naive estimation procedure based on this representation. A more detailed investigation of the local asymptotic efficiency behaviors is left for future work. 

%%%%%%%%%%%%%%%%%%%%%%%%%%%%%%%%%%%%%%%%%%%%%%%%

%%%%%%%%%%%%%%%%
% Nested IPW
%%%%%%%%%%%%%%%%
\section{Estimation of Any Identified Target Parameter}
\label{sec:nested}

Thus far we have discussed inference of the target $\psi(t)$ in a broad class of ADMGs defined by the primal fixability criterion. However, in arbitrary hidden variable causal models, $\psi(t)$ may be identified even if the treatment $T$ is not p-fixable. The resulting identifying functional is given by truncated factorization of the nested Markov model introduced in Section~\ref{subsub:nested_factorization}. This strategy for identification of the target is known to be sound and complete \citep{richardson2017nested}. That is, identification of the target parameter $\psi(t)$ in a hidden variable causal model associated with a DAG $\G(V \cup H)$ may be rephrased, without loss of generality, using its corresponding latent projection ADMG $\G(V)$.  Specifically, for $Y^* \equiv \an_{\G_{V \setminus T}}(Y)$,

{\small
	\begin{align}
		\psi(t) = \sum_{Y^*} \ Y\times \prod_{D \in {\cal D}(\G_{Y^*})} 
		\phi_{V \setminus D}(p(V); \G(V)) \bigg\vert_{T = t},
		\hspace{1cm} {\small \textit{(Truncated nested Markov factorization)}}
		\label{eq:kernel_id}
	\end{align}
}%
provided every $D \in {\cal D}(\G_{Y^*})$ is intrinsic in $\G(V)$; otherwise, $\psi(t)$ is not identifiable \citep{richardson2017nested}. Recall from the definition of the nested Markov model that a set $S \subseteq V$ is said to be intrinsic in $\G$ if $V \setminus S$ is fixable, and $\phi_{V \setminus S}(\G)$ contains a single district.

In special cases, when all observed variables are either discrete or multivariate normal, a parametric likelihood can be specified for the nested Markov model \citep{shpitser2018acyclic, evans2019smooth}, which leads naturally to estimation of $\psi(t)$ in Eq.~\ref{eq:kernel_id} by the plug-in principle. However, assuming a full parametric likelihood is often unrealistic. Here we describe estimators that use only subsets of the likelihood thus reducing the chance of model misspecification. 

\subsection{Nested IPW Estimators}
\label{subsec:nested_ipw}

In Algorithm~\ref{alg:simplify}, we describe a generalization of the primal IPW estimator, introduced in Section~\ref{sec:primal}, for \emph{any} $\psi(t)$ that is identifiable from the observed margin $p(V)$ corresponding to an ADMG $\G(V).$ 
As these estimators are derived from the nested Markov factorization of the latent projection ADMG $\G(V)$, we coin the term \textit{nested IPW} in referring to them. 

Algorithm~\ref{alg:simplify} takes as inputs the observed margin $p(V),$ the corresponding ADMG $\G(V),$ and a valid topological order $\tau;$ the algorithm then proceeds as follows. In line~\ref{alg:simplify-dstar} it identifies districts $D$ in $\G_{Y^*}$ such that $q_D(D \mid \pa_\G(D))$ does not appear in the district factorization of the original ADMG $\G(V).$ We denote this set of districts by ${\cal D}^*$ and note that this set corresponds to districts in $\G_Y^*$ that have some intersection with $D_T$ (the district of the treatment in $\G.$) We show that $\psi(t)$ is identifiable under the assumptions implied by $\G(V)$ if and only if each district in $D \in {\cal D}^*$ is intrinsic in $\G;$ the algorithm checks this criterion in line~2, and returns ``Fail'' when it is not satisfied (corresponding to a non-identified target.) When $\psi(t)$ is identified, the nested IPW functional created in line~\ref{alg:simplify-ipw-functional} of the algorithm can  be viewed as modifications to the district factorization of the ADMG $\G(V)$ involving the replacement of pieces of the kernel  $q_{D_T}(D_T \mid \pa_\G(D_T))$ with the relevant intrinsic kernels that recreate the truncated nested Markov factorization in Eq.~\ref{eq:kernel_id} (note there is no replacement intrinsic kernel involving $T$ as a random variable, hence the truncation.) Indeed, we show the equivalence of this nested IPW functional and the truncated nested Markov factorization in the proof of Theorem~\ref{thm:nested_ipw} in Appendix~\ref{app:proofs}.

Theorem~\ref{thm:nested_ipw} formalizes that the nested IPW algorithm  is sound and complete. That is, when Algorithm~\ref{alg:simplify} returns a nested IPW functional, $\psi(t)_{\text{nested}} = \psi(t)$ and when the algorithm fails to return a functional, $\psi(t)$ is not identifiable within the given model. 

\begin{algorithm}[t]
	\caption{\textproc{Nested IPW Functional} $(\G(V), p(V), \tau)$} \label{alg:simplify}
	\label{alg:simplify-toporder}
	\begin{algorithmic}[1]
		\State Let $Y^* = \an_{\G_{V\setminus T}}(Y)$ and $D_T = \dis_\G(T)$  and ${\cal D}^* \gets \{D \in {\cal D}(\G_{Y^*}) \mid D \cap D_T \not= \emptyset \}$ \label{alg:simplify-dstar}
		\If {$\exists D \in {\cal D}^*$ such that $D$ is not intrinsic in $\G$}
		\State \textbf{return} Fail
		\EndIf
		\State Define $q_D(D \mid \pa_{\G}(D)) \equiv \phi_{V \setminus D}(p(V); \G(V))$ 
		
		\State $\beta_{\text{nested}} \equiv \displaystyle \frac{\I(T=t)}{p(T \mid \tb_\G(T))} \times \prod_{D \in {\cal D}^*} \Big( \displaystyle \frac{q_D(D \mid \pa_{\G}(D))}{\prod_{D_i \in D} \ p(D_i \mid \tb_\G(D_i))}\Big) \times Y$ \label{alg:simplify-ipw-functional}
		
		\State \textbf{return} $\psi(t)_{\text{nested}} \equiv \E[\beta_{\text{nested}}]$ 
		\label{alg:simplify-return}
	\end{algorithmic}
\end{algorithm}

\begin{theorem}[Soundness and completeness of Algorithm~\ref{alg:simplify}]\ \\
	Let $p(V)$ and $\G(V)$ be the observed marginal distribution and ADMG induced by a hidden variable causal model associated with a DAG $\G(V \cup H)$.
	Then if $\psi(t)$ is identifiable in the model, $\psi(t) = \psi(t)_\text{nested}$ in the nested Markov model associated with $\G(V).$ 
	If $\psi(t)$ is not identifiable in the model, Algorithm~\ref{alg:simplify} returns `fail'. \label{thm:nested_ipw}
\end{theorem}

Note that $\psi(t)$ and $\psi(t)_\text{nested}$ are only equal in the model associated with $\G(V)$.  In other words, equality of these two functionals may depend on equality restrictions implied by the model, and the two functionals may not be equal in an unrestricted observed data model.

Nested IPW estimators are obtained by the empirical evaluation of the outer expectation in line~\ref{alg:simplify-return} and the plug-in principles. As the above algorithm suggests, such estimators rely on specifying \textit{only} a subset of the nested Markov likelihood that form the district of $T.$  If all variables in $D_T$ are discrete, this can be done using a (conditional) Moebius parameterization as described in \cite{evans2019smooth}. In general, for estimating $\psi(t)_{\text{nested}}$ we  rely on the correct specification of the kernel $q_{D_T}(D_T \mid \pa_{\G}(D_T)) \equiv \phi_{V \setminus D_T}(p(V); \G(V))$ as follows. Given the topological ADMG factorization in (\ref{eq:top_factorization}), we can write down $q_{D_T}(D_T \mid \pa_{\G}(D_T))  =  \prod_{V_i \in D_T} p(V_i \mid \tb_\G(V_i)).$ Each kernel $q_D(D\mid \pa_\G(D))$ for $D \in {\cal D}^*$ can then be obtained via conditioning, marginalization, or fixing operations on the top-level kernel $q_{D_T}(D_T \mid \pa_\G(D_T))$. Hence, in addition to some regularity constraints and positivity assumptions, the nested IPW remains consistent if all the conditional densities in the set ${p(V_i \mid \tb_\G(V_i)), \forall V_i \in D_T}$ are correctly specified. 

\subsubsection{Example: A Nested IPW Functional}
\label{subsubsec:nested-ex}

\begin{figure}[t]
	\begin{center}
		\scalebox{0.9}{
			\begin{tikzpicture}[>=stealth, node distance=2cm]
				\tikzstyle{format} = [thick, circle, minimum size=1.0mm, inner sep=0pt]
				\tikzstyle{square} = [draw, thick, minimum size=1.0mm, inner sep=3pt]
				
				\begin{scope}
					\path[->, thick]
					node[] (z) {$Z$}
					node[right of=z] (t) {$T$}
					node[right of=t] (r1) {$R_1$}
					node[above of=t, yshift=-0.25cm] (r2) {$R_2$}
					node[right of=r1] (m1) {$M$}
					node[right of=m1] (y) {$Y$}
					node[above of=m1, yshift=-0.25cm, xshift=-1cm] (c) {$C$}
					
					(z) edge[blue] (t)
					(t) edge[blue] (r1)
					(r1) edge[blue] (m1)
					(m1) edge[blue] (y)
					(z) edge[red, <->] (r2)
					(r2) edge[red, <->] (t)
					(z) edge[red, <->, bend right=25] (r1)
					(r2) edge[blue] (y)
					(c) edge[blue] (t)
					(c) edge[blue] (y)
					(t) edge[blue, bend right=23] (y)
					(c) edge[red, <->] (m1)
					(c) edge[red, <->, bend left] (y)
					;
				\end{scope}
			\end{tikzpicture}
		}
	\end{center}
	\caption{ An ADMG where the treatment is not p-fixable but $\psi(t)$ is still identified via the truncated nested Markov factorization. }
	\label{fig:general_id}
\end{figure}
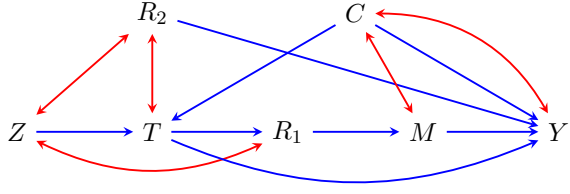

Consider the ADMG in Fig.~\ref{fig:general_id}. In this graph, $T$ is not p-fixable. However, $\psi(t)$ is still identifiable via the truncated nested Markov factorization as follows. $Y^* = \{Y, M, C, R_1, R_2\}$ and ${\cal D}(\G_{Y^*})=\{\{Y, M, C\}, \{R_1\}, \{R_2\} \}.$  Fix a valid topological order $\tau = R_2 \prec Z \prec C \prec T \prec R_1 \prec M \prec Y.$
Then from Eq.~\ref{eq:kernel_id} we have, 

{\small
	\begin{align}
		&\psi(t) = \sum_{Y^*} Y \times \phi_{V\setminus\{Y, M, C\}}(p(V);\G) \times \phi_{V\setminus R_1}(p(V);\G) \times \phi_{V\setminus R_2}(p(V);\G) \bigg\vert_{T =t} \nonumber \\
		&= \sum_{C, R_1, R_2, M} p(C) \times p(M \mid C, R_1) \times \E[Y \mid M, C, R_1, R_2, t] \times \sum_{Z} p(Z) \times p(R_1 \mid t, Z) \times p(R_2).
	\end{align}
}%
We use Algorithm~\ref{alg:simplify} to find another alternative for estimating the target $\psi(t).$ Note that ${\cal D}^*$ simply focuses on the districts related to $\G_{Y^*}$ that do not overlap with $D_T.$ Therefore, ${\cal D}^*$ in line~\ref{alg:simplify-dstar} of the algorithm is $\{\{R_1\}, \{R_2\}\}.$ Since both of these districts are intrinsic in $\G,$ Algorithm~\ref{alg:simplify} does not fail, and we get
{\small
	\begin{align}
		\psi(t)_{\text{nested}} &= \E\bigg[\ \frac{\I(T=t)}{p(T \mid R_2, Z, C)} \times \frac{\sum_{Z}p(Z) \times p(R_1 \mid T, Z)}{p(R_1 \mid T, Z, C, R_2)} \times \cancelto{1}{\frac{p(R_2)}{p(R_2)}} \times Y \ \bigg].
	\end{align}
}%
At first glance, the models for $p(R_1 \mid T, Z)$ and $p(R_1 \mid T, Z, C, R_2)$ might look incompatible. However, the topological factorization of $q_{D_T}$ provides us with a congenial representation in terms of specifying the conditional densities $p(V_i \mid \tb_\G(V_i)), \forall V_i \in D_T.$  As mentioned earlier, if variables in the district of $T$ can be assumed to be discrete, the Moebius parameterizaton \citep{evans2019smooth} may also be used to parameterize $q_{D_T}.$

\section{Alternative Strategies When the Treatment is not Primal Fixable} 
\label{subsec:kernel_aipw}

\begin{figure}[t]
	\begin{center}
		\scalebox{0.9}{
			\begin{tikzpicture}[>=stealth, node distance=2cm]
				\tikzstyle{format} = [thick, circle, minimum size=1.0mm, inner sep=0pt]
				\tikzstyle{square} = [draw, thick, minimum size=1.0mm, inner sep=3pt]
				
				\begin{scope}
					\path[->, thick]
					node[] (x) {$X$}
					node[above right of=x, xshift=1cm, yshift=-0.75cm] (z1) {$Z_1$}
					node[below right of=x, xshift=1cm, yshift=0.75cm] (z2) {$Z_2$}
					node[right of=x, xshift=2.5cm] (t) {$T$}
					node[right of=t] (y) {$Y$}
					node[above of=t, yshift=-0.1cm, xshift=-0.75cm] (c) {$C$}
					
					(x) edge[blue] (z1)
					(x) edge[blue] (z2)
					(z1) edge[blue] (t)
					(z2) edge[blue] (t)
					(t) edge[blue] (y)
					(c) edge[blue] (x)
					(c) edge[blue] (z1)
					(c) edge[blue] (z2)
					(c) edge[blue] (t)
					(c) edge[blue] (y)
					(x) edge[red, <->] (t)
					(x) edge[red, <->, bend right=35] (y)
					(z1) edge[red, <->] (z2)
					node[below right of=z2, xshift=-0.1cm, yshift=0.1cm] {(a) $\G(V)$}
					;
				\end{scope}
				
				\begin{scope}[yshift=0.0cm, xshift=9cm]
					\path[->, thick]
					node[] (x) {$X$}
					node[square, above right of=x, xshift=1cm, yshift=-0.75cm] (z1) {$z_1$}
					node[square, below right of=x, xshift=1cm, yshift=0.75cm] (z2) {$z_2$}
					node[right of=x, xshift=2.5cm] (t) {$T$}
					node[right of=t] (y) {$Y$}
					node[above of=t, yshift=-0.1cm, xshift=-0.75cm] (c) {$C$}
					
					(z1) edge[blue] (t)
					(z2) edge[blue] (t)
					(t) edge[blue] (y)
					(c) edge[blue] (x)
					(c) edge[blue] (t)
					(c) edge[blue] (y)
					(x) edge[red, <->] (t)
					(x) edge[red, <->, bend right=35] (y)
					%					(z1) edge[red, <->] (z2)
					node[below right of=z2, xshift=0.cm, yshift=0.1cm] {(b) $\phi_{Z}(\G(V))$}
					;
				\end{scope}
			\end{tikzpicture}
		}
	\end{center}
	\caption{(a) An ADMG where the treatment is not p-fixable but $\psi(t)$ is still identified via the truncated nested Markov factorization. (b) A valid sequence of fixing that yields a CADMG where $T$ is p-fixable and $p(Y(t))$ can be obtained as $p(Y(t, z_1, z_2))$ }
	\label{fig:kernel_aipw}
\end{figure}
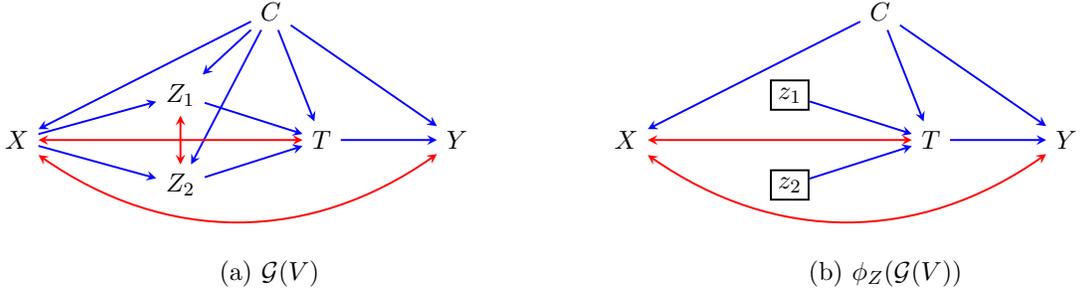

We close our theoretical discussions with an example of ADMGs where $T$ is not p-fixable (but the effect is still identifiable) and $V$ can be partitioned as follows: baseline confounders $C,$ the district of $T$ denoted by $D_T$ which includes the outcome $Y$, and instrumental variables $Z$ that affect the outcome only via the treatment. The ADMG in Fig.~\ref{fig:kernel_aipw}(a) is an example. Besides having the nested IPW to estimate the effect, we make further progress in setting up estimating equations that exhibit partial robustness to model misspecification. 

Consider the ADMG in Fig.~\ref{fig:kernel_aipw}(a) with $C = \{C\}, D_T = \{T, X, Y\},$ and $Z = \{Z_1, Z_2.\}$ The joint distribution district factorizes as follows, 
\begin{align}
	p(V) = q_C(C) \times q_Z(Z \mid \pa_\G(Z)) \times q_{D_T}(D_T \mid \pa_{\G}(D_T)), 
	\label{eq:ex-nested-factor}
\end{align}
where 
{\small
	\begin{align*}
		&q_C(C) \ =  \ p(C), \\
		&q_Z(Z \mid \pa_\G(Z)) \ = \ p(Z \mid C, X), \ \text{and} \\
		&q_{D_T}(D_T \mid \pa_{\G}(D_T)) \ = \ p(X \mid C) \times p(T, Y \mid C, X, Z). 
	\end{align*}
}%
The treatment $T$ is not primal fixable as it does not meet the condition that $\dis_\G(T ) \cap \ch_\G(T) = \emptyset.$ However, the target $\psi(t)$ is indeed identified via truncated nested Markov factorization as follows. $Y^* = \{C, Y\}$ and ${\cal D}(\G_{Y^*}) = \{\{Y\}, \{C\}\}.$
\begin{align*}
	\psi(t) 
	&= \sum_{Y,C} \ Y \times \phi_{V\setminus Y} (p(V); \G) \times \phi_{V\setminus C} (p(V); \G)\Big\vert_{T = t} \\
	&= \sum_{Y, C} \ Y \times q_Y(Y \mid T = t, C, X, Z) \times q_C(C), 
\end{align*}
where $q_C(C) = p(C)$ and
{\small $$q_Y(Y \mid T, C, X, Z) = \displaystyle \frac{\sum_{X} q_{D_T}(D_T \mid \pa_{\G}(D_T)) }{\sum_{X, Y} q_{D_T}(D_T \mid \pa_{\G}(D_T)) } = \displaystyle \frac{\sum_X p(X \mid C) \times p(T \mid C, X, Z) \times p(Y \mid T, C, X, Z)}{\sum_{X, Y} p(X \mid C) \times p(T \mid C, X, Z) \times p(Y \mid T, C, X, Z)}.$$}

Besides the nested IPW, we can estimate $\psi(t)$ by solving the estimating equation $\mathbb{P}_n[U^q_{\psi_t}] = 0,$ where 
{\small
	\begin{align*}
		U^q_{\psi_t} = &\frac{\I(T=t) \times \Big(Y - \E_{q_Y}\big[Y\mid T=t, C, X, Z\big]\Big)}{q_Z(Z \mid \pa_\G(Z)) \times q_{D_T \setminus Y}(D_T \setminus Y \mid \pa_{\G}(D_T))}  + \E_{q_Y}\big[Y\mid T=t, C, X, Z\big] - \psi(t)
	\end{align*}
}%
and {\small $\E_{q_Y}[ Y \mid T, C, X, Z] = \sum_{Y} Y \times q_{Y}(Y \mid T, C, X, Z)$ and $q_Z(Z \mid \pa_{\G}(Z)) = p(Z \mid \pa_{\G}(Z))$.} The resulting estimator resembles AIPW and exhibits a partial double robustness property; albeit using kernels that form pieces of the nested Markov likelihood. Upon correct specification of $p(X \mid C)$ and $p(T \mid C, X, Z),$ the obtained estimator is doubly robust in correct specification of either $p(Z \mid C, X)$ or $\E[Y \mid T, C, X, Z].$

The above example suggests the following question: is it always possible to find an adjustment set in a kernel obtained after a sequence of valid fixing that would yield  an estimating equation with partial robustness behaviours in the obtained estimator. In the types of ADMGs we considered in this subsection, we can always fix the instrumental variables and find a valid adjustment set in the obtained CADMG. In general, the structure preventing the validity of covariate adjustment is an inducing backdoor path, i.e., a bidirected path between $T$ and $Y$ where every non-endpoint is an ancestor of $T$ or $Y.$  It is known that there exists no separating set to block such paths \citep{verma1990equivalence}.\footnote{Verma constraints are independences that occur when such paths are broken via fixing operations.} As an example, it is easy to verify that no valid adjustment set exists for the effect of $T$ on $Y$ in n Fig~\ref{fig:kernel_aipw}(a), due to the presence of the inducing backdoor path $T \biedge X \biedge Y.$ However, upon fixing the instrumental variables $Z_1$ and $Z_2,$ the path $T \biedge X \biedge Y$ is no longer inducing and we can find an adjustment set. The resulting CADMG is shown in Fig.~\ref{fig:kernel_aipw}(b) which yields the post-intervention distribution $p(C, X, T(z), Y(z)).$ In this CADMG the effect of $T$ on $Y$ is identified by adjusting for $C$. 
%Adjustment is possible only in a kernel derived after a sequence of valid fixing on the set of ``instrumental variables." 

Finally, it is worth noting that even though $T$ is not p-fixable in Fig~\ref{fig:kernel_aipw}(a), it is p-fixable in a CADMG obtained after a valid sequence of p-fixing operation ($\{Z_1, Z_2, X\}$). This yields a general procedure for identification of the target parameter via a sequence of p-fixing operation. The details on this procedure are deferred to Appendix~\ref{app:primal-fixing-operator}, where we define the primal fixing operator that operationalizes primal IPW so that it can be recursively applied to simplify problems where the treatment is not directly p-fixable. This results in estimating equations that are sequentially reweighted and partially doubly robust in the final nuisance models used after reweighting.

%%%%%%%%%%%%%%%%%%%%%%%%%%%%%%%%%%%%%%%%%%%%%%%%

%%%%%%%%%%%%%%%%
% Experiments
%%%%%%%%%%%%%%%%
\section{Simulated Data Analysis}
\label{sec:experiments}

In this section, we  describe a set of simulations to illustrate the key results presented in this paper.
For each experiment, we generate data according to hidden variable DAGs that give rise to the latent projection ADMGs used in the motivating examples throughout the paper. Specifically, for each bidirected edge in the latent projection ADMG, we allow for the presence of unmeasured confounders that are parents of both end points of the bidirected edge; they are sampled from either a normal distribution, a uniform distribution, and/or a Bernoulli distribution. For example in Fig.~\ref{fig:motiv}(b), for the bidirected edge $T \leftrightarrow L$ the underlying hidden variable DAG contains variables $H_1$, $H_2$, and  $H_3$ which are parents of both $T$ and $L.$  We provide an example of a data generating process in Appendix~\ref{app:sims}. We use generalized additive models to fit all of the nuisance models. Our form of model misspecification involves dropping some of the appropriate conditioning variables and interaction terms in the nuisance models to demonstrate robustness to arbitrary model misspecification. The R code is available upon request.\footnote{For Python implementations, see the open source package Ananke (link: \url{https://ananke.readthedocs.io/en/latest/})}

We consider two examples where treatment is primal fixable and an example where treatment is not primal fixable but the effect is nonetheless nonparametrically identified. In primal cases, we generated data according to hidden variable DAGs $\G(V \cup H)$ that give rise to the latent projection ADMGs $\G(V)$ shown in Figures~\ref{fig:motiv}(b) and \ref{fig:p-fix_EIF}. In the case where treatment is not primal fixable, we generated data according to a hidden variable DAG  $\G(V \cup H)$ that gives rise to the latent projection ADMG $\G(V)$ shown in Figure~ \ref{fig:general_id}. We analyzed the bias, variance, and robustness behaviors of our proposed estimators (Primal IPW, Dual IPW, Augmented Primal IPW, and  Nested IPW) and compared them with the plug-in estimators. We further, evaluated the performance of the efficient influence function in the mb-shielded ADMG of Figure~\ref{fig:p-fix_EIF}.

\begin{figure}[!htbp]
	\begin{subfigure}[b]{1\textwidth}
		\centering
		\includegraphics[scale=0.33]{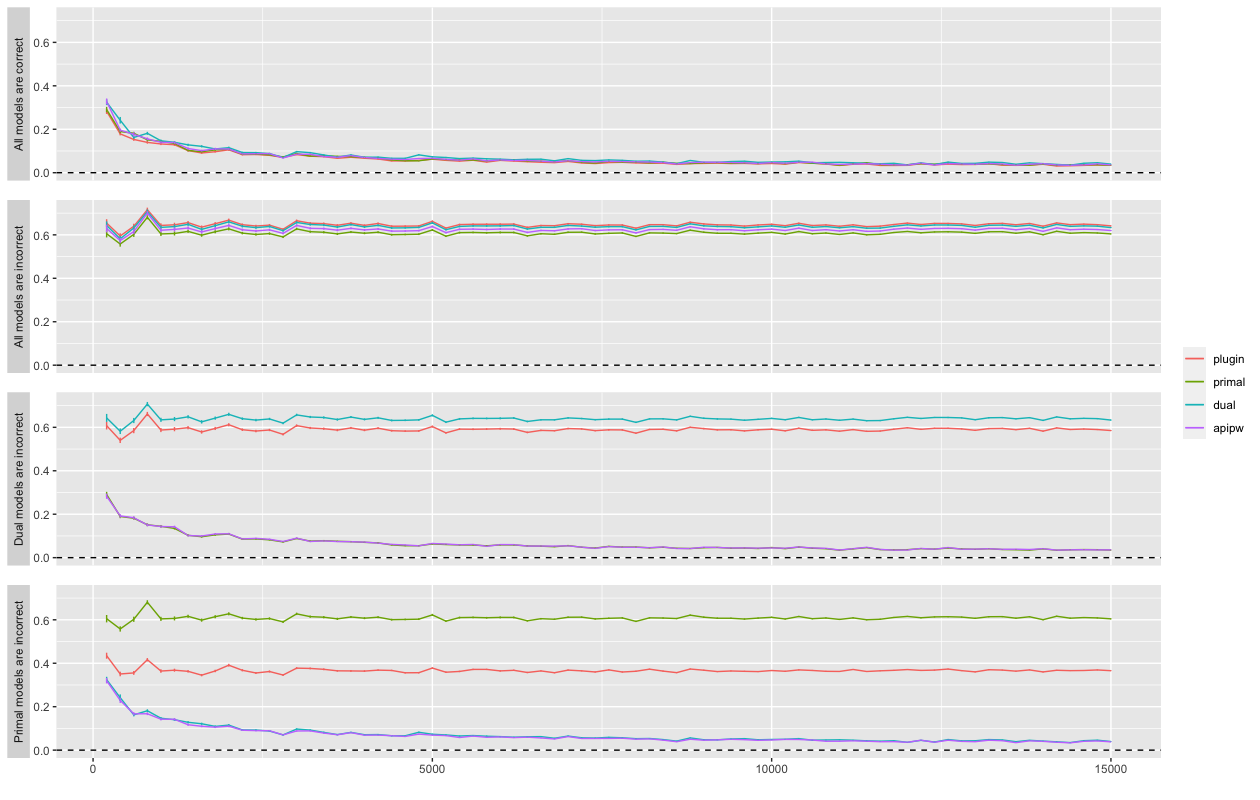}
		%		\label{fig:bias-ADMG-2b}
	\end{subfigure}
	\newline \newline
	\begin{subfigure}[b]{1\textwidth}
		\centering
		\includegraphics[scale=0.33]{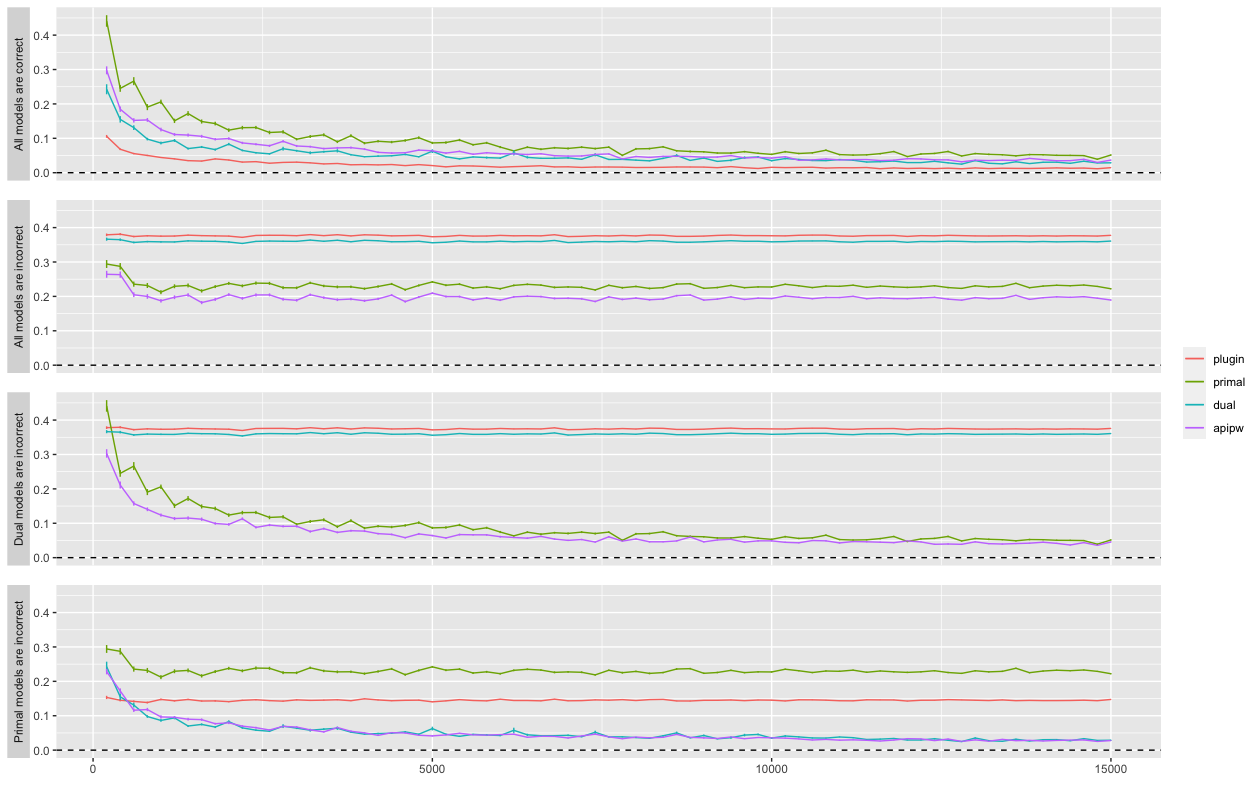}
		%		\label{fig:bias-ADMG-3}
	\end{subfigure}
	\caption{Bias behavior as a function of sample size when (1st row) all models are correctly specified, (2nd row) all models are incorrect, (3rd row) only dual models are misspecified, or (4th row) when only primal models are misspecified. The panel on (top) uses the ADMG in Fig.~\ref{fig:motiv}(b), and the (bottom) one uses the ADMG in Fig.~\ref{fig:p-fix_EIF}. The error bars illustrate deviations across the multiple iterations. }
	\label{fig:bias-ADMGs}
\end{figure}

\begin{figure}[!t]
	\begin{subfigure}[b]{0.5\textwidth}
		\centering
		\includegraphics[scale=0.18]{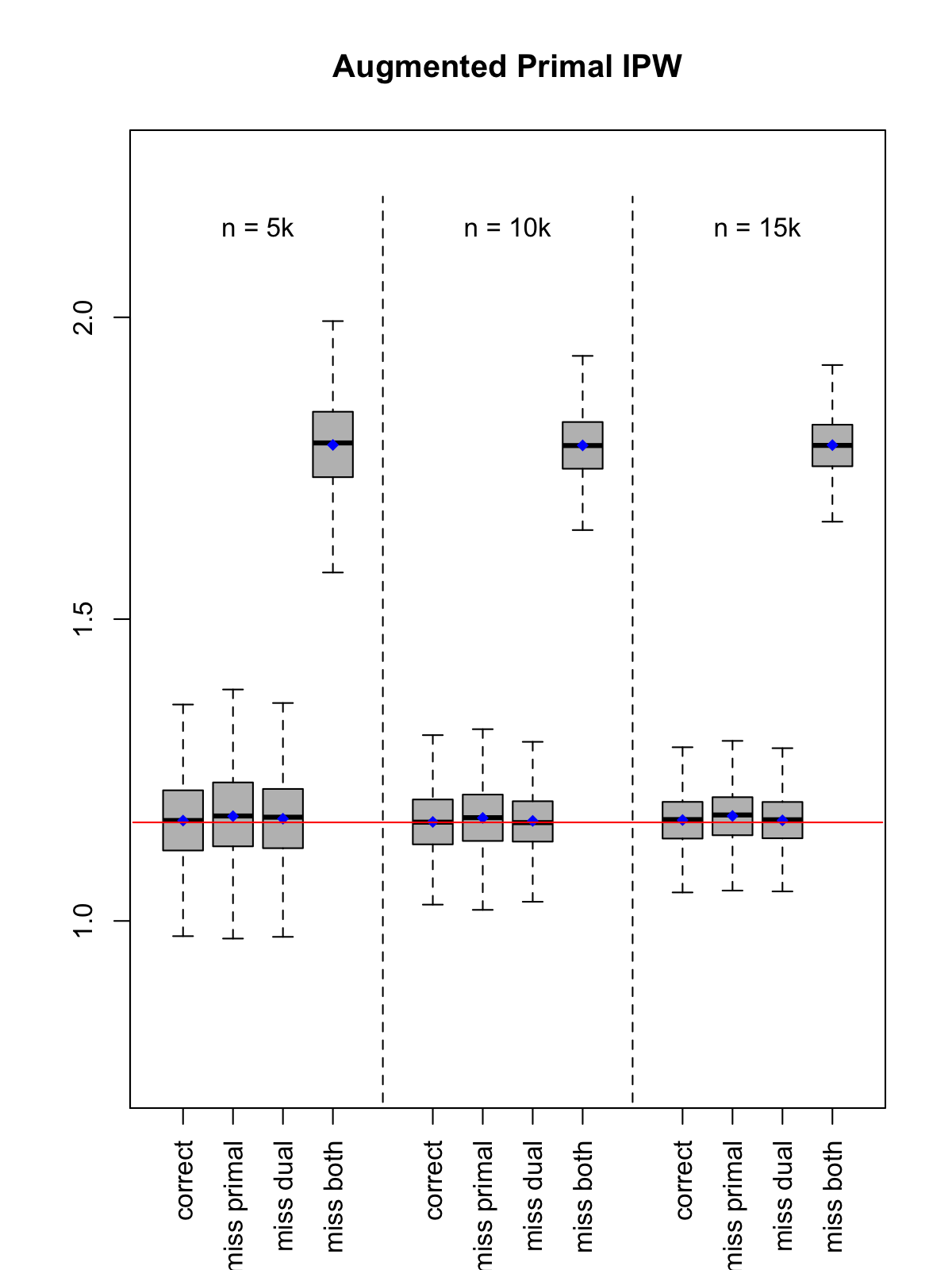}
		%		\label{fig:DR-ADMG-2b}
	\end{subfigure}
	\begin{subfigure}[b]{0.5\textwidth}
		\centering
		\includegraphics[scale=0.18]{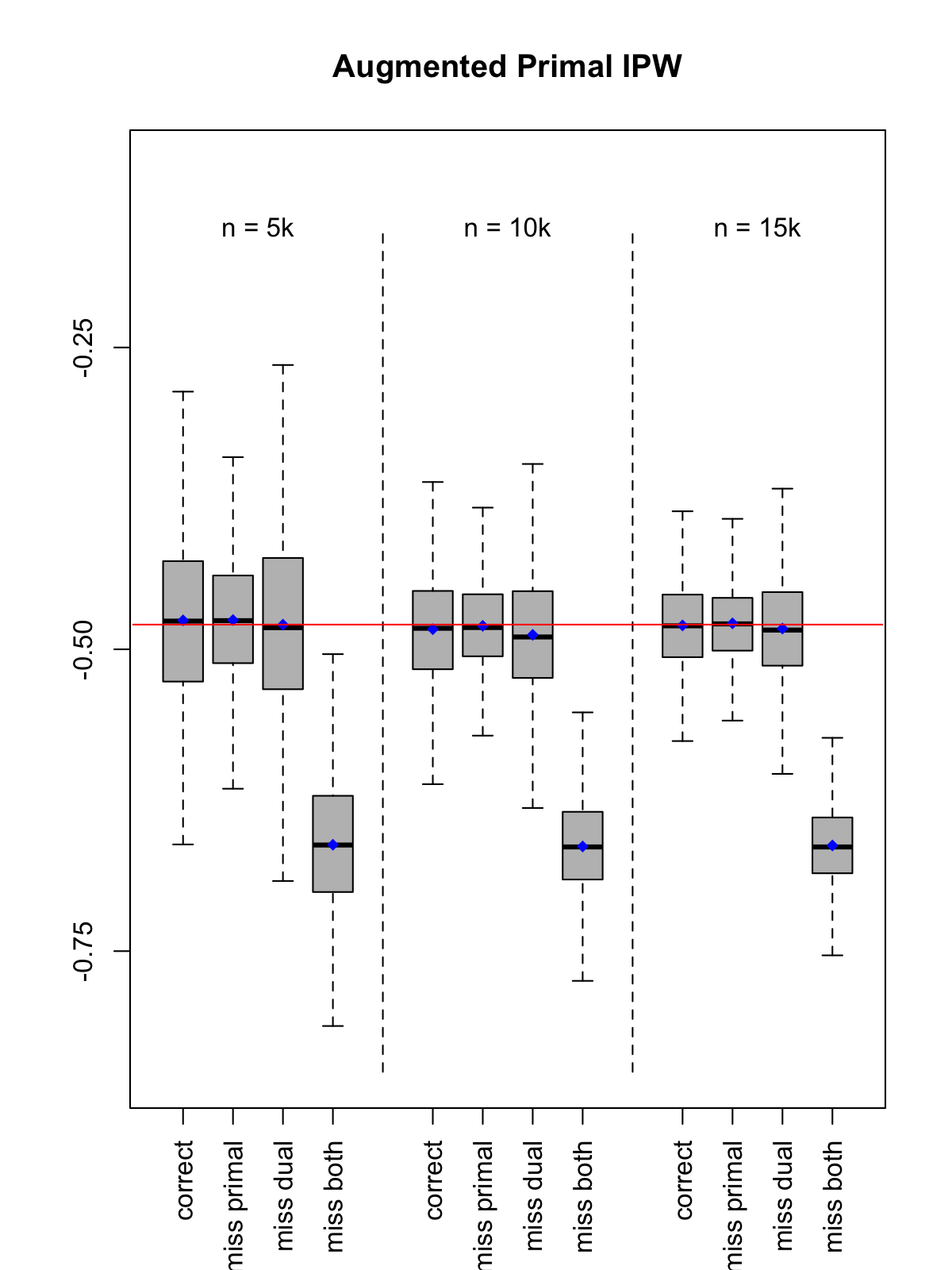}
		%		\label{fig:DR-ADMG-3}
	\end{subfigure}
	\caption{Demonstrating the double robustness of the Augmented Primal IPW estimator. The boxplot panel on the (left) uses the ADMG in Fig.~\ref{fig:motiv}(b), and the one on the (right) uses the ADMG in Fig.~\ref{fig:p-fix_EIF}. The red dashed lines indicate the true values of the ACE. }
	\label{fig:DR-ADMGs}
\end{figure}

\begin{figure}[!htbp]
	\centering
	\includegraphics[scale=0.28]{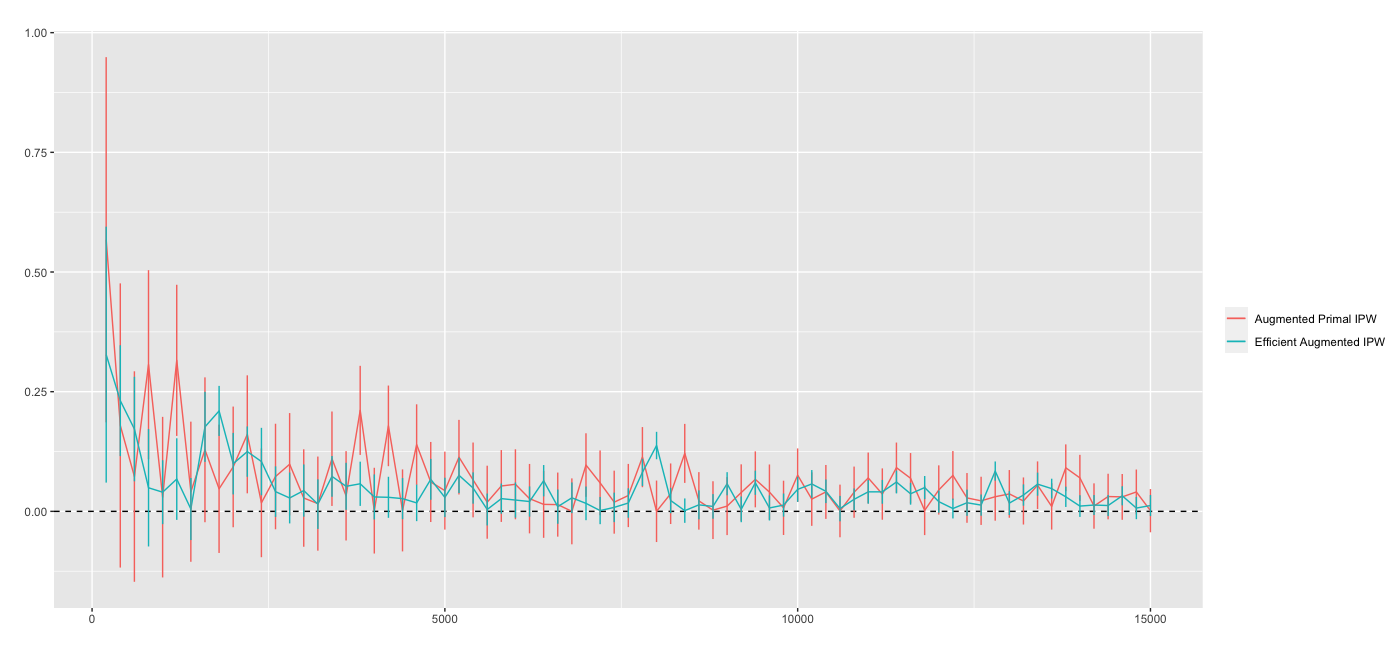}
	\caption{Comparing bias and variance behavior of the Augmented Primal IPW and efficient APIPW as a function of sample size, using the ADMG in Fig.~\ref{fig:p-fix_EIF}. The error bars illustrate the variance of the corresponding influence function. }
	\label{fig:var-ADMG-3}
\end{figure}

%##################################################
\vspace{0.3cm}
\noindent \textit{{Simulation 1.}} Bias behavior of Primal, Dual, and Augmented Primal IPW estimators
\vspace{0.10cm}

We evaluated the bias of the proposed estimators as a function of sample size for the causal effect of a binary treatment $T$ on a continuous outcome $Y$ in the ADMGs of Figures~\ref{fig:motiv}(b) and \ref{fig:p-fix_EIF}. The estimators for the corresponding counterfactual means are provided in Sections~\ref{subsec:ex_primalIPW}, \ref{subsec:ex_APlIPW}, and $5.1.$ These evaluations are carried under four different statistical models: (i) all conditional pieces in $\mathcal{M}_{\mathbb{M} \cup \mathbb{L}}$ are correctly specified, (ii) all conditional pieces in $\mathcal{M}_{\mathbb{M} \cup \mathbb{L}}$ are misspecified, (iii) only the conditional pieces in $\mathcal{M}_{\mathbb{M}}$ (terms in $\beta_{\text{dual}}$) are misspecified, and (iv) only the conditional pieces in $\mathcal{M}_{\mathbb{L}}$ (terms in $\beta_{\text{primal}}$) are misspecified. 
Results are reported and compared to the plug-in estimators in Figure~\ref{fig:bias-ADMGs}. The x-axis is the sample size $n$ with a range $(200, 15000)$ with increments of $200$. For a given sample size, we iterate over $100$ replications and report the absolute value of the estimator's bias for the causal effect, averaged over all the iterations. The error bars illustrate deviations from the mean where the length equals the standard error of the mean. 
The theory is borne out by the simulations: in both ADMGs, the plug-in estimator is unbiased only under scenario (i); the primal IPW estimator is unbiased under scenarios (i) and (iii); the dual IPW estimator is unbiased under scenarios (i) and (iv); the APIPW estimator remains unbiased when at least one set of models corresponding to those used in $\beta_{\text{primal}}$ or $\beta_{\text{dual}}$ are correctly specified, and is biased if they are all misspecified. That is, the influence function based estimator is unbiased in scenarios (i), (iii), and (iv). See Appendix  G for more simulations. 

%##################################################
\vspace{0.3cm}
\noindent \textit{{Simulation 2.}} Double robustness of Augmented Primal IPW 
\vspace{0.1cm}

To illustrate the  double robustness behavior of the APIPW estimator, we provide the boxplots of the causal effect of $T$ on $Y$ in the ADMGs of Figures~\ref{fig:motiv}(b) and \ref{fig:p-fix_EIF}, under different settings of model misspecifications. The simulations are replicated $1000$ times. For comparison, we plot the results for three sample sizes, $n=5k, 10k,$ and $15k$  in Figure~\ref{fig:DR-ADMGs}. The true causal effect is $1.16$ in Fig.~\ref{fig:motiv}(b) and is  $-0.48$ in Fig.~\ref{fig:p-fix_EIF}. The blue dot on each boxplot corresponds to the mean. The results can also be compared on the basis of  means and standard errors. 
We provide a similar set of boxplots in Appendix G to highlight the standard deviations across different scenarios.

%##################################################
\vspace{0.3cm}
\noindent \textit{{Simulation 3.}} Efficiency of Augmented Primal IPW in mb-shielded ADMGs
\vspace{0.1cm}

As pointed out in Section~\ref{sec:eff_if}, we can exploit constraints in a statistical model of an ADMG that is not nonparametrically saturated  to obtain more efficient estimators. We discussed the efficiency results for the special class of mb-shielded ADMGs. The ADMG in Figure~\ref{fig:motiv}(b) is nonparametrically saturated (no constraints). However, the ADMG in Figure~\ref{fig:p-fix_EIF} is not saturated (there are constraints) and it is an mb-shielded ADMG. Hence, we can obtain an estimator that is more efficient than the nonparametric influence function based estimator APIPW. 
The plots in Figure~\ref{fig:var-ADMG-3} summarize the simulation results comparing variances of  the APIPW and the efficient influence function based estimators for the causal effect in the ADMG of Figure~\ref{fig:p-fix_EIF}. The form of the efficient influence function is given in Section~$5.1.$ The y-axis is the bias and the error bars illustrate the variance of the estimators (which is the variance of the corresponding influence functions). As expected, though both estimators are unbiased, the one based on the efficient influence function offers lower variance with varying sample sizes. The absolute reduction in variance is greater and most beneficial at smaller sample sizes, but the relative reduction in variance remains the same across all sample sizes. This highlights the benefit of using the efficient IF in many practical applications where sample sizes may be small.

\begin{figure}[!t]
	\centering
	\includegraphics[scale=0.33]{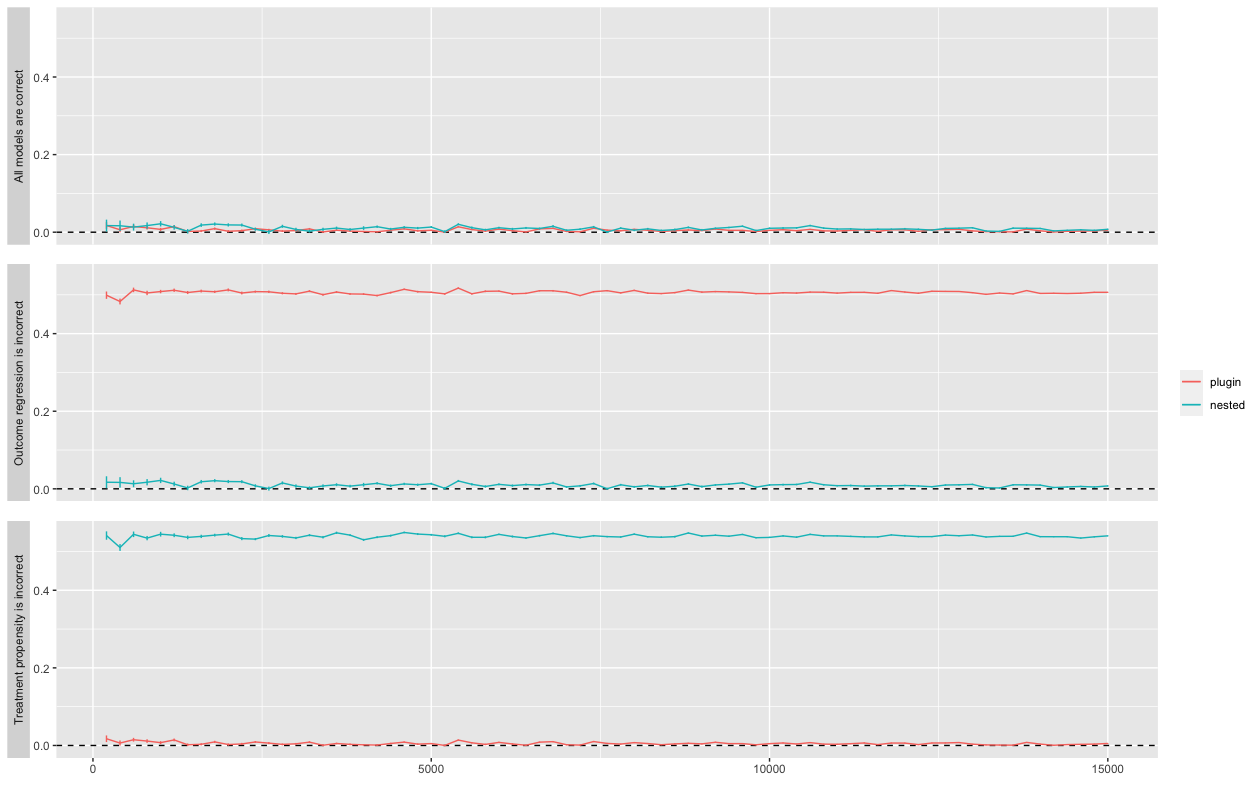}
	\caption{Bias behavior as a function of sample size when (1st row) all models are correctly specified, (2nd row) outcome regression is incorrect, and (3rd row) propensity score is incorrect, using the ADMG in Fig.~\ref{fig:general_id}. The error bars illustrate deviations across the multiple iterations. }
	\label{fig:bias-nested}
\end{figure}

%##################################################
\vspace{0.3cm}
\noindent \textit{{Simulation 4.}} Bias behavior of Nested IPW estimator when $T$ is not primal fixable
\vspace{0.1cm}

We evaluated the bias of the nested IPW estimator as a function of sample size for the causal effect of $T$ on $Y$ in the ADMG of Figure~\ref{fig:general_id} where treatment $T$ is not primal fixable. We compared this to the behavior of the plug-in estimator. These estimators  are discussed in Section~\ref{subsubsec:nested-ex}. The evaluations are carried under three different statistical models: (i) all conditional pieces in the district of $T$ along with all pieces outside of the district of $T$ are correctly specified, (ii) the outcome regression is misspecified (a term outside of the district of $T$), and (iii)  the district of $T$ is misspecified by misspecifying the treatment propensity. 
Results are reported and compared to the plug-in estimator in Figure~\ref{fig:bias-nested}. The x-axis is the sample size $n$ with a range $(200, 15000)$ with increments of $200$.  For a given sample size, we iterate over $100$ replications and report the absolute value of the bias for the causal effect, averaged across all iterations. The error bars illustrate deviations from the mean. and the length correspond to the standard error of the mean. The experiments support our theory: the nested IPW estimator is biased under scenario (iii), but remains unbiased under scenarios (i) and (ii).

%%%%%%%%%%%%%%%%%%%%%%%%%%%%%%%%%%%%%%%%%%%%%%%%

%%%%%%%%%%%%%%%%
% Conclusion
%%%%%%%%%%%%%%%%
\section{Conclusions}
\label{sec:conc}

In this paper, we bridged the gap between identification and estimation theory for the causal effect of a single treatment on a single outcome in hidden variable causal models associated with directed acyclic graphs (DAGs). 
We provided a simple graphical criterion, primal fixability, which when satisfied allows for the derivation of two novel IPW estimators -- primal and dual IPW. We further derived the nonparametric influence function under primal fixability of the treatment that yields the augmented primal IPW estimator and showed that it is doubly robust in the models used in primal and dual IPW estimators. We showed that in a strict subclass of primal fixability, when treatment is ordinary fixable, we can always find a valid adjustment set that permits the use of an influence function based estimator of augmented inverse probability weighted estimator to settings with hidden variables. 
We considered restrictions on the tangent space implied by the latent projection acyclic directed mixed graph (ADMG) of the hidden variable causal model. We provided an algorithm (Algorithm~\ref{alg:nps}), that is sound and complete for the purposes of checking the nonparametric saturation status of a hidden variable causal model as long as these hidden variables are unrestricted. Further, through the use of mb-shielded ADMGs, we provided a graphical criterion that defines a class of hidden variable causal models whose score restrictions resemble those of a DAG with no hidden variables. For the class of causal models that can be expressed as an mb-shielded ADMG, we then derived the form of the efficient influence function under primal fixability, that takes advantage of the Markov restrictions implied on the observed data.  
Finally, we developed a weighting estimation strategy for \emph{any} identifiable causal effect involving a single treatment and a single outcome. We call these estimators \emph{nested IPW} which only rely on the conditional densities involving variables in the district of the treatment. 
A natural extension of the present work is deriving influence function based estimators for any identifiable causal effect (including those that involve multiple treatment variables), and finding their most efficient versions by projecting onto the tangent space defined by equality restrictions, such as conditional independences and Verma constraints, implied by the causal model.

%%%%%%%%%%%%%%%%%%%%%%%%%%%%%%%%%%%%%%%%%%%%%%%%

%%%%%%%%%%%%%%%%
% Supplement
%%%%%%%%%%%%%%%%

\clearpage
%\appendix
\begin{appendices}

\makeatletter  \def\@seccntformat{\appendixname\   \csname thesection\endcsname. \quad} \makeatother

\section{Glossary of Terms and Notations}
\label{app:glossary}

\begin{table}[h]
	\addtolength{\tabcolsep}{8pt}
	\begin{center}
		
		{\small
			\begin{tabular}{ ll ll}
				\textbf{Symbol}     & \textbf{Definition}  &  \textbf{Symbol}     & \textbf{Definition} \vspace{0.2cm} \\ 
				$T$   & Treatment  &  $\G(V)$ & Graph $\G$ with vertices $V$    \\
				$Y, Y(t)$    & Outcome, potential outcome    & $\G(V, W)$ & A CADMG with fixed $W$   \\
				$V$    & Observed variables & $\G_{S}$ & Subgraph of $\G$ on vertices $S$    \\
				$H$    & Unmeasured variables   &   $\pa_\G(V_i)$ & Parents of $V_i$ in $\G$   \\ 
				$W$     & Fixed variables  & $\ch_\G(V_i)$ & Children of $V_i$ in $\G$  \\ 
				$\psi(t)$    & Target parameter $\E[Y(t)]$  &   $\an_\G(V_i)$ &  Ancestors of $V_i$ in $\G$   \\ 
				$U_{\psi_t}$ & Influence function for $\psi(t)$ & $\de_\G(V_i)$  & Descendants of $V_i$ in $\G$  \\
				$\mathbb{H}$ & Hilbert space&  $\mb_\G(V_i)$ & Markov blanket of $V_i$ in $\G$  \\
				$\Lambda$   & Tangent space  & $\tb_{\G}(V_i)$ & Markov pillow of $V_i$ in $\G$  \\  
				$\Lambda^{\perp}$ &  Orthogonal complement  &  $\dis_\G(V_i)$ & District of $V_i$ in $\G$  \\  
				$\cal U$ &  Class of all influence functions & $D_T$ &  District of $T$     \\ 
				$U^{\text{eff}}_{\psi}$ & Efficient influence function & ${\cal D}(\G)$  & Set of all districts in $\G$  \\
				$\pi[h \mid \Lambda]$ & Projection of $h$ onto $\Lambda$ & $\tau$ & A valid topological order  \\ 
				$\mathbb{C}$ & Pre-treatment variables & $V_i \prec V_j$ & $V_i$ precedes $V_j$   \\
				$\mathbb{L}$ & Post-treatment variables in $D_T$ & $\{\prec V_i\}$ & Vertices preceding $V_i$ \\
				$\mathbb{M}$ & Variables not in $\mathbb{C} \cup \mathbb{L}$ & $\phi_{V_i}(\G)$ & Fixing $V_i$ in $\G$ \\
				$\mathbb{M}^*$ & Inverse Markov pillow of $T$ & $\phi_{V_i}(q_V;\G)$ & Fixing $V_i$ in $q_V(\cdot \mid \cdot)$  \\
				$\mathbb{P}_n$ & Empirical distribution & $\phi_{\neg{S}}(\G)$ & CADMG from fixing $V\setminus S$ recursively  \\
				$S(V)$ & Score of $p(V)$ &  $Y^*$& $\an_{\G_{V \setminus T}}(Y)$   \\
			\end{tabular}
		}
	\end{center}
\end{table}

%%%%%%%%%%%%%%%%%%%%%%%%%%%%%%%%%%%%%%%%%%%%%

\section{Example of Latent Projection} 
\label{app:latent_proj}

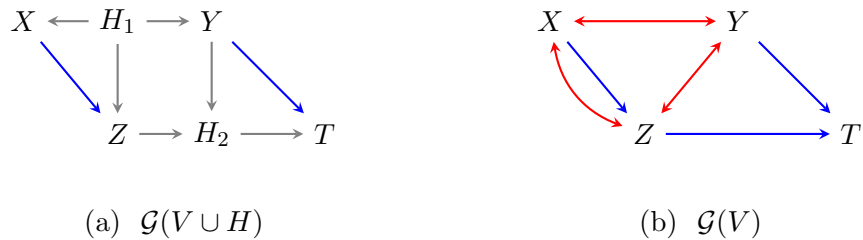
\begin{figure}[!h]
	\begin{center}
		\scalebox{1}{
			\begin{tikzpicture}[>=stealth, node distance=1.5cm]
				\tikzstyle{format} = [thick, circle, minimum size=1.0mm, inner sep=0pt]
				\tikzstyle{square} = [draw, thick, minimum size=1.0mm, inner sep=3pt]
				\begin{scope}
					\path[->,  thick]
					node[] (x) {$X$}
					node[right of=x, xshift=1cm] (y) {$Y$}
					node[right of=x, xshift=-0.25cm] (u1) {$H_1$} 
					node[below of=u1] (z) {$Z$}
					node[below of=y] (u2) {$H_2$}
					node[right of=u2] (t) {$T$}
					(u1) edge[gray] (x)
					(u1) edge[gray] (z)
					(u1) edge[gray] (y)
					(y) edge[gray] (u2)
					(z) edge[gray] (u2)
					(u2) edge[gray] (t)
					(x) edge[blue] (z)
					(y) edge[blue] (t)
					node[below of=z, xshift=0.75cm, yshift=0.3cm] (cap) {(a) \ $\G(V \cup H)$}
					;
				\end{scope}
				\begin{scope}[xshift=7cm]
					\path[->,  thick]
					node[] (x) {$X$}
					node[right of=x, xshift=1cm] (y) {$Y$}
					node[right of=x, xshift=-0.25cm] (u1) {} 
					node[below of=u1] (z) {$Z$}
					node[below of=y] (u2) {}
					node[right of=u2] (t) {$T$}
					node[left of=t, xshift=.1cm] (d1) {}
					node[above left of=t, xshift=.3cm, yshift=-.3cm] (d2) {}
					(x) edge[red, <->] (y)
					(z) edge[blue] (t)
					(x) edge[blue] (z)
					(x) edge[red, <->, bend right] (z)
					(z) edge[red, <->] (y)
					(y) edge[blue] (t)
					node[below of=z, xshift=0.75cm, yshift=0.3cm] (cap) {(b) \ $\G(V)$}
					;
				\end{scope}
			\end{tikzpicture}
		}
	\end{center}
	\caption{(a) A hidden variable DAG $\G(V \cup H). $ (b) The ADMG $\G(V)$ obtained via latent projection. }
	\label{fig:projection}
\end{figure}

Fig.~\ref{fig:projection}(a) shows an illustrative example of a hidden variable DAG $\G(V \cup H)$ and the ADMG $\G(V)$ obtained by applying rules of latent projection described in Section~\ref{sec:prelims}.

%%%%%%%%%%%%%%%%%%%%%%%%%%%%%%%%%%%%%%%%%%%%%

\section{Marginalization, Conditioning, and Fixing in Kernels}
\label{supp:kernels}

A kernel $q_V(V \mid W)$ is a mapping from values of $W$ to normalized densities over $V.$ That is, $\sum_V q_V(V \mid W=w) = 1, \forall w \in W.$ For any set of variables $X \subseteq V,$ marginalization and conditioning in a kernel are defined as follows.
\begin{align*}
	{q_{V\setminus X}(V\setminus X \mid W)} &\equiv \sum_X q_V(V \mid W), \text{ and} \\ 
	q_V(V\setminus X \mid X, W) &\equiv \frac{q_V(V \mid W)}{q_V(X \mid W)}.
\end{align*}

The notation $q_V(\cdot \mid X)$ makes clear which variables appearing past the ``conditioning'' bar in a kernel are fixed as opposed to simply conditioned on. That is, if a variable $X_i \not\in V,$ then it is fixed, else it is conditioned on. Occasionally, fixing operations may also simplify to marginalization or conditioning events. We illustrate these concepts with a simple example.

\begin{figure}[!h]
	\begin{center}
		\scalebox{0.8}{
			\begin{tikzpicture}[>=stealth, node distance=1.5cm]
				\tikzstyle{format} = [thick, circle, minimum size=1.0mm, inner sep=0pt]
				\tikzstyle{square} = [draw, thick, minimum size=1.0mm, inner sep=3pt]
				
				\begin{scope}
					\path[->, thick]
					node[] (c1) {$Z_1$}
					node[right of=c1] (c2) {$Z_2$}
					node[right of=c2] (t) {$T$}
					node[right of=t] (y) {$Y$}
					(c1) edge[blue] (c2)
					(c2) edge[blue] (t)
					(t) edge[blue] (y)
					(c1) edge[<->, red, bend right] (t)
					(c1) edge[<->, red, bend left] (y)
					node[below right of=c2, xshift=-0.35cm, yshift=0cm] (label) {(a) $\G$}
					node[below of=label, yshift=0.5cm] {$p(V) = p(Z_2 \mid Z_1) \times p(Z_1) \times p(T, Y \mid Z_1, Z_2)$}
					
					;
				\end{scope}
				
				\begin{scope} [yshift=0.0cm, xshift=9.5cm]
					\path[->, thick]
					node[] (c1) {$Z_1$}
					node[square, right of=c1] (c2) {$z_2$}
					node[right of=c2] (t) {$T$}
					node[right of=t] (y) {$Y$}
					(c2) edge[blue] (t)
					(t) edge[blue] (y)
					(c1) edge[<->, red, bend right] (t)
					(c1) edge[<->, red, bend left] (y)
					node[below right of=c2, xshift=-0.35cm, yshift=0cm] (label) {(b) $\phi_{Z_2}(\G)$}
					node[below of=label, yshift=0.5cm] {$q_{Z_1, T, Y}(Z_1, T, Y \mid Z_2) = p(Z_1) \times p(T, Y \mid Z_1, Z_2)$}
					;
				\end{scope}
				
				\begin{scope} [yshift=-3.5cm, xshift=0.0cm]
					\path[->, thick]
					node[square] (c1) {$z_1$}
					node[square, right of=c1] (c2) {$z_2$}
					node[right of=c2] (t) {$T$}
					node[right of=t] (y) {$Y$}
					(c2) edge[blue] (t)
					(t) edge[blue] (y)
					node[below right of=c2, xshift=-0.35cm, yshift=0cm] (label) {(c) $\phi_{\{Z_1, Z_2\}}(\G)$}
					node[below of=label, yshift=0.5cm] {$q_{T, Y}(T, Y \mid Z_1, Z_2) = \sum_{Z_1} p(Z_1) \times p(T, Y \mid Z_1, Z_2)$}
					;
				\end{scope}
				
				\begin{scope}[yshift=-3.5cm,xshift=9.5cm]
					\path[->, thick]
					node[square] (c1) {$z_1$}
					node[square, right of=c1] (c2) {$z_2$}
					node[square, right of=c2] (t) {$t$}
					node[right of=t] (y) {$Y$}
					(t) edge[blue] (y)
					node[below right of=c2, xshift=-0.35cm, yshift=0cm] (label) {(d) $\phi_{\{Z_1, Z_2, T\}}(\G)$}
					node[below of=label, yshift=0.5cm] {$q_Y(Y \mid T, Z_1, Z_2) = \frac{\sum_{Z_1} p(Z_1) \times p(T, Y \mid Z_1, Z_2)}{\sum_{Z_1} p(Z_1) \times p(T \mid Z_1, Z_2)}$}
					;
				\end{scope}
				
			\end{tikzpicture}
		}
	\end{center}
	\vspace{-0.5cm} 
	\caption{An example to illustrate fixing and kernel operations. }
	\label{fig:kernel_example}
\end{figure}
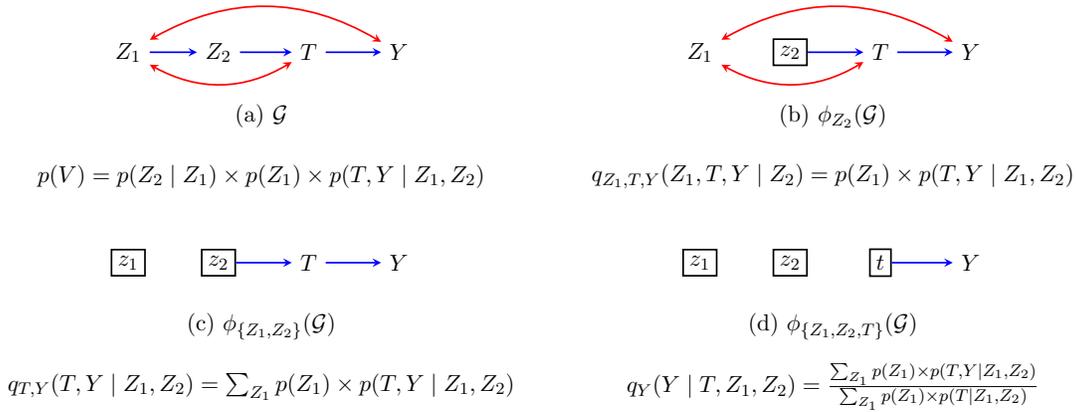

Consider the ADMG shown in Fig.~\ref{fig:kernel_example}(a) and fix the kernel of interest to be $q_Y(Y \mid T, Z_1, Z_2),$ i.e., a kernel where all other variables except $Y$ are fixed. A valid fixing sequence in order to obtain such a kernel from the joint $p(V)$ is $(Z_2, Z_1, T).$ Fixing $Z_2$ entails dividing by the simple conditional $p(Z_2 \mid Z_1)$ and yields the CADMG $\phi_{Z_2}(\G)$ and corresponding kernel $q_{Z_1, T, Y}(Z_1, T, Y \mid Z_2)$ shown in Fig.~\ref{fig:kernel_example}(b). In order to fix $Z_1,$ we must divide by the kernel $q_{Z_1, T, Y}(Z_1 \mid Z_2, T, Y).$ By rules of conditioning and marginalization in kernels,

\begin{align*}
	q_{Z_1, T, Y}(Z_1 \mid Z_2, T, Y) \equiv \frac{q_{Z_1, T, Y}(Z_1, T, Y \mid Z_2)}{q_{Z_1, T, Y}(T, Y \mid Z_2)} \equiv \frac{q_{Z_1, T, Y}(Z_1, T, Y \mid Z_2)}{\sum_{Z_1} q_{Z_1, T, Y}(Z_1, T, Y \mid Z_2)}
\end{align*}

Fixing $Z_1$ and evaluating the above expression gives us the CADMG and corresponding kernel shown in Fig.~\ref{fig:kernel_example}(c). That is, fixing $Z_1$ in the kernel $q_{Z_1, T, Y}(Z_1 \mid Z_2, T, Y),$ simplifies to marginalization of $Z_1.$ Finally, applying rules of conditioning and marginalization to the kernel $q_{T, Y}(T, Y \mid Z_1, Z_2)$ we can obtain the kernel $q_{T, Y}(T \mid Z_1, Z_2, Y).$ Dividing by this corresponds to fixing $T,$ giving us the CADMG and desired kernel shown in Fig.~\ref{fig:kernel_example}(d). 

\vspace{0.25cm} 

%%%%%%%%%%%%%%%%%%%%%%%%%%%%%%%%%%%%%%%%%%%%%

\section{An Overview of Semiparametric Estimation Theory}
\label{supp:semiparam}

Assume a statistical model $\mathcal{M} = \{p_\eta(Z): \eta \in \Gamma\}$ where $\Gamma$ is the parameter space and $\eta$ is the parameter indexing a specific model. We are often interested in a function $\psi: \eta \in \Gamma \mapsto \psi(\eta) \in \mathbb{R}$; i.e., a parameter that maps the distribution $P_\eta$ to a scalar number in $\mathbb{R}$, such as an identified average causal effect. 
(For brevity, we sometimes use $\psi$ instead of $\psi(\eta),$ which should be obvious from context.) 
Truth is denoted by $P_{\eta_0}$ and $\psi_0.$
An estimator $\widehat{\psi}_n$ of a scalar\footnote{Here, our focus is on estimation of $\psi = \E[Y(t)]$ which is a scalar parameter.  For an extension to a vector valued functional in $\mathbb{R}^q, q > 1,$  refer to \cite{tsiatis2007semiparametric,bickel1993efficient}.} parameter $\psi$ based on $n$ i.i.d copies $Z_1, \ldots, Z_n$ drawn from $p_{\eta}(Z)$, is \textit{asymptotically linear} if there exists a measurable random function $U_\psi(Z)$ with mean zero and finite variance such that 
\begin{align}
	\sqrt{n} \times (\widehat{\psi}_n - \psi) = \frac{1}{\sqrt{n}} \times \sum_{i = 1}^n U_\psi(Z_i) + o_p(1), 
	\label{eq:asym_linear}
\end{align}%
where $o_p(1)$ is a term that converges in probability to zero as $n$ goes to infinity. The random variable $U_\psi(Z)$ is called the \textit{influence function} of the estimator $\widehat{\psi}_n$. The term influence function comes from the robustness literature \citep{hampel1974influence}.

Before mentioning the asymptotic properties of an asymptotically linear estimator, it is worth noting that in asymptotic theory, we can sometimes construct \textit{super efficient} estimators, e.g. Hodges estimator, that have undesirable local properties associated with them. Therefore, the analysis is oftentimes restricted to \textit{regular}\footnote{
	Given a collection of probability laws $\mathcal{M}$, an estimator $\widehat{\psi}$ of $\psi (P)$ is said to be regular in $\mathcal{M}$ at $P$ if its convergence to $\psi (P)$ is locally uniform \citep{van2000asymptotic}.
}
and asymptotically linear (RAL) estimators to avoid such complications. Although most reasonable estimators are RAL, regular estimators do exist that are not asymptotically linear. However, as a consequence of \cite{hajek1970characterization} representation theorem, the most efficient regular estimator is asymptotically linear; hence, it is reasonable to restrict attention to RAL estimators. According to \cite{newey1990semiparametric}, the influence function of a RAL estimator is the same as the influence function of its estimand. Further, there is a bijective correspondence between RAL estimators and influence functions. 

By a simple consequence of the central limit theorem and Slutsky's theorem, it is straightforward to show that the RAL estimator $\widehat{\psi}_n$ is  \textit{consistent and asymptotically normal} (CAN), with asymptotic variance equal to the variance of its influence function $U_\psi$, 
\begin{align}
	\sqrt{n} \times (\widehat{\psi}_n - \psi) \ \xrightarrow[]{d} \ N\big(0, \ \text{var}(U_\psi) \big). 
\end{align}

The first step in dealing with a semiparametric model, is to consider a simpler finite-dimensional parametric submodel that is contained within the semiparametric model and it contains the truth. Consider a (regular) parametric submodel $\mathcal{M}_{\text{sub}} =  \{P_{\eta_\kappa}: \kappa \in {[0, 1)} \text{ where } P_{\eta_{\kappa = 0}} = P_{\eta_0} \}$ of the model $\mathcal{M}$. Given $P_{\eta_0}$, define the corresponding score to be $S_{\eta_0}(Z) = \displaystyle \frac{d}{d\kappa}\log p_{\eta_\kappa}(Z) \Big|_{\kappa = 0}.$ 
It is known that
\begin{align}
	\frac{d}{d\kappa} \psi(\eta_\kappa)\Big|_{\kappa = 0} = \E\Big[ U_\psi(Z) \times S_{\eta_0}(Z) \Big],
	\label{eq:pathwise_deriv}
\end{align} %
where $\psi(\eta_\kappa)$ is the target parameter in the parametric submodel, $U_\psi(Z)$ is the corresponding influence function evaluated at law $P_{\eta_0}$, $S_{\eta_0}(Z)$ is the score of the law $P_{\eta_0}$, and the expectation is taken with respect to $P_{\eta_0}.$ Equation~\ref{eq:pathwise_deriv} provides an easy way to derive an influence function for the parameter $\psi$. In the next subsection, we use this equation to derive an influence function for our target $\psi = \E[Y(t)]$ and discuss its properties. 

Influence functions provide a geometric view of the behavior of RAL estimators.
Consider a Hilbert space\footnote{The Hilbert space of all mean-zero scalar functions is the $L^2$ space. For a precise definition of Hilbert spaces see \cite{luenberger1997optimization}.} $\mathbb{H}$ of all mean-zero scalar functions, equipped with an inner product defined as $\E[h_1\times h_2], h_1, h_2 \in \mathbb{H}.$ The \emph{tangent space} in the model $\mathcal{M}$, denoted by $\Lambda$, is defined to be the mean-square closure of parametric submodel tangent spaces, where a parametric submodel tangent space is the set of elements $\Lambda_{\eta_\kappa} = \{\alpha S_{\eta_\kappa}(Z)\}, \alpha$ is a constant and $S_{\eta_\kappa}$ is the score for the parameter $\psi_{\eta_\kappa}$ for some parametric submodel. In mathematical form, $\Lambda =  \overline{[\Lambda_{\eta_\kappa}]}$. 

The tangent space $\Lambda$ is a closed linear subspace of the Hilbert space $\mathbb{H}$ ($\Lambda \subseteq \mathbb{H}$). The orthogonal complement of the tangent space, denoted by $\Lambda^{\perp},$ is defined as $\Lambda^{\perp} = \{h \in \mathbb{H} \mid \E[h\times h'] = 0, \forall h' \in \Lambda\}.$ Note that $\mathbb{H} = \Lambda \oplus \Lambda^\perp$, where $\oplus$ is the direct sum, and $\Lambda \cap \Lambda^\perp = \{0\}$. Given an arbitrary element $h \in \Lambda^\perp,$ it holds that for any submodel ${\cal M}_{\text{sub}},$ with score $S_{\eta_0}$ corresponding to $P_{\eta_0},$ $\E[h\times S_{\eta_0}] = 0$. Consequently, using Eq.~\ref{eq:pathwise_deriv}, $h + U_\psi(Z)$ is also an influence function. The vector space $\Lambda^\perp$ is then of particular importance because we can now construct the class of all influence functions, denoted by $\cal U$, as ${\cal U} = U_\psi(Z) + \Lambda^\perp.$ Upon knowing a single IF $U_\psi(Z)$ and the tangent space orthogonal complement $\Lambda^\perp$, we can obtain the class of all possible RAL estimators that admit the CAN property. 

Out of all the influence functions in $\cal U$ there exists a unique one which lies in the tangent space $\Lambda,$ and which yields the most efficient RAL estimator by recovering the \emph{semiparametric efficiency bound}. This efficient influence function can be obtained by projecting any influence function, call it $U^*_\psi$, onto the tangent space $\Lambda$. This operation is denoted by $U^{\text{eff}_\psi} = \pi[U^*_\psi \mid \Lambda],$ where ${U^{\text{eff}}_\psi}$ denotes the efficient IF.

On the other hand, if the tangent space contains the entire Hilbert space, i.e., $\Lambda = \mathbb{H},$ then the statistical model $\cal M$ is called a \textit{nonparametric} model. In a nonparametric model, we only have one influence function since $\Lambda^\perp = \{0\}.$ This unique influence function can be obtained via Eq.~\ref{eq:pathwise_deriv} and corresponds to the efficient influence function ${U^{\text{eff}}_\psi}$ (the unique element in the tangent space $\Lambda$) in the nonparametric model $\cal M.$ For a detailed description of the concepts outlined here, please refer to \cite{tsiatis2007semiparametric, bickel1993efficient}.

\subsection{ An Overview of Inference for the Adjustment Functional}
\label{supp:DAG_inference}

Having briefly discussed causal models of a DAG in Section~\ref{sec:prelims}, we now provide a short overview of estimation theory surrounding the target $\psi(t)$ in such a model. 

If a parametric likelihood can be correctly specified for the statistical DAG model of the observed data distribution, then an efficient estimator for $\psi(t)$ may be derived using the plug-in principle.  In the commonly assumed  case where the DAG corresponding to the observed data distribution is complete, the plug-in estimator for $\psi(t)$ reduces to $\mathbb{P}_n[ {\mu}_t(C; \widehat{\eta_1})],$ where $\mathbb{P}_n[.] \coloneqq \frac{1}{n}\sum_{i=1}^n (.)$, $\mu_t(C; \eta_1)$ is the correctly specified parametric form for $\E[Y \mid T=t, C]$, and $\widehat{\eta_1}$ are the maximum likelihood values of ${\eta_1}$.

Since assuming a correctly specified parametric observed data likelihood, or even a correctly specified outcome regression $\mu_t(C; \eta)$ is unrealistic in practice, a variety of other estimators have been developed that place \emph{semiparametric} restrictions on the observed data distribution.
One such estimator, based on \emph{inverse probability weighting (IPW)}, seeks to compensate for a biased treatment assignment by reweighing observed outcomes of units assigned $T=t$ by the inverse of the normalized treatment assignment probability $p(T=t \mid C)$.  If this probability has a known parametric form $\pi_t(C; \eta_2) \equiv p(T=t \mid C)$, the IPW estimator takes the form
$\mathbb{P}_n [ \frac{\mathbb{I}(T = t)}{{\pi}_t(C; \widehat{\eta_2})} \times Y ],$ where $\mathbb{I}(.)$ is the indicator function, and $\widehat{\eta_2}$ are the maximum likelihood estimates of $\eta_2.$ While the IPW estimator is inefficient, it is simple to implement, and is often used in cases where the treatment assignment model $\pi_t(C; \eta_2)$ is known by design, as is often the case in controlled trials.

The plug-in and IPW estimators of $\psi(t)$ are both \emph{$\sqrt{n}$-consistent and asymptotically normal} if the models they rely on, $\mu_t(C; \eta_1)$ and $\pi_t(C; \eta_2)$ respectively, are parametric and correctly specified. Otherwise, these estimators are no longer consistent.  If flexible models are used for $\mu_t(C)$ and $\pi_t(C)$ instead, the resulting estimators may remain consistent, but converge to the true value of $\psi(t)$ at unacceptably slow rates; see \cite{chernozhukov2018double} for examples.

A principled alternative is to consider influence functions and RAL estimators. In the nonparametric saturated model, corresponding to the complete DAG, the unique influence function for $\psi(t)$ is given by {\small $U_{\psi_t} = \frac{\mathbb{I}(T=t)}{{\pi}_t(C)} \times \{ Y - {\mu}_t(C) \} + {\mu}_t(C) - \psi(t),$} yielding the \emph{AIPW} estimator: 
{\small$ \mathbb{P}_n \Big[ \frac{\mathbb{I}(T=t)}{{\pi}_t(C; \widehat{\eta_2})} \times \{ Y - {\mu}_t(C; \widehat{\eta_1}) \} + {\mu}_t(C; \widehat{\eta_1}) \Big].$} Given the standard factorization of the complete DAG as {\small$p(Y\mid A,C) \times p(A\mid C) \times p(C),$} the propensity score model $\pi_t(C)$ and the outcome regression model $\mu_t(C)$ are variationally independent. Further, the bias of this estimator is a product of the biases of its nuisance functions $\pi_t(C)$ and $\mu_t(C)$.  As a result, the AIPW estimator exhibits the \emph{double robustness property}, where it remains consistent if \emph{either} of the two nuisance models $\pi_t(C)$ or $\mu_t(C)$ is specified correctly, even if the other is arbitrarily misspecified.

In a semiparametric model of a DAG, which is defined by conditional independence restrictions on the tangent space implied by the DAG factorization, the above influence function can be projected onto the tangent space of the model to improve efficiency; see \cite{rotnitzky2019efficient} for details.

%%%%%%%%%%%%%%%%%%%%%%%%%%%%%%%%%%%%%%%%%%%%%
\newpage
\section{Intuitions for  APIPW in the Nonparametric Model}
%\st{Nonparametric IF in Primal Fixability}
\label{supp:intuition_primal_IF}

Given a post treatment variable $V_i$ and its conditional density $p(V_i \mid \tb_\G(V_i))$ in the identified functional of $\psi(t)$ in Eq.~\ref{eq:tian2002rearrange}, there is a corresponding term in the influence function $U_{\psi_t}$ in Theorem~\ref{thm:IF_childless} of the form 
\begin{align}
	\label{eq:intuit_IF_supp}
	f_1(\prec V_i) \times \Big( f_2(\preceq V_i) - \sum_{V_i} f_2(\preceq V_i) \times p\big(V_i \mid \tb_\G(V_i)\big) \Big),
\end{align} %
where $f_1(\prec V_i)$ denotes a function of variables that come before $V_i$ in the topological order, a.k.a history/past of $V_i.$ Similarly, $f_2(\preceq V_i)$ is a function of past of $V_i$ and including $V_i$ itself. $f_1(\prec V_i)$ is defined as follows,

\begin{align}
	f_1(\prec V_i)  = 
	\begin{cases} 
		\displaystyle \frac{\I(T=t)}{ \prod_{L_i \prec V_i} \ p(L_i \mid \tb_\G(L_i))}, & \mbox{if } V_i \in \mathbb{M} 
		\\
		\\
		\displaystyle \frac{\prod_{M_i \prec V_i} p(M_i \mid \tb_\G(M_i)) \vert_{T=t} }{\prod_{M_i \prec V_i} p(M_i \mid \tb_\G(M_i))}, & \mbox{if } V_i \in \mathbb{L}
		\label{eq:weights_IF}
	\end{cases} 
	\\ \nonumber 
\end{align}

Interestingly, these weights resemble the ones in $\psi_{\text{primal}}$ and $\psi_{\text{dual}}$ that we introduced in Lemmas~\ref{lem:primal_ipw} and \ref{lem:dual_ipw}, if the target were the counterfactual mean $\E[V_i(t)].$ That is,

\begin{align*}
	\psi_{v_i, \text{primal}}
	&= \E\Big[ \ \frac{\I(T=t)}{ \prod_{L_i \prec V_i} \ p(L_i \mid \tb_\G(L_i))} \times \sum_T \prod_{L_i \prec V_i} \ p(L_i \mid \tb_\G(L_i)) \times V_i \ \Big] 
	\\
	\psi_{v_i, \text{dual}}
	&= \E\Big[ \ \frac{\prod_{M_i \prec V_i} p(M_i \mid \tb_\G(M_i)) \vert_{T=t} }{\prod_{M_i \prec V_i} p(M_i \mid \tb_\G(M_i))} \times V_i \ \Big].\\
\end{align*}

However, to calculate the effect of $T$ on $Y$, we do not need to worry about the effect of $T$ on intermediate variables $V_i.$ In Lemma~\ref{lem:dr}, we show that the influence function $U_{\psi_t}$ in Theorem~\ref{thm:IF_childless} cleverly uses the information in these intermediate primal and dual estimators and yields a doubly robust estimator for $\psi_t.$

%%%%%%%%%%%%%%%%%%%%%%%%%%%%%%%%%%%%%%%%%%%%%
\newpage
\section{Primal Fixing Operator} 
\label{app:primal-fixing-operator}

In this section we introduce the primal fixing operator, which is a generalization of the fixing operator used in the definition of the nested Markov model. We show how a valid sequence of primal fixing serves as a useful identification strategy whose complexity lies between a single use of the primal fixing criterion (as seen in Section~\ref{sec:nps}) and truncated nested Markov factorization (as seen in Section~\ref{sec:nested}.) Proofs are deferred to Appendix~\ref{app:proofs}. 

Given a kernel $q_V(V \mid W)$ that is nested Markov with respect to a CADMG $\G(V, W),$ we define a vertex $V_i$ to be primal fixable (p-fixable) if it has no bidirected path to any of its children, i.e., $\dis_\G(V_i) \cap \ch_\G(V_i) = \emptyset.$ If $V_i$ is primal fixable in $\G(V, W)$, the graphical operation of primal fixing $V_i$ applied to $\G,$ denoted by $\primal_{V_i}(\G),$ yields a new CADMG $\G(V\setminus V_i, W \cup V_i)$ where all incoming edges into $V_i$ are removed and $V_i$ is fixed to some value $v_i.$  That is, the graphical operation of primal fixing is exactly the same as ordinary fixing. However, the two forms of fixing differ in the definition of the probabilistic operators; the probabilistic operator for primal fixing is defined as follows. Given a kernel $q_V(V \mid W)$ that nested factorizes with respect to a CADMG $\G(V, W)$ in which $V_i$ is primal fixable, let $D_{V_i}$ denote the district of $V_i$ in $\G(V, W)$. Then the probabilistic operation of primal fixing $V_i$ denoted by $\primal_{V_i}(q_V; \G)$ is given by,

{\small
	\begin{align}
		\label{eq:primal_fixing}
		\primal_{V_i}(q_V; \G) \equiv q_{V\setminus T}(V\setminus V_i \mid W \cup V_i) &\equiv \frac{q_V(V \mid W)}{q_{D_{V_i}}(V_i \mid \mb_\G(V_i), W)} \nonumber \\
		&= q_V(V \mid W) \times \frac{\sum_{V_i} \prod_{D_i \in D_{V_i}} q_{V}(D_i \mid \tb_\G(D_i), W)}{\prod_{D_i \in D_{V_i}} q_{V}(D_i \mid \tb_\G(D_i), W)}.
	\end{align}
}
The second equality follows from the application of algebraic manipulations to the kernel $q_{D_{V_i}}(V_i \mid \mb_\G(V_i), W)$ in almost the exact same manner as the ones shown in the primal IPW formulation in the main draft (paragraph below Lemma~\ref{lem:primal_ipw}.) In fact, when $q_V(V \mid W)$ is defined to be the observed margin $p(V)$ of a hidden variable causal DAG, it is easy to see that the primal fixing operator recovers the primal IPW formula for $p(V(v_i))$.

Similar to the ordinary fixing operator $\phi$, the primal fixing operation can be applied to sequences of vertices $\sigma_S = ( S_1, \ldots, S_p)$ in a set $S$ provided this forms a valid p-fixing sequence for $S$ in $\G(V,W)$. The sequence is valid if $S_1$ is p-fixable in $\G,$ $S_2$ is p-fixable in $\primal_{S_1}(\G),$ and so on. Given a sequence $\sigma_S = (S_1, \ldots, S_p)$ p-fixable in $\G(V,W)$, define $\primal_{\sigma_S}(\G(V,W))$ as $\G$ if $S$ is empty, and $\primal_{\sigma_{S}\setminus S_1}(\primal_{S_1}(\G))$ otherwise.
Similarly, define $\primal_{\sigma_S}(q_V; \G)$ to be $q_V(V \mid W)$ if $S$ is empty and $\primal_{\sigma_{S}\setminus S_1} (\primal_{S_1}(q_V; \G); \primal_{S_1}(\G))$ otherwise.
We say $S$ is p-fixable in ${\cal G}(V,W)$ if there exists a valid sequence $\sigma_S$.
Since the graphical operator of p-fixing is equivalent to ordinary fixing, it follows trivially that two valid p-fixing sequences yield the same CADMG. The following lemma formalizes that any two valid p-fixing sequences also yield the same kernel.

\begin{lemma}[Commutativity of p-fixing]\ \\
	If $\sigma^1_S$ and $\sigma^2_S$ are both valid sequences for $S$ p-fixable in the ADMG $\G(V)$, then for any $p(V \cup H)$ Markov relative to a DAG $\G(V \cup H)$ that yields the latent projection $\G(V)$, $\primal_{\sigma^1_S}(p(V); \G(V)) = \primal_{\sigma^2_S}(p(V); \G(V)) = p(V \setminus S \mid \doo(S=s)),$ if $\G(V \cup H)$ is interpreted as a causal diagram.
	\label{lem:p-commute}
\end{lemma}
Due to this lemma, for marginals $p(V)$ induced by causal models associated with a hidden variable DAG $\G(V \cup H)$, if $S$ is p-fixable in the latent projection $\G(V)$, we can define $\primal_S(p(V); \G(V))$ to be the result of applying the fixing operator $\primal$ to any p-fixable sequence on $S$ and $p(V),\G(V)$. As suggested by the result, the proof of commutativity comes from viewing each valid p-fixing operation as a step in an identification procedure for the post intervention distribution $p(V \setminus S \mid \doo(S=s)).$ Thus,  any valid sequence for $S$ should yield the same post intervention distribution. This also suggests the following general procedure for identification of the target parameter via a sequence of p-fixing operations.

\begin{corollary}[Identification via a sequence of p-fixing]\ \\
	Fix a causal model associated with a hidden variable DAG $\G(V \cup H)$, that induces the observed marginal distribution $p(V)$. %nested Markov with respect to the latent projection $\G(V)$.
	Given $Y^* \equiv \an_{\G_{V\setminus T}}(Y)$, $\psi(t)$ is identified if there exists a subset $Z \subseteq V \setminus (Y^* \cup T )$ that is p-fixable in $\G(V)$ such that $T$ is p-fixable in $\primal_{Z}(\G(V)).$
	When $\psi(t)$ is identified in this manner we have, \label{lem:p-fix_sequence_id}
	{\small 
		\begin{align}
			\psi(t) = \sum_{V \setminus \{Z \cup T\}} Y \times \primal_{Z \cup T}(p(V); \G(V)) \bigg\vert_{T=t}.
			\label{eq:p-fix-rec-id}
		\end{align}
	}
	\label{lem:sequence-p-fixing}
\end{corollary} 

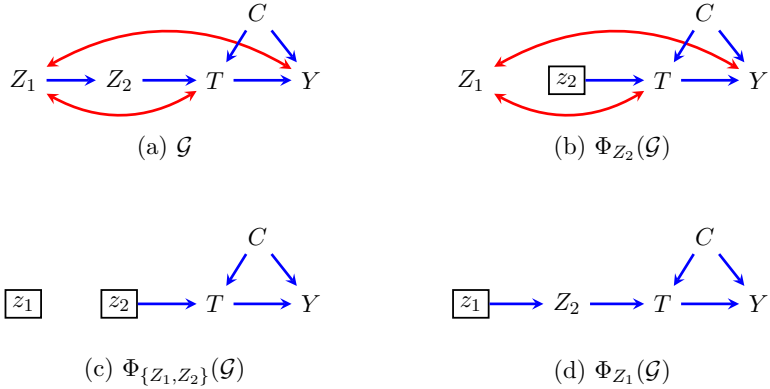
\begin{figure}[t]
	\begin{center}
		\scalebox{0.85}{
			\begin{tikzpicture}[>=stealth, node distance=1.5cm]
				\tikzstyle{format} = [thick, circle, minimum size=1.0mm, inner sep=0pt]
				\tikzstyle{square} = [draw, thick, minimum size=1.0mm, inner sep=3pt]
				
				\begin{scope}
					\path[->, very thick]
					node[] (c1) {$Z_1$}
					node[right of=c1] (c2) {$Z_2$}
					node[right of=c2] (t) {$T$}
					node[right of=t] (y) {$Y$}
					node[above right of=t, xshift=-0.4cm] (c) {$C$}
					(c1) edge[blue] (c2)
					(c2) edge[blue] (t)
					(t) edge[blue] (y)
					(c1) edge[<->, red, bend right] (t)
					(c1) edge[<->, red, bend left] (y)
					(c) edge[blue] (t)
					(c) edge[blue] (y)
					node[below right of=c2, xshift=-0.35cm, yshift=0cm] {(a) $\G$}
					;
				\end{scope}
				
				\begin{scope}[yshift=-3.5cm,xshift=7cm]
					\path[->, very thick]
					node[square] (c1) {$z_1$}
					node[right of=c1] (c2) {$Z_2$}
					node[right of=c2] (t) {$T$}
					node[right of=t] (y) {$Y$}
					node[above right of=t, xshift=-0.4cm] (c) {$C$}
					(c2) edge[blue] (t)
					(t) edge[blue] (y)
					(c1) edge[blue] (c2)
					(c) edge[blue] (t)
					(c) edge[blue] (y)
					node[below right of=c2, xshift=-0.35cm, yshift=0cm] {(d) $\primal_{Z_1}(\G)$}
					;
				\end{scope}
				
				\begin{scope} [yshift=0.0cm, xshift=7cm]
					\path[->, very thick]
					node[] (c1) {$Z_1$}
					node[square, right of=c1] (c2) {$z_2$}
					node[right of=c2] (t) {$T$}
					node[right of=t] (y) {$Y$}
					node[above right of=t, xshift=-0.4cm] (c) {$C$}
					(c2) edge[blue] (t)
					(t) edge[blue] (y)
					(c1) edge[<->, red, bend right] (t)
					(c1) edge[<->, red, bend left] (y)
					(c) edge[blue] (t)
					(c) edge[blue] (y)
					node[below right of=c2, xshift=-0.35cm, yshift=0cm] {(b) $\primal_{Z_2}(\G)$}
					;
				\end{scope}
				
				\begin{scope} [yshift=-3.5cm, xshift=0.0cm]
					\path[->, very thick]
					node[square] (c1) {$z_1$}
					node[square, right of=c1] (c2) {$z_2$}
					node[right of=c2] (t) {$T$}
					node[right of=t] (y) {$Y$}
					node[above right of=t, xshift=-0.4cm] (c) {$C$}
					(c2) edge[blue] (t)
					(t) edge[blue] (y)
					(c) edge[blue] (t)
					(c) edge[blue] (y)
					node[below right of=c2, xshift=-0.35cm, yshift=0cm] {(c) $\primal_{\{Z_1, Z_2\}}(\G)$}
					;
				\end{scope}
			\end{tikzpicture}
		}
	\end{center}
	\caption{ (a) An ADMG where $T$ is not fixable. 
		(b, c) A valid sequence of fixing that yields a CADMG  where $T$ is p-fixable and $p(Y(t))$ can be obtained as $p(Y(t))=p(Y(t, z_1, z_2)).$ 
		(d) A graph obtained after $Z_1$ is p-fixed, where $p(Y(t))$ can be obtained by p-fixing $T,$ as $p(Y(t)) = p(Y(t,z_1))$. }
	\label{fig:sequential_fixing}
\end{figure}

The result in Corollary~\ref{lem:sequence-p-fixing} directly yields an inverse weighted estimator. We illustrate this via the following example. 

\subsubsection*{Example: Identification via a sequence of p-fixing steps}

Consider the ADMG shown in Fig.~\ref{fig:sequential_fixing}(a). Clearly $T$ is not p-fixable as it does not meet the condition that $\dis_\G(T) \cap \ch_\G(T) = \emptyset.$ However, $Z_1$ is p-fixable, and yields a CADMG $\primal_{Z_1}(\G)$ where $T$ is p-fixable as shown in Fig.~\ref{fig:sequential_fixing}(d). Further, $Z_1$ is an ancestor of $Y$, but only via a directed path through $T$. Hence, while the CADMG in Fig.~\ref{fig:sequential_fixing}(d) corresponds to the post-intervention distribution $p(C, Z_2(z_1), T(z_1), Y(z_1)),$ fixing $T$ in this CADMG yields the post-intervention distribution $p(C, Z_2(z_1), Y(t))$ from which $p(Y(t))$ can be easily obtained as $Y(t,z_1)=Y(t)$ \citep{malinsky2019potential}.  A similar argument can be made to show that p-fixing according to the sequence $(Z_2, Z_1),$ resulting in the ADMGs shown in Figs.~\ref{fig:sequential_fixing}(b, c), also gives us the desired post-intervention distribution as $p(C, Y(t,z_1,z_2)) = p(C, Y(t)).$

Consider the first scenario where we p-fix $Z_1$ prior to p-fixing $T$ in the graph $\primal_{Z_1}(\G(V))$ shown in Fig.~\ref{fig:sequential_fixing}(d). We can estimate $\psi(t)$ via the following estimating equation: 

{\scriptsize
	\begin{align*}
		\mathbb{P}_n\bigg[p^*(Z_1) \times \frac{\sum_{Z_1} \ p(Y \mid T, Z, C) \times p(T \mid Z, C) \times  p(Z_1)}{p(Y \mid T, Z, C) \times p(T \mid Z, C) \times p(Z_1)}\times U\big(\psi(t); \primal_{Z_1}(p(V); \G(V))\big) \bigg] = 0,
	\end{align*}
}%
where  
{\small $U(\psi(t); \primal_{Z_1}(p(V); \G(V)))$} is given in Theorem~\ref{thm:IF_childless}, with nuisance models fitted with respect to the weighted distribution $p(V) / \pi_{Z_1}$, rather than with respect to $p(V)$.  In this example, the nuisances are simply the propensity score model for the treatment given covariates $C$ and the outcome regression model for $Y$ given the treatment and $C,$ as seen from the CADMG in Fig.~\ref{fig:sequential_fixing}(d). Further, $\pi_{Z_1}$ is defined as follows ({\small $Z = \{Z_1, Z_2\}$}).

{\scriptsize
	\begin{align*}
		\pi_{Z_1} =
		\frac{
			p(Y \mid T, Z, C) \times p(T \mid Z, C) \times p(Z_1)}
		{\sum_{Z_1} \ p(Y \mid T, Z, C) \times p(T \mid Z, C) \times p(Z_1)}.
	\end{align*}
}%

An estimation strategy is similar to the one used in marginal structural models \citep{robins2000marginal}. That is, we can fit a weighted regression for $\E[Y \mid T=t, C]$ using weights $1/\pi_{Z_1}$ estimated via appropriate nuisance models, a similarly weighted model for $p(A\mid C)$, and then plugging these in to solve the final estimating equation.

Using the alternative p-fixing sequence {\small $(Z_2, Z_1)$} also yields a CADMG $\primal_{Z_1, Z_2}(\G)$ where $T$ is p-fixable as in Fig.~\ref{fig:sequential_fixing}(c). In this case, $\pi_{Z_2}$ is simply {\small $p(Z_2 \mid Z_1).$} In the CADMG $\primal_{Z_2}(\G),$ the variable $Z_1$ is childless. Therefore, p-fixing $Z_1$ in the corresponding distribution corresponds to marginalization of $Z_1$ \citep{richardson2017nested}. Any p-fixings that correspond to marginalization in this manner, do not require the specification of an additional p-fixing weight. Hence, the estimating equation in this case is simply,

{\small
	\begin{align}
		\E\bigg[\frac{p^*(Z_1, Z_2)}{p(Z_2 \mid Z_1)} \times U\big(\psi(t); \primal_{Z_1, Z_2}(p(V); \G(V))\big) \bigg] = 0. \label{eq:reweight_fig7(a)}
	\end{align}
}
A similar strategy for estimation can be followed here  using the weights $1/\pi_{Z_2}$.

In cases where multiple p-fixing operations must be performed in some sequence, say $(Z_1, \dots, Z_k)$, the first set of weights $1/\pi_{Z_1}$ are  obtained by fitting appropriate nuisance models  using the observed data. Subsequent weights $1/\pi_{Z_i}$ for $1 < i \leq k$ are obtained by fitting weighted regressions that use a product of the prior  weights $1/(\pi_{Z_1} \times \dots \times \pi_{Z_{i-1}})$. The final nuisance models in {\small $U(\psi(t); \primal_{Z}(p(V); \G(V)))$} are then fit using weights $1/(\pi_{Z_1}\times \dots \times \pi_{Z_k}).$ The outer expectation is simply evaluated empirically.

An interesting special case is when the nuisance models for $\pi_{Z}$ can be estimated using pieces of the observed data likelihood that are variationally independent from the final nuisance models in $U(\cdot; \cdot)$. The resulting estimators in these cases exhibit double robustness \emph{after} correct specification of models involved in estimating $\pi_{Z}$. In the above example, when we use the order $(Z_2, Z_1)$ we only had to estimate weights $1/\pi_{Z_2}.$ Since $Z_2$ is in a different district from $Y$ and $T$, we obtain the desired variational independence for any natural parameterization of the observed data likelihood. However, generally (e.g., when we first p-fix $Z_1$ above) the exact form of robustness is an interesting question for future work.
%An implementation strategy that would allow us to estimate $\psi(t)$ using the above estimating equation is as follows. We first fit a model for the conditional density {\small $p(Z_2 \mid Z_1).$} We then use this model to predict inverse weights for each sample. We then fit the nuisance models $q_{CTY}(T \mid C)$ and $\E_{q_{CTY}}[Y \mid T=t, C]$ using logistic/linear regression or any flexible method that can be fit by utilizing the inverse weights in its objective function.  If additional weights $\pi_{Z_i}$ were needed, these models would be fit recursively using a product of the inverse weights from the previous stages. The final nuisance models in $U_{\psi_t}$ would then use  $1/\prod_{Z_i \in Z}\pi_{Z_i}$ as weights. Finally, the estimate for $\psi(t)$ is obtained by plugging in predictions from these models according to the estimating equation, which in the case of Fig.~\ref{fig:sequential_fixing}(c) reduces to the form of the AIPW estimator and empirically evaluating the expectation.

%%%%%%%%%%%%%%%%%%%%%%%%%%%%%%%%%%%%%%%%%%%%%
\newpage
\section{{Details on Simulated Data}} 
\label{app:sims}

The R code is available upon request.

\vspace{0.2cm}
\noindent To generate data from the ADMGs in Figures~\ref{fig:motiv}(b) and \ref{fig:p-fix_EIF}, we first generated six hidden variables that are used across the first three simulations: 
$H_1$ and $H_4$ are sampled from a Binomial distribution, with $p(H_1=1) = 0.4$ and $p(H_4=1) = 0.6$. 
$H_2$ and $H_5$ are sampled from a Uniform distribution with corresponding lower bounds $0, -1$ and upper bounds $1.5, 1$. 
$H_3$ and $H_6$ are sampled from a Normal distribution with means $0$ and standard deviations $1$ and $1.5$. For the observed variables in each simulation, continuous variables are  sampled from either a Normal or Uniform distributions and binary variables are sampled from Bernoulli distributions. In both ADMGs, we assume all variables, except for the outcome and part of the baselines, are binary random variables. In Fig.~\ref{fig:motiv}(b), we assume we have three baselines $C_1, C_2, C_3$ with Binomial ($p=0.3$), Uniform ($l=-1, u=2$), and Normal ($\mu=1, \text{sd}=1$) distributions. In Fig.~\ref{fig:p-fix_EIF}, we assume $C_1$ consists of $C_{11}$ and $C_{12}$ with Normal ($\mu=1, \text{sd}=1$) and Uniform $(l=-1, u=1)$ distributions, and $C_2$ consists of $C_{21}$ and $C_{22}$ with standard Normal and Binomial ($p = 0.4$) distributions. We projected out $Z_1$ and $Z_2$ from the DGP. 
Below we illustrate the data generating processes. $F_V(v)$ denotes the CDF of a standard normal distribution. 

\vspace{-0.2cm}
{\scriptsize
	\begin{align*}
		C_4 &= F_{C_3}(c_3), \ C_5 = C_3^{C_1} + (1-C_1)\times\sin(|C_3|\pi), \ C_6 = C_1\times C_2 + |C_3|  
		\hspace{2.3cm} 	\text{(Fig.~\ref{fig:motiv}(b))} 
		\\
		p(T = 1 \mid C,U) &\sim \text{expit}(0.5 + 0.9C_4 -0.5C_5 + 0.2C_6 + 0.3U_1 - 0.8U_2 + 0.8U_3)
		\\ 
		p(M = 1 \mid C, T, U) &\sim \text{expit}(0.5 - 0.7C_1 + 0.8C_2 - C_3 - 1.2T - 0.2U_4 + 0.5U_5 + 0.4U_6
		+ (1.5C_4 + 1.2C_5+ 0.6C_6)T)
		\\
		p(L = 1 \mid C, M, U) &\sim \text{expit}(-0.5 + 0.8C_4 + 1.2C_5 - 0.6C_6 - 1.2M + 0.3U_1 + 0.6U_2 - 0.4U_3
		- (0.8C_4 + 1.5C_5 + 0.4C_6)M) 
		\\
		Y \mid C, T, L, U  &\sim 0.5 + 0.5C_4 - 2C_5 + 0.8C_6 + 0.5T + 0.6L - 0.6U_4 + 0.5U_5 -0.5U_6 + 1.3C_4A +  \\ 
		&\hspace{0.5cm} 2.3C_5L + 2C_6TL + 1.2AL + \mathbb{N}(0, 1.5).
	\end{align*}
}%
\vspace{-0.75cm}
{\scriptsize
	\begin{align*}
		C_3 &= F_{C_{11}}(c_{11}c_{12}) + (1-C_{12})\times \sin(|C_{11}|\pi), \ C_4 = (C_{21}^{C_{22}}) + (1-C_{22})\times\sin(|C_{21}|\pi)) 
		\hspace{1.25cm} 	\text{(Fig.~\ref{fig:p-fix_EIF})} 
		\\
		p(T = 1 \mid C,U) &\sim \text{expit}(-0.5 + 0.9C_{11} -0.7C_{12}+ 0.6C_{21} - 0.7C_{22} + 0.3U_1 - 0.5U_2 + 0.4U_3 
		+ 1.6C_3 - 0.8C_4 )
		\\ 
		p(M = 1 \mid C, T) &\sim \text{expit}(-0.5 - 1.4C_{21} + 1.3C_{22}- 1.2A + 2.2C_4A -  C_4)
		\\
		p(L = 1 \mid C, M, U) &\sim \text{expit}(0.5 - 0.5*C_{11} - 0.4C_{12} + 0.8C_{21} +  0.9C_{22} - 1.2M + 0.3U_1 + 0.6U_2 - 0.4U_3  \\
		&\hspace{1.2cm} 	- 1.8C_{3}M - 1.5C_4M + 1.2C_{3} + 0.8C_4) 
		\\
		Y \mid C, L  &\sim  0.5 + 0.7C_{21} - 0.5C_{22} + 1.6L +  1.1C_{4}L + 0.8C_{4}  + \mathbb{N}(0, 1.5).
	\end{align*}
}%

\noindent To generate data from the ADMG in Figure~\ref{fig:general_id}, we first generated ten hidden variables that are used for the last simulation: 
$H_1, H_3, H_5, H_7, H_9$ are sampled from a Binomial distribution with $p_1=0.4, p_3=0.3, p_5=0.4, p_7=0.3, p_9=0.3$. 
$H_2, H_4, H_6, H_8, H_{10}$ are sampled from standard Normal distribution. We assume $C$ consists of two baseline covariates. All variables, except for the baselines and outcome, are binary random variables. 

\vspace{-0.2cm}
{\scriptsize
	\begin{align*}
		p(R_2 = 1 \mid U) &\sim \text{expit}(-0.2 + 0.3U_1 - 0.8U_2 + 0.4U_3 + 0.6U_4 ) 
		\hspace{5.5cm} 	\text{(Fig.~\ref{fig:general_id})}  
		\\
		C_1 \mid U &\sim 0.4U_7 - 0.1U_8 + 0.6U_9 + 0.8U_{10} + \mathbb{N}(0, 1) 
		\\
		C_2 \mid U &\sim -0.3U_7 - 0.7U_8 + 0.8U_9 + 1.2U_{10} + \mathbb{N}(0, 1) 
		\\
		C_3 &= {|C_1C_2|^{0.5}} + \sin(|C_1+C_2|\pi), \ C_4 =  F_{C_1}(c_{1}),
		\\
		p(Z = 1 \mid U) &\sim \text{expit}(-0.5 + U_1 + 0.2U_2 - 0.8U_5 + 0.3U_6)
		\\ 
		p(T = 1 \mid C, Z, U) &\sim \text{expit}(0.5 - 0.5C_1 + 0.5C_2 + 0.3Z + 0.5U_3 - 0.4U_4 + 0.8C_3 - 1.3C_4)
		\\ 
		p(R_1 = 1 \mid T, U) &\sim \text{expit}(0.2 + 0.7T - 0.6U_5 - 0.6U_6)
		\\ 
		p(M = 1 \mid R_1, U) &\sim \text{expit}(0.5 - 0.8R_1 + 1.2U_7 -1.5U_8)
		\\
		Y \mid R_2, C, T, M, U  &\sim   -1 + 0.5C_1 + 0.2C_2 + 1.2T + 0.8R_2 + 0.8M + 0.2U_9 - 0.4U_{10} + 0.8C_3 - 1.2C_4 + MA + \mathbb{N}(0, 1).
	\end{align*}
}%

%+++++++++++++++++++++++++++++++++++++++++

\begin{figure}[!t]
	\begin{subfigure}[b]{0.5\textwidth}
		\centering
		\includegraphics[scale=0.19]{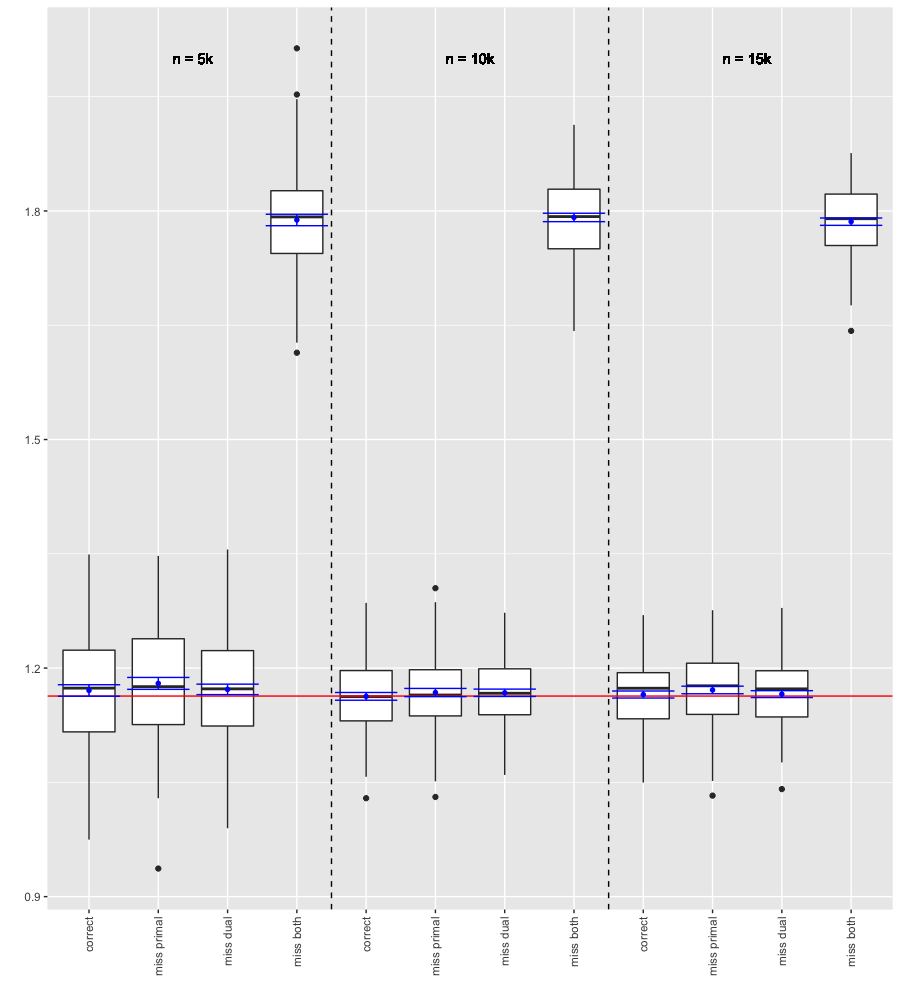}
		%		\label{fig:DR-ADMG-2b}
	\end{subfigure}
	\begin{subfigure}[b]{0.5\textwidth}
		\centering
		\includegraphics[scale=0.19]{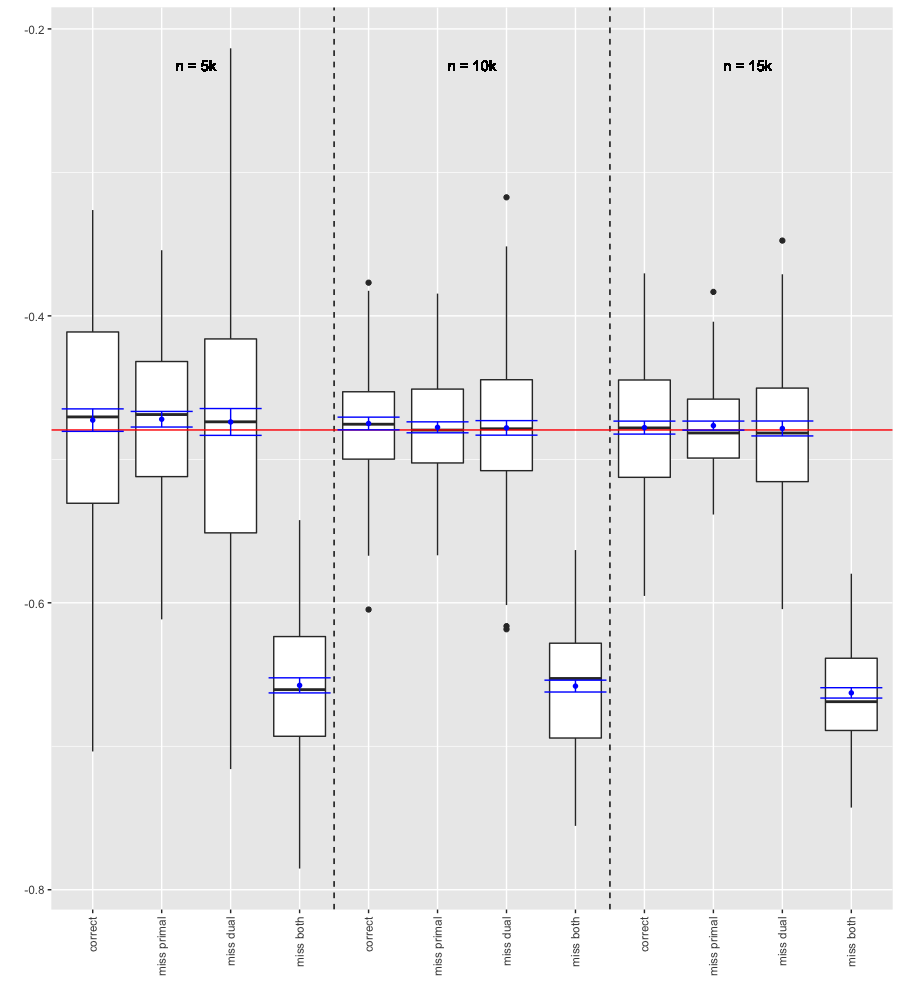}
		%		\label{fig:DR-ADMG-3}
	\end{subfigure}
	\caption{Double robustness of the Augmented Primal IPW estimator. The boxplot panel on the (left) uses the ADMG in Fig.~\ref{fig:motiv}(b), and the one on the (right) uses the ADMG in Fig.~\ref{fig:p-fix_EIF}. The red dashed lines indicate the true values of the ACE.  The blue dots denote the mean and the blue error bars denote the length of the standard error of the mean. }
	\label{fig:DR-ADMGs-supp}
\end{figure}

{\bf Additional Experiments}: In continuation of Simulation 2, in Figure~\ref{fig:DR-ADMGs-supp} we highlight  the summary statistics for the case of using Augmented Primal IPW estimator for computing the causal effect of $T$ on $Y$ in ADMGs of Figures~\ref{fig:motiv}(b) and \ref{fig:p-fix_EIF}. 

%+++++++++++++++++++++++++++++++++++++++++

\begin{figure}[!h]
	
	\begin{subfigure}[b]{1\textwidth}
		\centering
		\includegraphics[scale=0.33]{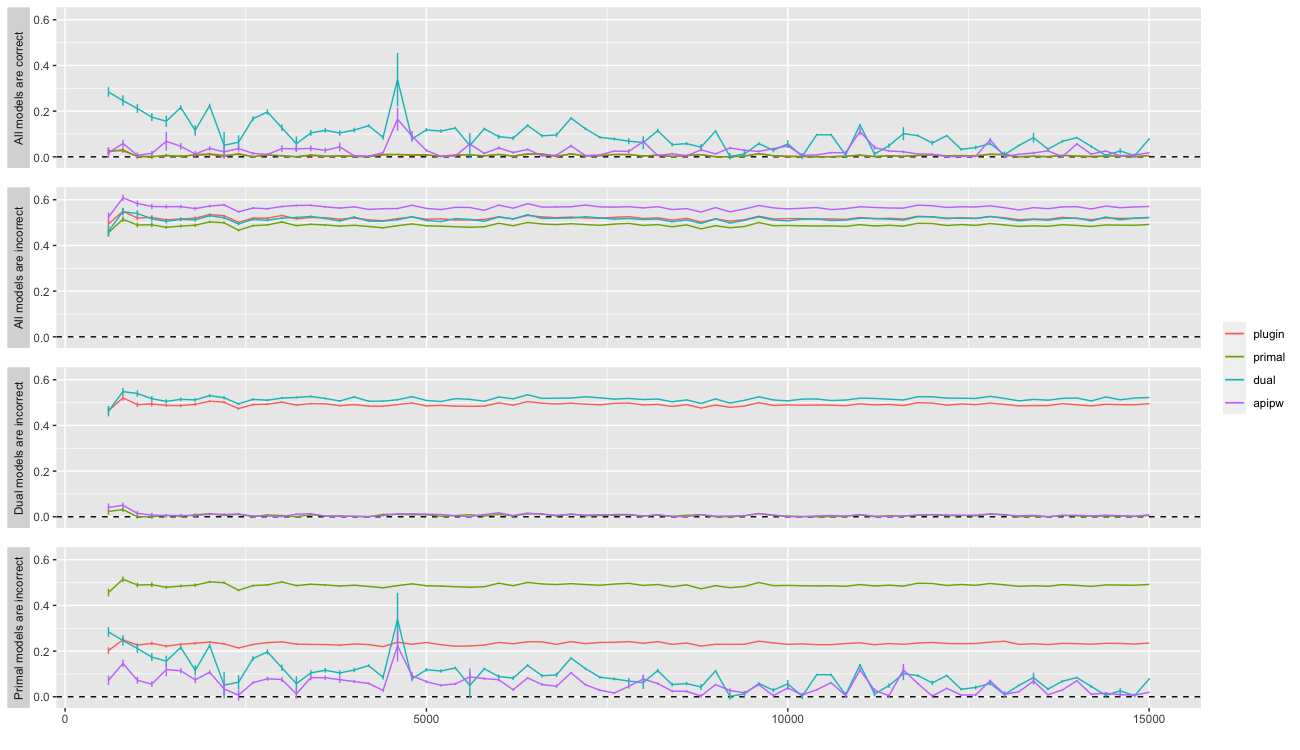}
	\end{subfigure}
	\newline 
	\newline
	\begin{subfigure}[b]{1\textwidth}
		\centering
		\includegraphics[scale=0.33]{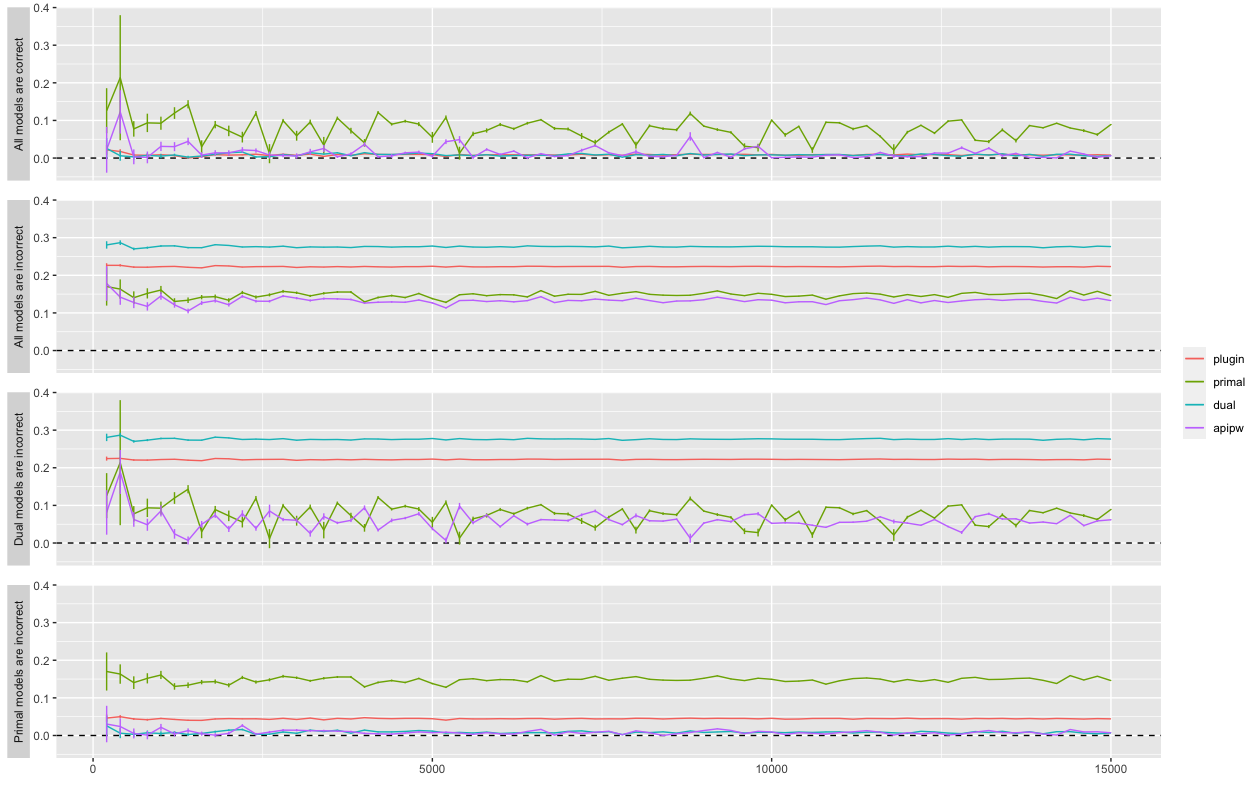}
	\end{subfigure}
	
	\caption{Bias behavior as a function of sample size with an alternative set of parameters for the DGPs that can result in extreme weights in the IPW estimators. (top) using the ADMG in Fig.~\ref{fig:motiv}(b). (bottom) using the ADMG in Fig.~\ref{fig:p-fix_EIF}. }
	\label{fig:bias-ADMG-supp}
\end{figure}

\begin{figure}[!t]
	
	\begin{subfigure}[b]{1\textwidth}
		\centering
		\includegraphics[scale=0.33]{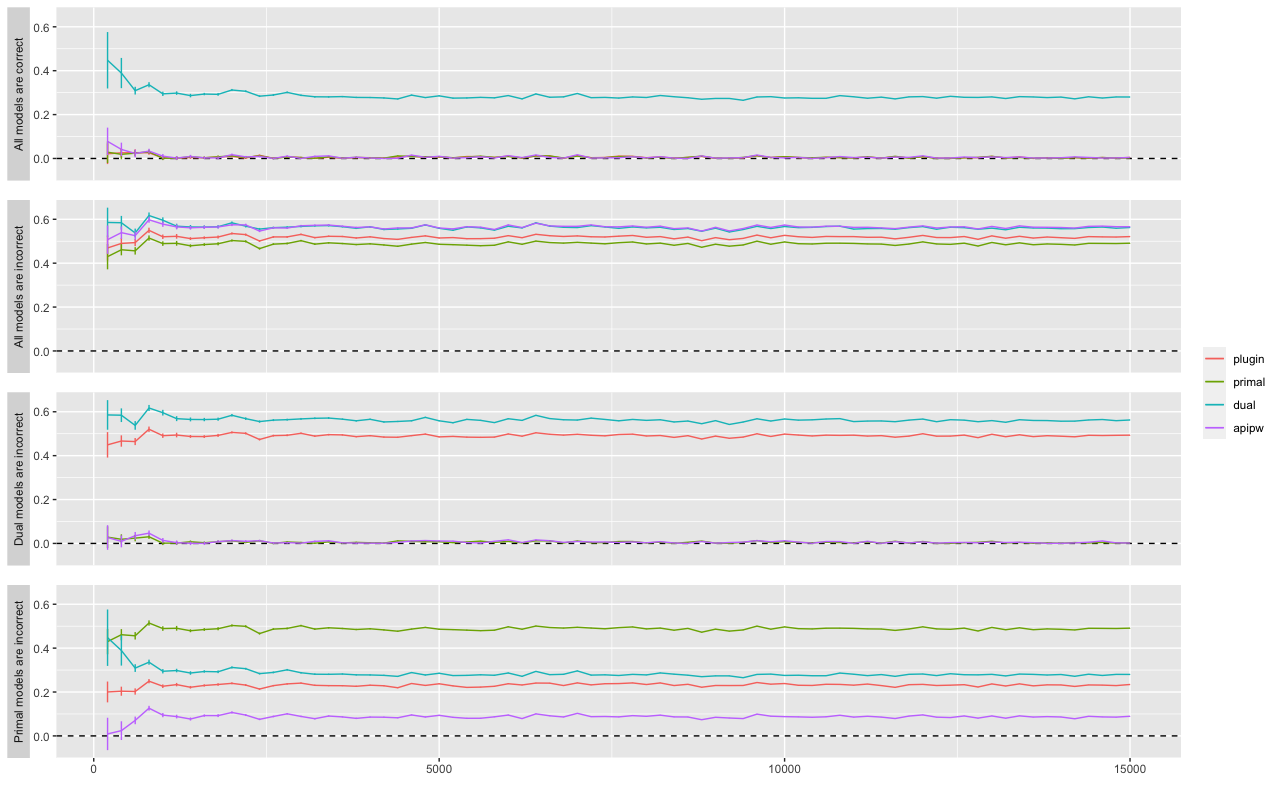}
	\end{subfigure}
	\newline 
	\newline
	\begin{subfigure}[b]{1\textwidth}
		\centering
		\includegraphics[scale=0.33]{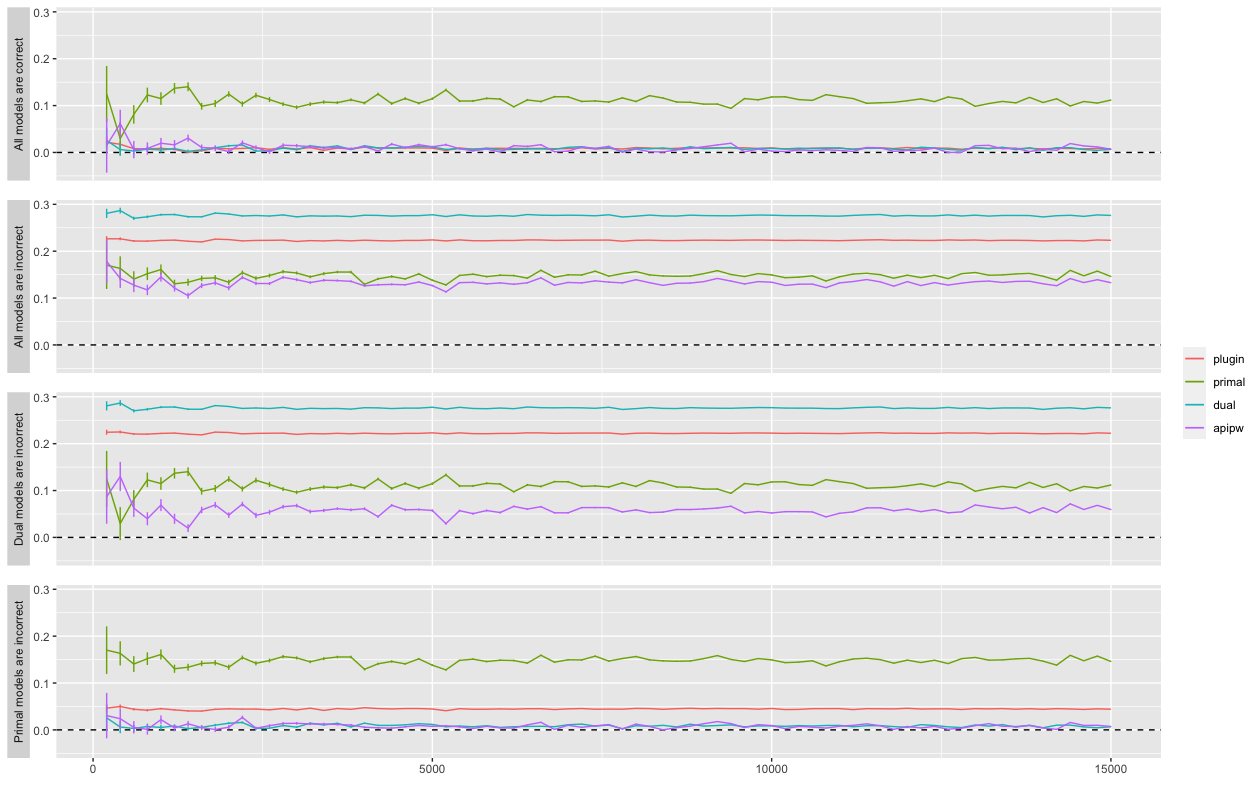}
	\end{subfigure}
	
	\caption{Bias behavior as a function of sample size with truncated weights in Fig~\ref{fig:bias-ADMG-supp}. (top) using the ADMG in Fig.~\ref{fig:motiv}(b). (bottom) using the ADMG in Fig.~\ref{fig:p-fix_EIF}. The plots have fewer outliers. However, truncating weights in primal and dual terms introduces bias that does not disappear as sample size increases. }
	\label{fig:cutoff-ADMG-supp}
\end{figure}

We reran the simulations for primal fixability with various different DGPs (different parameter sets.) Our conclusions remain consistent across all simulations. It is worth pointing out that in general, IPW-type estimators are instable due to extreme values for the weights used in the estimation procedures. The followings are examples of two DGPs (using the ADMGs in Figures~\ref{fig:motiv}(b) and \ref{fig:p-fix_EIF}, where we encountered some instability in the weights and consequently in how primal and dual IPW estimators behave as sample size increases. We plot the bias against the sample size in Figure~\ref{fig:bias-ADMG-supp}.  We truncated the extreme weights in both examples and were able to observe a more smoothed behavior, however, such procedures introduce bias that does not disappear as sample size increases. This is shown in Figure~\ref{fig:cutoff-ADMG-supp}. Exploring  possibilities  for dealing with such instabilities is an interesting future direction. 

\vspace{-0.25cm}
{\scriptsize
	\begin{align*}
		C_4 &= F_{C_3}(c_3), \ C_5 = C_3^{C_1} + (1-C_1)\times\sin(|C_3|\pi), \ C_6 = C_1\times C_2 + |C_3|  
		\hspace{2.3cm} 	\text{(Fig.~\ref{fig:motiv}(b))} 
		\\
		p(T = 1 \mid C,U) &\sim \text{expit}(0.5 + 0.9C_4 -0.5C_5 + 0.2C_6 + 0.3U_1 - 0.8U_2 + 0.8U_3)
		\\ 
		p(M = 1 \mid C, T, U) &\sim \text{expit}(0.5 - 0.7C_1 + 0.8C_2 - C_3 - 1.2T - 0.2U_4 + 0.5U_5 + 0.4U_6
		+ (1.5C_4 + 1.2C_5+ 1.6C_6)T)
		\\
		p(L = 1 \mid C, M, U) &\sim \text{expit}(-0.5 + 0.8C_4 + 1.2C_5 - 0.6C_6 - 1.2M + 0.3U_1 + 0.6U_2 - 0.4U_3
		- (0.8C_4 + 1.5C_5 + 0.4C_6)M) 
		\\
		Y \mid C, T, L, U  &\sim 0.5 + 0.5C_4 - 2C_5 + 0.8C_6 + 0.5T + 0.6L - 0.6U_4 + 0.5U_5 -0.5U_6 + 1.3C_4A +  \\ 
		&\hspace{0.5cm} 2.3C_5L + 2C_6TL - 1.2AL + \mathbb{N}(0, 1.5).
	\end{align*}
}%
\vspace{-0.75cm}
{\scriptsize
	\begin{align*}
		C_3 &= F_{C_{11}}(c_{11}), \ C_4 = (C_{21}^{C_{22}}) + (1-C_{21})\times\sin(|C_{21}|\pi)) 
		\hspace{4.75cm} 	\text{(Fig.~\ref{fig:p-fix_EIF})} 
		\\
		p(T = 1 \mid C,U) &\sim \text{expit}(0.5 + 0.9C_{11} -0.5C_{12}+ 0.2C_{21} + 1.2C_{22} + 0.3U_1 - 0.8U_2 + 0.8U_3 
		+ 0.7C_3 - 2C_4 )
		\\ 
		p(M = 1 \mid C, T) &\sim \text{expit}(0.5 - 0.7C_3 + 0.8C_4 - 1.2A + 1.5C_3A + 1.2C_4A)
		\\
		p(L = 1 \mid C, M, U) &\sim \text{expit}(-0.5 - 0.5C_3 - 0.1C_{12} + 0.8C_{21} + 1.2C_{22} + 0.3U_1 + 0.6U_2 - 0.4U_3 - (1.2 + 0.8C_{21} + 1.5C_{22})M) \\
		%				&\hspace{1.2cm} 	- 0.8C_{21}M - 1.5C_{22}M) 
		%				\\
		Y \mid C, L  &\sim  0.5 + 0.5C_{21} - 2C_{22} + 0.6L +  1.3C_{21}L + 2.3C_{22}L  + \mathbb{N}(0, 1.5).
	\end{align*}
}%

%+++++++++++++++++++++++++++++++++++++++++

%+++++++++++++++++++++++++++++++++++++++++

%%%%%%%%%%%%%%%%%%%%%%%%%%%%%%%%%%%%%%%%%%%%%
%%%%%%%%%%%%%%%%%%%%%%%%%%%%%%%%%%%%%%%%%%%%%
%%%%%%%%%%%%%%%%%%%%%%%%%%%%%%%%%%%%%%%%%%%%%

\clearpage
\section{Proofs}
\label{app:proofs}

%%%%%%%%%%%%%%%%%%%%%%%%%%%%%%%%%%%%%%%%%%%%%

\subsection*{Theorem~\ref{thm:nps} (Soundness and completeness of Algorithm~\ref{alg:nps})}
\begin{proof}
	
	The construction of Algorithm~\ref{alg:nps} is closely related to the maximal arid projection described by \cite{shpitser2018acyclic}. Given any ADMG $\G$ the maximal arid projection yields a maximal arid graph (MArG) $\G^a$ that is nested Markov equivalent to $\G.$ The MArG $\G^a$ has
	\begin{enumerate}
		\item A directed edge $V_j \diedgeright V_i$ if and only if $V_j\in \an_\G(V_i)$ and $V_j$ is a parent of any element in the \emph{reachable closure} of $V_i$ -- the set of vertices that are still random in $\phi_{\neg \{V_i\}}(\G).$
		\item A bidirected edge $V_i \biedge V_j$ if and only if the previous condition fails, and $\phi_{\neg \{V_i, V_j\}}(\G)$ contains a single district.
	\end{enumerate}
	
	We now prove that Algorithm~\ref{alg:nps} declares $\G$ to be NPS if and only if $\G^a$ is a complete graph (no missing edges). We will then use this observation to reason about the soundness and completeness of the algorithm. Notice that for any pair $(V_i, V_j)$ and assuming wlog that $V_j \prec_\tau V_i,$ the first check in line~\ref{alg:nps-check} of the algorithm evaluates to True when $V_j$ is not a parent of the reachable closure of $V_i,$ as the district of $V_i$ in $\phi_{\neg \{V_i\}}(\G)$ is the same as the reachable closure of $V_i$ (elements of the district of $V_i$ are the only vertices that can remain random in such a CADMG.) That is, this condition evaluates to True when $V_j \diedgeright V_i$ is not added to  $\G^a$ fails. It is also easy to see that the second check in line~\ref{alg:nps-check} evaluates to True when  $V_i \biedge V_j$ is not added to $\G^a.$ Hence, when both checks evaluate to True the algorithm returns ``not NPS'' and the MArG $\G^a$ has at least one pair $(V_i, V_j)$ that are not adjacent. On the other hand, when the truth value output in line~\ref{alg:nps-check} evaluates to False for every pair $(V_i, V_j)$ resulting in the algorithm returning ``NPS'', it is clear that either $V_j$ is a parent of the reachable closure of $V_i,$ or if not, $\phi_{\neg \{V_i, V_j\}}(\G)$ contains a single district. That is, $\G^a$ either contains $V_j \diedgeright V_i$ or $V_j \biedge V_i$ for every pair  when the algorithm returns ``NPS.''
	
	\subsubsection*{Completeness}
	
	We now prove that the algorithm is complete -- when the algorithm declares the model of $\G(V)$ to be NPS it is indeed NPS. We have seen that when the algorithm returns NPS, the input ADMG $\G$ is nested Markov equivalent to a MArG $\G^a$ where every pair of vertices is pairwise connected with a directed or bidirected edge. The nested Markov model of any complete ADMG is NPS (per Corollary~\ref{cor:admg-complete}), and so it is indeed the case that $\G$ is NPS, and the output of the algorithm is correct.
	
	\subsubsection*{Soundness}
	The algorithm returns ``not NPS'' when the input ADMG $\G$ is nested Markov equivalent to a MArG $\G^a$ where at least one pair of vertices, say $(V_i, V_j)$, are not connected by a directed or bidirected edge. Assume wlog that $V_i \prec V_j$ according to a valid topological order, and $(V_i, V_j)$ is the only non-adjacent pair in $\G^a.$ We assume the edge between $V_i$ and $V_j$ is the only missing edge for simplicity, but it is easy to see  if there are additional missing edges this simply corresponds to a submodel where there is still a constraint between $V_i$ and $V_j$.
	
	We know the complete graph where the edge is present between $V_i$ and $V_j$ is NPS (no constraints in the nested Markov model.)
	We now ask whether the constraint finding algorithm in \cite{tian2002testable}, which returns a list of constraints that imply the nested Markov model of an ADMG, detects a constraint between $V_i$ and $V_j$ in $\G$, implying it is not NPS.
	
	The algorithm in \cite{tian2002testable}, and reformulated in terms of kernels and fixing in \cite{richardson2017nested}, starts by examining the subgraph of $\G$  consisting of $V_j$ and all vertices preceding $V_j$ under a valid topological order, namely $\G_{\{\preceq_\tau V_j\}}$.
	Let $S$ be the district of $\G_{\{\preceq_\tau V_j\}}$
	containing $V_j$.  If $V_i$ is not in the district $S$ nor a parent of the district $S$, then we immediately get an ordinary conditional independence constraint $V_i \ci V_j \mid \tb_{\G}(V_j)$ in step (A1) of the algorithm. We now examine the recursion that occurs in step (A2) to handle the remaining cases: (i) $V_i$ is in the district $S,$ but not a parent of $S,$ and (ii) $V_i$ is a parent of $S$ and potentially in the district $S$.
	Recursive applications of step (A2) in the algorithm involve examining CADMGs 
	with random vertices that are subsets of $S$, and corresponding kernels obtained from $p(V)$, obtained by one of two steps.  Given a recursively obtained CADMG
	${\cal G}(\tilde{S},W)$, the first step considers subgraphs $\G_A$ of ${\cal G}(\tilde{S},W)$ consisting of all ancestral subsets $A$ of $\tilde{S}\cup W$ in ${\cal G}(\tilde{S},W)$ that contains $V_j$.\footnote{An ancestral margin of an CADMG $\G$ is a subgraph with a vertices that is closed under the ancestral relation in $\G$.}  The second step considers a district $E$ containing $V_j$ within ${\cal G}_A$.  These steps, occuring within recursive applications of (A2), visit every intrinsic set in the original ADMG ${\cal G}(V)$, potentially multiple times.
	
	Case (i): $V_i$ and $V_j$ belong to the same district in $\G_{\{\preceq_\tau V_j\}}$.
	Consider a sequence of alternating ancestral and district steps within (A2), starting from $S$, which always include $\{ V_i, V_j \}$ in the random vertex set $\tilde{S}$ for the CADMG $\G(\tilde{S},\tilde{W})$ associated with each step.  This sequence is non-empty, since $V_i$ and $V_j$ are in $S$, by assumption.  Moreover, $V_i$ and $V_j$ will always be childless in each CADMG ${\cal G}(\tilde{S},\tilde{W})$.
	Since the projection $\G^a$ is a MArG where $V_i$ and $V_j$ are not adjacent, we know $\phi_{\neg{\{V_i, V_j\}}}(\G_S) = \phi_{\neg\{ V_i, V_j \}}(\G)$ is a CADMG where $V_i$ and $V_j$ are not bidirected connected.
	Thus, there must exist a point in this sequence where $V_i$ and $V_j$ will no longer be in the same district after an ancestral step applied to some $\G(\tilde{S},\tilde{W})$, where an ancestral set $A$ is retained.  Since $V_i$ is childless in the resulting CADMG,
	$V_i$ is not in the Markov blanket of $V_j$, and the algorithm in \cite{tian2002testable} adds the corresponding constraint, namely that the kernel $q(\tilde{E} \mid \pa_{\G}(\tilde{E}))$
	is not a function of $V_i$, where $\tilde{E}$ is the district containing $V_j$ in $\G(\tilde{S},\tilde{W})_A$, and
	$q(\tilde{E} \mid \pa_{\G}(\tilde{E}))$ is the corresponding kernel obtained from
	$\sum_{\tilde{S}\setminus A} q(\tilde{S} \mid \pa_{\G}(\tilde{S}))$. 
	
	Case (ii): $V_i$ is a parent of the district $S$ and potentially also a part of the district $S$ in $\G_{\{\preceq_\tau V_j\}}$.
	Consider a sequence of alternating ancestral and district steps within (A2), starting from $S$, which always include $V_j \in \tilde{S}$ and $V_i \in \tilde{S} \cup \tilde{W}$ for the CADMG $\G(\tilde{S},\tilde{W})$ associated with each step.
	This sequence is non-empty, since $V_i$ is in $\pa_{\G}(S)$.
	Since the projection $\G^a$ is a MArG where $V_i$ and $V_j$ are not adjacent, we know $\phi_{\neg{\{V_j\}}}(\G)$ is a CADMG where $V_i$ is not a parent of the district of $V_j$.
	Thus, there must exist a point in this sequence where after either an ancestral or district step,
	$V_i$ will no longer be a parent of the district of $V_j$ in the CADMG $\G(\tilde{S},\tilde{W})_A$ (if an ancestral step was taken) or
	$\G(\tilde{E},\tilde{W})_{\tilde{E}}$ (if the district step was taken).
	At this point, the algorithm in \cite{tian2002testable} adds the corresponding constraint, namely that the kernel $\sum_{\tilde{S} \setminus A} q(\tilde{S} \mid \pa_{\G}(\tilde{S}))$ (if an ancestral step was taken) or $q(\tilde{E} \mid \pa_{\G}(\tilde{E}))$ (if a district step was taken)
	is not a function of $V_i$.
	
	Hence, we see that in either case, the algorithm gives a non-empty list of constraints associated with $\G$ whenever 	there is an edge missing between $V_i$ and $V_j$ in the MArG projection $\G^a$.
\end{proof}

%%%%%%%%%%%%%%%%%%%%%%%%%%%%%%%%%%%%%%%%%%%%%

\subsection*{Theorem~\ref{thm:mb-shielded} (mb-shielded ADMGs) }

\begin{proof} The proof relies on the fact that the constraint finding algorithm provided in \cite{tian2002testable} finds a list of equality constraints that is sufficient to define the nested Markov model of an ADMG (this was shown by \cite{richardson2017nested}). Here we show that the only non-trivial equality constraints found by applying this algorithm to an arbitrary mb-shielded ADMG $\G$ are of the form $V_i \ci \{\prec V_i \} \mid \tb_\G(V_i)$, for a topological order $\prec$, thus implying that all equality constraints in the nested Markov model of such an ADMG are implied by ordinary conditional independences of that form.
	
	Given a topological order $\prec$ on the vertices in an ADMG ${\cal G}(V)$, the constraint finding algorithm in \cite{tian2002testable} iterates over each vertex $V_i$ in the order and attempts to find constraints between $V_i$ and $\{\prec V_i\}.$ In substep (A1) of the algorithm (see \cite{tian2002testable} for details), it identifies constraints of the form $V_i \ci \{\prec V_i \} \mid \tb_\G(V_i)$.  Step (A2) and recursive applications of it, attempts to find constraints between $V_i$ and subsets of $\{\prec V_i\} \cap \mb_\G(V_i)$. In the rest of this proof, we temporarily switch to using a disjunctive definition of parents, i.e., $\pa_\G(S) = \bigcup_{S_i \in S} \pa_\G(S_i),$ to align with the presentation of the constraint finding algorithm in \cite{tian2002testable}.
	
	The top level calls to step (A2) always involve a subset $S$ of a district $D$ of ${\cal G}(V)$, along with $\pa_{{\cal G}(V)}(S) \setminus S$.  
	These subsets $S$ are districts of $V_i$ in subgraphs ${\cal G}_{\{ \prec V_i \}}$ of ${\cal G}(V)$.  Each call to (A2) involves constraints on a particular $V_i$.
	Since ${\cal G}(V)$ is mb-shielded, the induced subgraph ${\cal G}(V)_{S \cup \pa_{{\cal G}(V)}(S)}$ is complete.  In particular,
	since every $V_j \in \pa_{{\cal G}(V)}(S) \setminus S$ is in the Markov blanket of every $V_k \in S$, $V_j \in \pa_{{\cal G}(V)}(V_k)$.
	The top level calls to step (A2) proceed as follows.
	
	First, a subset $\tilde{S}$ of $S$ ancestral in ${\cal G}(V)_{S \cup \pa_{{\cal G}(V)}}(S)$ is found.  A constraint is found if
	$\{ [\pa_{{\cal G}(V)}(S) \cup S] \setminus (S \setminus \tilde{S}) \} \setminus [\pa_{{\cal G}(V)}(\tilde{S}) \cup \tilde{S}] \neq \emptyset$. Note that $\{ [\pa_{{\cal G}(V)}(S) \cup S] \setminus (S \setminus \tilde{S}) \} = ([\pa_{{\cal G}(V)}(S)\cup S] \cap \tilde{S}) \cup (\pa_{{\cal G}(V)}(S) \setminus S)$.
	Since ${\cal G}(V)$ is mb-shielded, and every element $V_j \in \pa_{{\cal G}(V)}(S) \setminus S$ is in the Markov blanket of every element in $S$ (and thus in $\tilde{S}$) in ${\cal G}_{\{ V_i \}}$.
	Thus, every element $V_j \in \pa_{{\cal G}(V)}(S) \setminus S$ is in $\pa_{{\cal G}(V)}(V_k)$ for every $V_k \in \tilde{S}$.  Any element $V_j$ in $[\pa_{{\cal G}(V)}(S)\cup S] \cap\tilde{S}$ must be in
	$\pa_{{\cal G}(V)}(\tilde{S}) \cup \tilde{S}$ by definition.  Consequently, $\{ [\pa_{{\cal G}(V)}(S) \cup S] \setminus (S \setminus \tilde{S}) \} \setminus [\pa_{{\cal G}(V)}(\tilde{S}) \cup \tilde{S}] = \emptyset$, and no restrictions are added at this stage.
	
	Second, the algorithm considers a district $\tilde{E}$ of $V_i$ in ${\cal G}_{\tilde{S} \cup \pa_{{\cal G}(V)}(\tilde{S})}$,
	and $\pa_{{\cal G}(V)}(\tilde{E})$.  A constraint is found if $[\pa_{{\cal G}(V)}(\tilde{S}) \cup \tilde{S}] \setminus [\pa_{{\cal G}(V)}(\tilde{E}) \cup \tilde{E}] \neq \emptyset$.
	Any element $V_j$ in $\pa_{{\cal G}(V)}(\tilde{S}) \setminus \tilde{S}$ must be in $\pa_{\cal G}(S) \setminus S$, since $\tilde{S}$ is ancestral in ${\cal G}(V)_{S \cup \pa_{{\cal G}(V)}}(S)$.
	Consequently $V_j \in \pa_{{\cal G}(V)}(V_k)$ for every $V_k \in \tilde{E} \subseteq S$.
	
	Any element $V_j \in \tilde{S}$ is in the Markov blanket of every element in $\tilde{E}$, in particular $V_i$.  Thus, $V_j$ and $V_i$ must share an edge.  Since $V_i$ is the $\prec$-largest element in $S$ under a topological order, this edge must be $V_j \diedgeright V_i$ or $V_j \biedge V_i$.  In the former case, $V_j \in \pa_{{\cal G}(V)}(\tilde{E})$, while the latter case, $V_j \in \tilde{E}$.
	In either case, $[\pa_{{\cal G}(V)}(\tilde{S}) \cup \tilde{S}] \setminus [\pa_{{\cal G}(V)}(\tilde{E}) \cup \tilde{E}] = \emptyset$, and no restrictions are added at this stage.
	
	Step (A2) is then recursively applied to ${\cal G}(V)_{\tilde{E} \cup \pa_{{\cal G}(V)}(\tilde{E})}$.  Since for any $V_j,V_k \in \tilde{E} \cup \pa_{{\cal G}(V)}(\tilde{E})$, one of the pair is in the Markov blanket of the other, they must share an edge in ${\cal G}(V)_{\tilde{E} \cup \pa_{{\cal G}(V)}(\tilde{E})}$.  Moreover, any element in $\pa_{{\cal G}(V)}(\tilde{E}) \setminus \tilde{E}$ must be connected to any elements in $\tilde{E}$ by a directed edge.  This allows us to repeat the above argument inductively, with $\tilde{E}$ replacing $S$ and $\pa_{{\cal G}(V)}(\tilde{E}) \setminus \tilde{E}$ replacing $\pa_{{\cal G}(V)}(S) \setminus S$, to establish that no constraints are added on any recursive call in step (A2).
	
	Thus, running the algorithm on an mb-shielded ADMG returns a list of constraints consisting of only ordinary conditional independence constraints (those that are found by substep (A1)), and specifically ones that are of the form $V_i \ci \{\prec V_i \} \mid \tb_\G(V_i)$.
\end{proof}

\begin{corollary}[Complete ADMGs are nonparametric saturated]\ \\
	Any complete ADMG $\G(V)$ is mb-shielded, and observationally equivalent to a complete DAG. Hence, the nested Markov model of a complete ADMG is nonparametric saturated.
	\label{cor:admg-complete}
\end{corollary}
\begin{proof}
	The first claim follows from the definition of mb-shieldedness, and by application of Theorem~\ref{thm:mb-shielded}. Since complete DAGs are nonparametric saturated, so is the ADMG $\G(V).$
\end{proof}

%%%%%%%%%%%%%%%%%%%%%%%%%%%%%%%%%%%%%%%%%%%%%

\subsection*{Lemma~\ref{lem:Lambda} ($\Lambda$ and $\Lambda^{\perp}$ in mb-shielded ADMGs) }

\begin{proof}
	The proof here is similar to the proofs of Theorems 4.4 and 4.5 in \citet{tsiatis2007semiparametric}. 
	
	\noindent Given $p(V)$ that factorizes with respect to an mb-shielded ADMG $\G(V),$ we can write down the following factorization using the ordinary local Markov property $V_i \ci \{\prec V_i\} \setminus \tb_\G(V_i) \mid \tb_\G(V_i):$ $p(V) = \prod_{V_i \in V} \ p(V_i \mid \tb_\G(V_i)).$	Under no restriction, the conditional density $p(V_i \mid \tb_\G(V_i)), \forall V_i \in V,$ is any positive function such that $\int p(V_i \mid \tb_\G(V_i)) d\nu(V_i) = 1,$ for all values of $ \tb_\G(V_i)),$ where $d\nu(V_i)$  is the dominating measure. Note that $p(V_i \mid \tb_\G(V_i))$s  are variationally independent. %We denote $p(V_i \mid \tb_\G(V_i)$ by $p_i.$ 
	
	The tangent space $\Lambda^*,$ corresponding to the model of the mb-shielded ADMG  $\G(V),$ is defined as the mean square closure of all parametric submodel tangent spaces. Assume there are $k$ variables in $V.$ The parametric submodel is defined as ${\cal{M}}_{\text{sub}} = \{\prod_{V_i \in V} p(V_i \mid \tb_\G(V_i); \gamma_i)\},$ where $\gamma_i, i = 1, \dots, k$ are parameters that are variationally independent and $p(V_i \mid \tb_\G(V_i); \gamma_{0i})$ denotes the true conditional density of $p(V_i \mid \tb_\G(V_i)).$ The parametric submodel tangent space is defined as the space spanned by the joint score $S_\gamma(V_1, \dots, V_k)$ given as follows, 
	{\small
		\[
		S_\gamma(V_1, \dots, V_k) = \frac{\partial}{\partial \gamma} \log p(V) = S_{\gamma_1}(V_1) + \dots + S_{\gamma_k}(V_k, \tb_\G(V_k)). 
		\]}%
	Therefore, the parametric  submodel tangent space is $\Lambda^*_{\gamma} = a_1\times S_{\gamma_1}(V_1) + \dots + a_k \times S_{\gamma_k}(V_k, \tb_\G(V_k)),$ where $a_i$'s are constants. Due to variational independence of $\gamma_i$s, $\Lambda^*_\gamma = \Lambda^*_{\gamma_1} \oplus \dots \oplus \Lambda^*_{\gamma_k},$ where $\Lambda^*_{\gamma_i} = \{a_i \times S_{\gamma_i}(V_i, \tb_\G(V_i))\}.$ The tangent space $\Lambda^*$ is then the mean-square closure of all parametric submodel tangent spaces, i.e., $\Lambda^* = \Lambda^*_1 \oplus \dots \oplus \Lambda^*_k,$ where $\Lambda^*_i$ is the mean-square closure of the parametric submodel tangent space $\Lambda^*_{
		\gamma_i}$ which corresponds to the term $p(V_i \mid \tb_\G(V_i)).$
	
	By the ordinary local Markov property, i.e., $V_i \ci \{\prec V_i\} \setminus \tb_\G(V_i) \mid \tb_\G(V_i),$ and properties of score functions for parametric models of conditional densities, the score function $S_i(.)$ must be a function of only $\{V_i, \tb_\G(V_i)\}$ and must have conditional expectation $\E[S_i(V_i, \tb_\G(V_i)) \mid \tb_\G(V_i)] = 0.$ Consequently, any element spanned by $S_i(V_i, \tb_\G(V_i))$ must belong to $\Lambda^*_i;$ hence $\Lambda^*_i = \{\alpha(V_i, \tb_\G(V_i)) \mid \E[\alpha \mid \tb_\G(V_i)] = 0\}.$ Further, in order to show that $\Lambda^*_i$'s are orthogonal, we need to show that $\E[h_i \times h_j] = 0,$ where $h_i \in \Lambda^*_i$ and $h_j \in \Lambda^*_j, $
	
	{\small
		\begin{align*}
			\E\big[ h_i \times h_j \big] 
			&= \E\Big[ h_i \times \E\big[ h_j \ \big| \ V_i, \tb_\G(V_i) \big]  \Big] \\
			&= \E\bigg[ h_i \times \E\Big[ \E\big[ h_j \ \big| \ V_i, \tb_\G(V_i), \tb_\G(V_j) \big] \ \Big| \ V_i, \tb_\G(V_i) \Big] \bigg] \\
			&= \E\bigg[ h_i \times \E\Big[ \E\big[ h_j \ \big| \ \tb_\G(V_j) \big] \ \big| \ V_i, \tb_\G(V_i) \Big] \bigg] = 0.
		\end{align*}
	}%
	The projection $h_i$ is in $\Lambda^*_i.$ Therefore, we only need to show that $h - h_i$ is orthogonal to all elements in $\Lambda^*_i.$ Consider an arbitrary element $\ell \in \Lambda^*_i, $
	
	{\small
		\begin{align*}
			\E[(h - h_i) &\times \ell] 
			= \E\Big[ \ell \times \big( \E[ h  \mid V_i, \tb_\G(V_i) ] - h_i \big) \Big]
			= \E[\ell \times \E[h \mid \tb_\G(V_i)]] \\
			&= \E\Big[ \E\big[ \ell \times  \E[h \mid \tb_\G(V_i)] \ \big| \ V_i, \tb_\G(V_i)  \big]  \Big] 
			= \E\Big[  \E[h \mid \tb_\G(V_i)] \times  \E\big[ \ell \ \big| \ V_i, \tb_\G(V_i)  \big]  \Big]= 0. 
		\end{align*}
	}

	\noindent Now regarding the orthogonal complement $\Lambda^{*\perp}$ in mb-shielded ADMGs, we know that $\Lambda^{*\perp} = \{ h - \pi[h \mid \Lambda^*], \forall h \in \mathbb{H}\},$ by definition. For a given $h \in \mathbb{H},$ we have $h = h_1 + \dots + h_k,$ where $h_i \in \Lambda_i$ as $\Lambda_i$ is defined in Section~\ref{sec:nps}. According to Section~\ref{sec:nps}, $h_i$ is any function of $V_1, \dots, V_i$, such that $\E[h_i \mid V_1, \dots, V_{i-1}] = 0.$ Therefore, 
	\begin{align*}
		h - \pi[h \mid \Lambda^*] 
		&= \big(h_1 + \dots + h_k \big) - \pi[h_1 + \dots + h_k \mid \Lambda^*_1 \oplus \dots \Lambda^*_k] \\
		&= \sum_{i = 1}^k h_i - \pi[h_i \mid \Lambda^*_1 \oplus \dots \Lambda^*_k] 
		= \sum_{i = 1}^k h_i - \pi[h_i \mid \Lambda^*_i] \\
		%&= \sum_{i = 1}^k h_i - \Big( \E[h_i \mid V_i, \tb_\G(V_i)] - \E[h_i \mid \tb_\G(V_i)]  \Big) \\
		&= \sum_{i = 1}^k h_i - \E[h_i \mid V_i, \tb_\G(V_i)] + \E[h_i \mid \tb_\G(V_i)].
	\end{align*}%
	The third equality holds since $\Lambda_i$ is orthogonal to $\Lambda^*_j,$ for $i, j = 1, \dots, k,$ such that $i \not= j.$ 
	Note that $h_i \equiv \E[h(V) \mid V_1, \dots, V_i] - \E[h(V) \mid V_1, \dots, V_{i-1}].$ Since $h(V)$ is an arbitrary element of the Hilbert space, without loss of generality, we can replace $ \E[h(V) \mid V_1, \dots, V_i]$ with $\alpha_i(V_1, \dots, V_i).$ Therefore, $h_i = \alpha_i - \E[\alpha_i \mid V_1, \dots, V_{i-1}].$ Substituting $h_i$ in the above equation yields the following. 
	
	\begin{align*}
		h_i - \E[h_i \mid V_i, &\tb_\G(V_i)] + \E[h_i \mid \tb_\G(V_i)] \\
		&=  \alpha_i - \E[\alpha_i \mid V_1, \dots, V_{i-1}] \\
		&\hspace{0.5cm} - \E[\alpha_i \mid V_i, \tb_\G(V_i)] + \E\Big[ \E[\alpha_i \mid V_1, \dots, V_{i-1}]   \Big| \  V_i, \tb_\G(V_i) \Big] \\
		&\hspace{0.5cm} + \E[\alpha_i \mid \tb_\G(V_i)] - \E\Big[ \E[\alpha_i \mid V_1, \dots, V_{i-1}]   \Big| \  \tb_\G(V_i) \Big] \\
		&= \alpha_i - \E[\alpha_i \mid V_1, \dots, V_{i-1}] - \E[\alpha_i \mid V_i, \tb_\G(V_i)] + \E[\alpha_i \mid \tb_\G(V_i)] \\
		&= \Big\{\alpha_i - \E[\alpha_i \mid V_1, \dots, V_{i-1}] \Big\} - \bigg\{ \E\Big[\alpha_i  - \E[\alpha_i \mid V_i, \dots, V_{i-1}] \ \Big| \ V_i, \tb_\G(V_i) \Big] \bigg\}.
	\end{align*}
	Consequently, the orthogonal complement of the tangent space is the following, 
	\begin{align*}
		\Lambda^{*\perp} = \Bigg\{ \sum_{V_i \in V} \ \alpha_i(V_1, \dots, V_i) - \E[\alpha_i \mid V_i, \tb_\G(V_i)] \bigg\},
	\end{align*}
	where $\alpha_i$ is any function of $V_1, \dots, V_i$ such that $\E[\alpha_i \mid V_1, \dots, V_{i-1}] = 0,$ i.e., $\alpha_i \in \Lambda_i.$
\end{proof}

%%%%%%%%%%%%%%%%%%%%%%%%%%%%%%%%%%%%%%%%%%%%%

\subsection*{Lemma~\ref{lem:primal_ipw} (Primal IPW formulation)}

\begin{proof}
	Our goal is to demonstrate that the primal IPW formulation is equivalent to the identifying functional of the target parameter $\psi(t)$ shown in Eq.~\ref{eq:tian2002rearrange} and restated below.
	{\small
		\begin{align*}
			\psi(t) = \sum_{V\setminus T}\  \prod_{V_i \in V\setminus D_T} p(V_i \mid \tb_\G(V_i)) \bigg\vert_{T=t} \times \sum_T \prod_{D_i \in D_T} p(D_i \mid \tb_{\G}(D_i)) \times Y.
		\end{align*}
	}%
	The primal IPW formulation for the target $\psi(t)$ is,
	\begin{align*}
		\E[\beta_{\text{primal}}(t)] &\equiv \E\bigg[ \frac{\I(T=t)}{q_{D_T}(T \mid \mb_\G(T))} \times Y\ \bigg], 
	\end{align*} % 
	where $q_{D_T}(D_T \mid \pa_\G(D_T)) = \prod_{V_i \in D_T} \ p(V_i \mid \tb_\G(V_i)),$
	and 
	{\small
		\begin{align*}
			q_{D_T}(T \mid \mb_\G(T)) 
			&=  q_{D_T}(T \mid D_T \cup \pa_{\G}(D_T) \setminus T) = \frac{q_{D_T}(D_T \mid \pa_{\G}(D_T))}{q_{D_T}(D_T \setminus T \mid \pa_\G(D_T))} \\
			&= \frac{q_{D_T}(D_T \mid \pa_\G(D_T))}{\sum_T q_{D_T}(D_T \mid \pa_\G(D_T))} = \frac{\prod_{V_i \in D_T} p(V_i \mid \tb_\G(V_i))}{\sum_T \prod_{V_i \in D_T} p(V_i \mid \tb_\G(V_i))} \\
			&=  \frac{\prod_{V_i \in \mathbb{L}} p(V_i \mid \tb_\G(V_i))}{\sum_T \prod_{V_i \in \mathbb{L}} p(V_i \mid \tb_\G(V_i))}. 
		\end{align*}
	}%
	The last equality holds because the conditional densities of $V_i \in \mathbb{C},$ does not depend on $T,$ and they cancel out from the numerator and denominator.  Therefore, product in the ratio is over the variables in $D_T \cap \{\succeq T\}$ which we have denoted by $\mathbb{L}.$
	Therefore,
	{\small
		\begin{align*}
			\E[\beta_{\text{primal}}(t)] 
			&= \E\bigg[ \I(T=t) \times \frac{ \sum_T\ \prod_{D_i \in \mathbb{L}}\ p(D_i \mid \tb_\G(D_i))}{\prod_{D_i \in \mathbb{L}} \ p(D_i \mid \tb_\G(D_i))} \times Y\ \bigg] 
			\\
			&= \sum_{V} \prod_{V_i \in V} p(V_i \mid \tb_\G(V_i)) \times \I(T=t) \times \frac{ \sum_T\ \prod_{D_i \in \mathbb{L}}\ p(D_i \mid \tb_\G(D_i))}{\prod_{D_i \in \mathbb{L}} \ p(D_i \mid \tb_\G(D_i))} \times Y
			\\
			&= \sum_{V}\  \I(T=t) \times \prod_{V_i \in V \setminus \mathbb{L}} p(V_i \mid \tb_\G(V_i)) \\
			&\hspace{3.5cm}\times \prod_{D_i \in \mathbb{L}} p(D_i \mid \tb_\G(D_i)) \times \frac{ \sum_T\ \prod_{D_i \in \mathbb{L}}\ p(D_i \mid \tb_\G(D_i))}{\prod_{D_i \in \mathbb{L}} \ p(D_i \mid \tb_\G(D_i))} \times Y
			\\
			&= \sum_{V}\  \I(T=t) \times \prod_{V_i \in V\setminus \mathbb{L}} p(V_i \mid \tb_\G(V_i)) \times \sum_T\ \prod_{D_i \in \mathbb{L}}\ p(D_i \mid \tb_\G(D_i)) \times Y.
		\end{align*}
	}
	In the second equality, we evaluated the outer expectation with respect to the joint $p(V).$
	In the third equality, we partitioned the joint into factors for the set $\mathbb{L}$ and factors for $V \setminus \mathbb{L}$. In the fourth equality, we canceled out the factors involved in the denominator of the primal IPW with the corresponding terms in the joint. 
	
	We can then move the conditional factors of pre-treatment variables in the district of $T$ past the summation over $T$ as these factors are not functions of $T.$ Finally, we evaluate the indicator function, concluding the proof. That is,
	{\small
		\begin{align*}
			\psi_{\text{primal}}&=\sum_V \I(T=t)\times \prod_{V_i \in V\setminus D_T} p(V_i \mid \tb_\G(V_i)) \times \sum_T \prod_{D_i \in D_T} p(D_i \mid \tb_{\G}(D_i)) \times Y\\[1em]
			&=\sum_{V\setminus T}\  \prod_{V_i \in V\setminus D_T} p(V_i \mid \tb_\G(V_i)) \bigg\vert_{T=t} \times \sum_T \prod_{D_i \in D_T} p(D_i \mid \tb_{\G}(D_i)) \times Y 
			= \psi(t)
		\end{align*}
	}
\end{proof}

%%%%%%%%%%%%%%%%%%%%%%%%%%%%%%%%%%%%%%%%%%%%%

\subsection*{Lemma~\ref{lem:dual_ipw} (Dual IPW formulation)}

\begin{proof}
	The proof strategy is similar to the one used for the primal IPW. The dual IPW formulation for the target $\psi(t)$ is,
	
	{\small
		\begin{align*}
			\E[\beta_{\text{dual}}(t)] 
			&= \E\bigg[ \frac{\prod_{M_i \in \mathbb{M}^*} \ p(M_i \mid \tb_\G(M_i)) \ \vert_{T=t}}{\prod_{M_i \in \mathbb{M}^*} \ p(M_i \mid \tb_\G(M_i))}\times Y  \bigg] 
			\\
			&= \sum_V \prod_{V_i \in V} p(V_i \mid \tb_\G(V_i)) \times \frac{\prod_{M_i \in \mathbb{M}^*} \ p(M_i \mid \tb_\G(M_i)) \ \vert_{T=t}}{\prod_{M_i \in \mathbb{M}^*} \ p(M_i \mid \tb_\G(M_i))}\times Y 
			\\
			&= \sum_V \prod_{V_i \in V\setminus \mathbb{M}^*} p(V_i \mid \tb_\G(V_i)) \\
			&\hspace{1.5cm} \times \prod_{M_i \in \mathbb{M}^*} p(M_i \mid \tb_\G(M_i)) \times \frac{\prod_{M_i \in \mathbb{M}^*} \ p(M_i \mid \tb_\G(M_i)) \ \vert_{T=t}}{\prod_{M_i \in \mathbb{M}^*} \ p(M_i \mid \tb_\G(M_i))}\times Y
			\\
			&= 	\sum_V \prod_{V_i \in V\setminus \mathbb{M}^*} p(V_i \mid \tb_\G(V_i)) \times \prod_{M_i \in \mathbb{M}^*}\ p(M_i \mid \tb_\G(M_i)) \ \vert_{T=t} \times Y
			\\
			&= \sum_{V\setminus T}\  \prod_{V_i \in V\setminus \{\mathbb{M}^* \cup D_T\}} \ p(V_i \mid \tb_\G(V_i)) \times \prod_{M_i \in \mathbb{M}^*}\ p(M_i \mid \tb_\G(M_i)) \ \vert_{T=t}\\
			&\hspace{5cm} \times \sum_T \ \prod_{D_T} \ p(D_i \mid \tb_\G(D_i)) \times Y.
		\end{align*}
	}
	In the above derivation, we first evaluated the outer expectation with respect to the joint $p(V).$ We then partitioned the joint into factors corresponding to $\mathbb{M}^*$ and $V\setminus \mathbb{M}^*$. The factors involved in the denominator of the dual IPW then canceled out with the corresponding terms in the joint. The last equality holds because by the definition of the inverse Markov pillow, $\mathbb{M}^*$ contains all variables not in the district of $T$ such that $T$ is a member of its Markov pillow. In the above expression, factors corresponding to the inverse Markov pillow of $T$ are evaluated at $T=t.$ Consequently, the only factors above that are still functions of $T$ are the ones corresponding to the district of $T.$ This allows us to push the summation over $T$. 
	
	Finally, since the summation over $T$ will prevent factors within the district of $T$ from being evaluated at $T=t,$ we can simply apply the evaluation to the entire functional and merge the sets not involved in the district of $T$ above. That is,
	{\small
		\begin{align*}
			\psi_{\text{dual}} 
			&= \sum_{V\setminus T}\  \prod_{V_i \in V\setminus D_T} p(V_i \mid \tb_\G(V_i)) \bigg\vert_{T=t} \times \sum_T \prod_{D_i \in D_T} p(D_i \mid \tb_{\G}(D_i)) \times Y
			= \psi(t).
		\end{align*}
	}%	
\end{proof}

%%%%%%%%%%%%%%%%%%%%%%%%%%%%%%%%%%%%%%%%%%%%%

\subsection*{Theorem~\ref{lem:ipw_consistency} (Primal and Dual IPW estimators)}

\begin{proof} 
	The required positivity assumptions for $\widehat{\psi(t)}_{\text{primal}}$ and $\widehat{\psi(t)}_{\text{dual}}$ are as follows.
	{\small
		\begin{align*}
			\widehat{\psi(t)}_{\text{primal}}: & \quad \forall L_i \in \mathbb{L}, \ p(L_i \mid \tb_\G(L_i)) > \epsilon_{l_i} \ \text{ almost surely for some non-negative } \epsilon_{l_i},
			\\
			\widehat{\psi(t)}_{\text{dual}}: & \quad \forall M_i \in \mathbb{M}^*, \ p(M_i \mid \tb_\G(M_i)) > \epsilon_{m_i} \ \text{ almost surely for some non-negative } \epsilon_{m_i}.
		\end{align*}
	}%
	For the usual regularity conditions see Theorem 1A in \cite{robins1992estimating}. Under these conditions, the proof of asymptotic normality is fairly standard once we show the corresponding estimating equations are unbiased. In Lemmas~(\ref{lem:primal_ipw}) and (\ref{lem:dual_ipw}), we proved the unbiasedness of these two estimators, i.e., we showed $\E[\beta(t)_{\text{primal}}] =  \E[\beta(t)_{\text{dual}}] = \psi(t).$ 
	
	For the sake of completeness, we walk through finding the asymptotic variance. Let $U(\psi, \eta)$ denote either $\beta(t)_{\text{primal}} - \psi(t)$ or $\beta(t)_{\text{dual}} - \psi(t).$ ($\eta$ denotes the set of nuisance parameters.)
	Given our proof in Lemmas~(\ref{lem:primal_ipw}) and (\ref{lem:dual_ipw}), we know that $\E[U(\psi_0, \eta_0)] = 0.$ In order to estimate $\psi_0$, we use the estimating equation: $\frac{1}{n} \times \sum_{i = 1}^n U_i\big(\psi(\hat{\eta}), \hat{\eta}\big) = 0.$ The Taylor series expansion of $U_i\big(\psi(\hat{\eta}), \hat{\eta}\big)$ around $\psi_0$ is
	{\small
		\begin{align*}
			0 
			= &\frac{1}{\sqrt{n}} \sum_{i=1}^n U_i\Big(\psi(\hat{\eta}), \hat{\eta}\Big)  
			\\
			= & \frac{1}{\sqrt{n}}  \sum_{i=1}^n \ U_i\Big(\psi_0, \hat{\eta} \Big) +  
			\frac{1}{\sqrt{n}}  \sum_{i=1}^n
			\frac{\partial}{\partial \psi} \ U_i\Big(\psi_0, \hat{\eta} \Big) \times ({\psi}(\hat{\eta}) - \psi_0) + o_p(1)
			\\
			= & \underset{(a)}{\underbrace{ \frac{1}{\sqrt{n}}  \sum_{i=1}^n \ U_i\Big(\psi_0, \hat{\eta} \Big)}} +  
			\underset{(b)}{\underbrace{	\frac{1}{n}  \sum_{i=1}^n
					\frac{\partial}{\partial \psi} \ U_i\Big(\psi_0, \hat{\eta} \Big) } } \times \sqrt{n}  ({\psi}(\hat{\eta}) - \psi_0) + o_p(1)  \nonumber 
		\end{align*}
	}
	{\small
		\begin{align*}
			(a) 
			&= \frac{1}{\sqrt{n}}  \sum_{i=1}^n \ U_i\Big(\psi_0, \hat{\eta} \Big) 
			\\
			&= \frac{1}{\sqrt{n}}  \sum_{i=1}^n \ U_i\Big(\psi_0, \eta_0 \Big)  
			+ \frac{1}{n} \sum_{i=1}^n \frac{\partial U_i(\psi_0, \eta_0)}{\partial \eta} \times \sqrt{n}(\widehat{\eta} - \eta_0) +  o_p(1)
			\\
			\text{as }n \rightarrow \infty \ (a) &\rightarrow  \frac{1}{\sqrt{n}}  \sum_{i=1}^n \ U_i\Big(\psi_0, \eta_0 \Big)  
			+ \E\left[ \frac{\partial U(\psi_0, \eta_0)}{\partial \eta}  \right] \times \sqrt{n}(\widehat{\eta} - \eta_0) 
			\\
			\\
			(b) 
			&= \frac{1}{n}  \sum_{i=1}^n \frac{\partial}{\partial \psi} \ U_i\Big(\psi_0, \hat{\eta} \Big)  
			\\
			&= \frac{1}{n}  \sum_{i=1}^n \frac{\partial}{\partial \psi} \bigg\{ U_i\Big(\psi_0, \eta_0 \Big)  +  \frac{\partial}{\partial \eta} U_i(\psi_0, \eta_0) \times (\widehat{\eta} - \eta_0)  \bigg\}
			\\
			\text{as }n \rightarrow \infty \ (b) &\rightarrow  \E\bigg[  \frac{\partial}{\partial \psi} U\Big(\psi_0, \eta_0 \Big)  \bigg]
		\end{align*}
	}
	
	\noindent By substituting (a) and (b), we get:
	
	{\small
		\begin{align*}
			&0 
			= \frac{1}{\sqrt{n}}  \sum_{i=1}^n \ U_i\Big(\psi_0, \eta_0 \Big)  
			+ \E\left[ \frac{\partial U(\psi_0, \eta_0)}{\partial \eta}  \right] \times \sqrt{n}(\widehat{\eta} - \eta_0)  +  \E\bigg[  \frac{\partial}{\partial \psi} U\Big(\psi_0, \eta_0 \Big)  \bigg] \times \sqrt{n} \Big({\psi}(\hat{\eta}) - \psi_0 \Big) + o_p(1).	
		\end{align*}
	}
	\noindent Consequently, 
	{\small
		\begin{align*}
			&\sqrt{n} \Big({\psi}(\hat{\eta}) - \psi_0 \Big) = - \E\bigg[  \frac{\partial}{\partial \psi} U\Big(\psi_0, \eta_0 \Big)  \bigg]^{-1} \times \Bigg(  \frac{1}{\sqrt{n}}  \sum_{i=1}^n \ U_i\Big(\psi_0, \eta_0 \Big)  
			+ \E\left[ \frac{\partial U(\psi_0, \eta_0)}{\partial \eta}  \right] \times \sqrt{n}(\widehat{\eta} - \eta_0)   \Bigg) + o_p(1).
		\end{align*}
	}
	
	\noindent The regularity conditions guarantee that,
	{\small
		\begin{align*}
			\sqrt{n}(\widehat{\eta} - \eta_0) = -\E\bigg[ \frac{\partial S}{\partial \eta} \bigg]^{-1} \times \frac{1}{\sqrt{n}} S(\eta_0) + o_p(1).
		\end{align*}
	}
	
	{\small
		\begin{align*}
			\sqrt{n} \Big({\psi}(\hat{\eta}) - \psi_0 \Big) &= - \E\bigg[  \frac{\partial}{\partial \psi} U\Big(\psi_0, \eta_0 \Big)  \bigg]^{-1} \times \frac{1}{\sqrt{n}} \times \Bigg( \sum_{i=1}^n \ U_i\Big(\psi_0, \eta_0 \Big)  
			-\E\left[ \frac{\partial U(\psi_0, \eta_0)}{\partial \eta}  \right] \times \E\bigg[ \frac{\partial S}{\partial \eta} \bigg]^{-1} \times S(\eta_0)   \Bigg) + o_p(1). 
			\\
			&= - \tau^{-1} \times \frac{1}{\sqrt{n} }\times \sum_{i} \text{ Resid}(U_i) + o_p(1). 
		\end{align*}
	}
	
	As a consequence of central limit theorem and Slutzky's theorem, we conclude that $\sqrt{n} \big({\psi}(\hat{\eta}) - \psi_0 \big)$ is asymptotically normal with mean zero and variance 
	$$\text{Var}(\psi) =  \text{Var}(\tau^{-1} \text{Resid}) = \tau^{-1}\times \text{Var}(\text{Resid}) \times \tau^{-1}, $$
	that can be consistently estimated using the sample variance of $\text{Var}(\psi).$	
	
\end{proof}

%%%%%%%%%%%%%%%%%%%%%%%%%%%%%%%%%%%%%%%%%%%%%

\subsection*{Theorem~\ref{thm:disjoint} (Variational independence of primal IPW and dual IPW)}

\begin{proof}
	Consider the topological factorization of the observed distribution $p(V)$ for the ADMG as shown in Eq.~\ref{eq:top_factorization}.
	\begin{align*}
		p(V) = \prod_{V_i \in V} p(V_i \mid \tb_\G(V_i)).
	\end{align*}
	Note by definition, the inverse Markov pillow of $T$ does not contain elements in the district of $T,$ i.e., $\mathbb{M}^* \cap D_T = \emptyset.$ Thus, we can partition $V$ into three disjoint sets as follows:
	\begin{align*}
		\mathbb{L} &= D_T \cap \{\succeq T \}, \qquad
		\mathbb{M}^*, \qquad
		\mathbb{R} = V \setminus (\mathbb{L} \cup \mathbb{M}^*)
	\end{align*}
	The set $\mathbb{L}$ is the same as what we introduced in Section~\ref{sec:primal}. $\mathbb{M}^*$ is a subset of $\mathbb{M},$ and the remaining variables are in set $\mathbb{R} = \mathbb{C} \cup (\mathbb{M} \setminus \mathbb{M}^*).$
	The topological factorization of the observed joint can then be restated as,
	\begin{align*}
		p(V) = \prod_{R_i \in \mathbb{R}} p(R_i \mid \tb_\G(R_i)) \prod_{M_i \in \mathbb{M}^*} p(M_i \mid \tb_\G(M_i)) \prod_{L_i \in \mathbb{L}} p(L_i \mid \tb_{\G}(L_i)).
	\end{align*}
	It is then clear from the above factorization that the components of the primal IPW estimator which sit in $\mathbb{L},$ and the components of the dual IPW estimator which sit in $\mathbb{M}^*,$ form congenial and variationally independent pieces of the joint distribution $p(V).$
\end{proof}

%%%%%%%%%%%%%%%%%%%%%%%%%%%%%%%%%%%%%%%%%%%%%

\subsection*{Theorem~\ref{thm:IF_childless} (The efficient influence function of $\psi(t)$ in $\mathcal{M}_{\text{np}}$) }

\begin{proof} 
	As a reminder, we partition the set of nodes $V$ into three disjoint sets: \\ $V = \{ \mathbb{C}, \mathbb{L}, \mathbb{M} \}, $ where 
	\begin{align*}
		\mathbb{C} &= \{ C_i \in V \mid C_i \prec T\},  \quad 
		\mathbb{L} = \{L_i \in V \mid L_i \in D_T, L_i \succeq T\},  \ \text{and } \
		\mathbb{M} = V \setminus \mathbb{C} \cup \mathbb{L}. 
	\end{align*}
	The target parameter is identified via the following function of the observed data, 
	{\small
		\begin{align}
			\psi_\kappa(t) =  \sum_{V \setminus T} \ Y \times \prod_{M_i \in  \mathbb{M}} \ p_\kappa(M_i \mid \tb_\G(M_i)) \vert_{T=t}  \times \sum_T \prod_{L_i \in \mathbb{L}} p_\kappa(L_i \mid \tb_\G(L_i)) \times  p_\kappa(\mathbb{C}),
			\label{eq:id_2}
		\end{align}
	}%
	and according to Eq.~\ref{eq:pathwise_deriv},
	${\small \frac{d}{d\kappa}\psi_\kappa(t) \big|_{\kappa = 0} = \E\big[ U_{\psi_t} \times S_{\eta_0}(V) \big]}.$ Therefore,
	
	{\scriptsize
		\begin{align*}
			\frac{d}{d\kappa} \psi_\kappa(t) 
			&=  \frac{d}{d\kappa} \Big\{ \sum_{V \setminus T} \ Y \times \prod_{M_i \in  \mathbb{M}} \ p_\kappa(M_i \mid \tb_\G(M_i)) \vert_{T=t}  \times \sum_T \prod_{L_i \in \mathbb{L} \setminus T} p_\kappa(L_i \mid \tb_\G(L_i)) \times  p_\kappa(T, \mathbb{C}) \Big\} 
			\\
			&= \sum_{V \setminus T} \ Y \times  \blue{\frac{d}{d\kappa} \Big\{} \prod_{M_i \in  \mathbb{M}} \ p_\kappa(M_i \mid \tb_\G(M_i)) \vert_{T=t} \blue{\Big\}} \times \sum_T \prod_{L_i \in \mathbb{L} \setminus T} p_\kappa(L_i \mid \tb_\G(L_i)) \times  p_\kappa(T, \mathbb{C})  \quad \text{\blue{(1st Term)}} 
			\\
			&+ \sum_{V \setminus T} \ Y \times  \prod_{M_i \in  \mathbb{M}} \ p_\kappa(M_i \mid \tb_\G(M_i)) \vert_{T=t} \times \sum_T \blue{\frac{d}{d\kappa} \Big\{} \prod_{L_i \in \mathbb{L} \setminus T} p_\kappa(L_i \mid \tb_\G(L_i))  \blue{\Big\}} \times  p_\kappa(T, \mathbb{C})   \quad \text{\blue{(2nd Term)}}  
			\\
			&+ \sum_{V \setminus T} \ Y \times  \prod_{M_i \in  \mathbb{M}} \ p_\kappa(M_i \mid \tb_\G(M_i)) \vert_{T=t} \times \sum_T \prod_{L_i \in \mathbb{L} \setminus T} p_\kappa(L_i \mid \tb_\G(L_i)) \times \blue{\frac{d}{d\kappa} \Big\{} p_\kappa(T, \mathbb{C})  \blue{\Big\}}.   \quad \text{\blue{(3rd Term)}}  
		\end{align*}
	}
	
	\vspace{0.5cm}
	\noindent \textbf{First Term:} The contribution of the first term to the final IF is made of individual contributions of the elements in $\mathbb{M}.$ Since the derivation is similar, we only derive it for an element $M_j \in \mathbb{M}.$
	{\scriptsize
		\begin{align*}
			& \sum_{V \setminus T} \ 
			Y 
			\times  
			\prod_{M_i \in \{\prec M_j\} \cap \mathbb{M} } \ p_\kappa(M_i \mid \tb_\G(M_i)) \vert_{T=t} 
			\times 
			\blue{\frac{d}{d\kappa} \Big\{} p_\kappa(M_j \mid \tb_\G(M_j)) \vert_{T=t} \blue{\Big\}}   \\
			&\hspace{1cm} \times 
			\prod_{M_i \in \{\succ M_j\} \cap \mathbb{M} }  \ p_\kappa(M_i \mid \tb_\G(M_i)) \vert_{T=t}
			\times
			\sum_T \prod_{L_i \in \mathbb{L}} p_\kappa(L_i \mid \tb_\G(L_i)) \times  p_\kappa(\mathbb{C})  
			\\  
			&\overset{(1)}{=} \sum_{V \setminus \{T, \{\preceq M_j\}\}} \ 
			\prod_{M_i \in \{\prec M_j\} \cap \mathbb{M} }  \ p_\kappa(M_i \mid \tb_\G(M_i)) \vert_{T=t}
			\times
			\blue{\frac{d}{d\kappa} \Big\{} p_\kappa(M_j \mid \tb_\G(M_j)) \vert_{T=t} \blue{\Big\}} 
			\\
			&\hspace{1cm} 
			\times 
			\sum_{T \cup \{\succ M_j\}} \ 
			Y 
			\times  
			\prod_{V_i \in \mathbb{L} \cup \{\{\succ M_j\} \cap \mathbb{M} \} } p_\kappa(V_i \mid \tb_\G(V_i)) \Big\vert_{T=t\text{ if } V_i \in \mathbb{M}} \times  p_\kappa(\mathbb{C}) 
			\\
			&\overset{(2)}{=} \sum_{\preceq M_j} \ \frac{\mathbb{I}(T = t)}{\prod_{L_i \prec M_j } \ p(L_i \mid \tb_\G(L_i))} \times 
			S(M_j \mid \tb_\G(M_j)) 
			\times 
			\prod_{V_i \in \{\preceq M_j \}}  \ p(V_i \mid \tb_\G(V_i))
			\\
			&\hspace{1cm} \times 
			\sum_{T \cup \{\succ M_j\}} \ 
			Y 
			\times  
			\prod_{V_i \in \mathbb{L} \cup \{\{\succ M_j\} \cap \mathbb{M} \} } p(V_i \mid \tb_\G(V_i)) \Big\vert_{T=t\text{ if } V_i \in \mathbb{M}} 
			\\
			&\overset{(3)}{=} \E\Bigg[
			\frac{\mathbb{I}(T = t)}{\prod_{L_i \prec M_j } \ p(L_i \mid \tb_\G(L_i))} 
			\times 
			\underbrace{
				\sum_{T \cup \{\succ M_j\}} \ Y \times  
				\prod_{V_i \in \mathbb{L} \cup  \{\{\succ M_j\} \cap \mathbb{M} \} } p(V_i \mid \tb_\G(V_i)) \Big\vert_{T=t\text{ if } V_i \in \mathbb{M}}}_{\coloneqq f( \preceq M_j)}
			\\
			&\hspace{6cm} \times   S(M_j \mid \tb_\G(M_j))
			\Bigg]
			\\
			&\overset{(4)}{=} \E\Bigg[
			\frac{\mathbb{I}(T = t)}{\prod_{L_i \prec M_j } \ p(L_i \mid \tb_\G(L_i))} \times \Big( f( \preceq M_j) - \sum_{M_j} f( \preceq M_j) \times p(M_j \mid \tb_\G(M_j)) \Big) 
			\times  S(M_j \mid \tb_\G(M_j))
			\Bigg]
			\\
			&\overset{(5)}{=} \E\Bigg[
			\frac{\mathbb{I}(T = t)}{\prod_{L_i \prec M_j } \ p(L_i \mid \tb_\G(L_i))} \times \Big( f( \preceq M_j) - \sum_{M_j} f( \preceq M_j) \times p(M_j \mid \tb_\G(M_j)) \Big) 
			\times  S(V)
			\Bigg] \\
		\end{align*}
	}
	The first equality follows from the fact that terms corresponding to $M_i \in \{\prec M_j\}$ are not functions of elements in $\{ \succ M_j \} $ and of $Y$.
	The second equality follows by term grouping, the definition of conditional scores, and term cancellation.
	The third equality is by definition of joint expectation.
	The fourth and fifth equalities are implied by the fact that conditional scores have expected value of $0$ (given their conditioning set). Therefore, the contribution of $M_j \in \mathbb{M}$ is the following:
	
	{\scriptsize
		\begin{align*}
			&\frac{\mathbb{I}(T = t)}{\prod_{L_i \prec M_j } \ p(L_i \mid \tb_\G(L_i))} \times \Big( \sum_{T \cup \{\succ M_j\} } \ Y \times  
			\prod_{V_i \in \mathbb{L} \cup \{\{\succ M_j\} \cap \mathbb{M} \} } p(V_i \mid \tb_\G(V_i)) \Big\vert_{T=t\text{ if } V_i \in \mathbb{M}} 
			\\
			&\hspace{5cm} -  \sum_{T \cup \{\succeq M_j\}} \ Y \times  
			\prod_{V_i \in \mathbb{L} \cup \{\{\succeq M_j\} \cap \mathbb{M} \} } p(V_i \mid \tb_\G(V_i)) \Big\vert_{T=t\text{ if } V_i \in \mathbb{M}}
			\ \Big).
		\end{align*}
	}
	
	\vspace{0.5cm}
	\noindent \textbf{Second Term:} The contribution of the second term to the final IF is made of individual contributions of the elements in $\mathbb{L} \setminus T.$ Since the derivation is similar, we only derive it for an element $L_j \in \mathbb{L} \setminus T.$
	{\scriptsize
		\begin{align*}
			& \sum_{V \setminus T} \ Y \times  \prod_{M_i \in  \mathbb{M}} \ p_\kappa(M_i \mid \tb_\G(M_i)) \vert_{T=t} \times \sum_T \bigg\{ \prod_{L_i \in \{\prec L_j\} \cap \mathbb{L} \setminus T} \ p_\kappa(L_i \mid \tb_\G(L_i))
			\\
			&\hspace{2cm} \times \blue{\frac{d}{d\kappa} \Big\{} p_\kappa(L_j \mid \tb_\G(L_j)) \blue{\Big\}}  \times \prod_{L_i \in \{\succ L_j\} \cap \mathbb{L} \setminus T} p_\kappa(L_i \mid \tb_\G(L_i))
			\bigg\} \times  p_\kappa(T, \mathbb{C})  
			\\
			&\overset{(1)}{=}  \sum_V \ Y \times \prod_{V_i \in \{ \succ L_j \} } \ p_\kappa(V_i \mid \tb_\G(V_i))\big\vert_{T=t\text{ if } V_i \in \mathbb{M}} \times  \blue{\frac{d}{d\kappa} \Big\{} p_\kappa(L_j \mid \tb_\G(L_j)) \blue{\Big\}} 
			\\
			&\hspace{6cm} \times \prod_{V_i \in \{\prec L_j\}} \ p_\kappa(V_i \mid \tb_\G(V_i))\big\vert_{T=t\text{ if } V_i \in \mathbb{M}} 
			\\
			&\overset{(2)}{=} \sum_{\preceq L_j} \
			\underbrace{ \sum_{\succ L_j} \ Y \times \prod_{V_i \in \{ \succ L_j \} } \ p(V_i \mid \tb_\G(V_i))\big\vert_{T=t\text{ if } V_i \in \mathbb{M}}}_{f(\preceq L_j)} \times 
			S(L_j \mid \tb_\G(L_j)) \\
			&\hspace{6cm} \times \prod_{V_i \in \{\preceq L_j\}} \ p(V_i \mid \tb_\G(V_i))\big\vert_{T=t\text{ if } V_i \in \mathbb{M}} 
			\\
			&\overset{(3)}{=} \sum_{\preceq L_j} \ f(\preceq L_j)
			\times  \frac{\prod_{M_i \in \mathbb{M} \cap \{\prec L_j\}} \ p(M_i \mid \tb_\G(M_i)) \big\vert_{T=t}}{\prod_{M_i \in \mathbb{M} \cap \{\prec L_j\}} \ p(M_i \mid \tb_\G(M_i))}
			\times  S(L_j \mid \tb_\G(L_j))  \times \prod_{V_i \in \{\preceq L_j\}} \ p(V_i \mid \tb_\G(V_i))
			\\
			&\overset{(4)}{=} \E\Bigg[
			\frac{\prod_{M_i \in \mathbb{M} \{\cap \prec L_j\}} \ p(M_i \mid \tb_\G(M_i)) \big\vert_{T=t}}{\prod_{M_i \in \mathbb{M} \cap \{\prec L_j\}} \ p(M_i \mid \tb_\G(M_i))}
			\times f(\preceq L_j) \times  S(L_j \mid \tb_\G(L_j)) 
			\Bigg] 
			\\
			&\overset{(5)}{=} \E\Bigg[
			\frac{\prod_{M_i \in \mathbb{M} \cap \{\prec L_j\}} \ p(M_i \mid \tb_\G(M_i)) \big\vert_{T=t}}{\prod_{M_i \in \mathbb{M} \cap \{\prec L_j\}} \ p(M_i \mid \tb_\G(M_i))}
			\times \Big(f(\preceq L_j) - \sum_{L_j} \ f(\preceq L_j) \times p(L_j \mid \tb_\G(L_j)) \Big) \times  S(L_j \mid \tb_\G(L_j)) 
			\Bigg] 
			\\
			&\overset{(6)}{=} \E\Bigg[
			\frac{\prod_{M_i \in \mathbb{M} \cap \{\prec L_j\}} \ p(M_i \mid \tb_\G(M_i)) \big\vert_{T=t}}{\prod_{M_i \in \mathbb{M} \cap \{\prec L_j\}} \ p(M_i \mid \tb_\G(M_i))}
			\times \Big(f(\preceq L_j) - \sum_{L_j} \ f(\preceq L_j) \times p(L_j \mid \tb_\G(L_j)) \Big) \times  S(V) 
			\Bigg] 
		\end{align*}
	}
	The first equality follows from the fact that terms corresponding to $M_i \in \mathbb{M}$ are not functions of $T$, the fact that $\mathbb{C,M,L}$ partition $V$, and term grouping.
	The second equality is by definition of conditional scores.
	The third equality is by term cancellation.
	The fourth is by definition of joint expectations, the fifth and sixth equalities are implied by the fact that conditional scores have expected value of $0$ (given their conditioning set). Therefore, the contribution of $L_j \in \mathbb{L} \setminus T$ is the following:
	
	{\scriptsize
		\begin{align*}
			&\frac{\prod_{M_i \in \mathbb{M} \cap \{\prec L_j\}} \ p(M_i \mid \tb_\G(M_i)) \big\vert_{T=t}}{\prod_{M_i \in \mathbb{M} \cap \{\prec L_j\}} \ p(M_i \mid \tb_\G(M_i))}
			\times \Big( \sum_{\succ L_j} \ Y \times \prod_{V_i \in \{ \succ L_j \} } \ p(V_i \mid \tb_\G(V_i))\big\vert_{T=t\text{ if } V_i \in \mathbb{M}} \\
			&\hspace{6cm} - \sum_{\succeq L_j} \ Y \times \prod_{V_i \in \{ \succeq L_j \} } \ p(V_i \mid \tb_\G(V_i))\big\vert_{T=t\text{ if } V_i \in \mathbb{M}} \ \Big).
		\end{align*}
	}
	
	\vspace{0.5cm}
	\noindent \textbf{Third Term:} The contribution of the last term to the final IF is as follows. 
	{\scriptsize
		\begin{align*}
			&\sum_{V \setminus T} \ Y \times  \prod_{M_i \in  \mathbb{M}} \ p_\kappa(M_i \mid \tb_\G(M_i)) \vert_{T=t} \times \sum_T \prod_{L_i \in \mathbb{L} \setminus T} p_\kappa(L_i \mid \tb_\G(L_i)) \times \blue{\frac{d}{d\kappa} \Big\{} p_\kappa(T, \mathbb{C})  \blue{\Big\}} 
			\\
			&\overset{(1)}{=} \sum_{T, \mathbb{C}} \bigg\{ 
			\underbrace{\sum_{V \setminus T, \mathbb{C}} \ Y \times  \prod_{M_i \in  \mathbb{M}} \ p_\kappa(M_i \mid \tb_\G(M_i)) \vert_{T=t} \times  \prod_{L_i \in \mathbb{L} \setminus T} p_\kappa(L_i \mid \tb_\G(L_i))
			}_{f(T, \mathbb{C})}  
			\bigg\} \times \blue{\frac{d}{d\kappa} \Big\{} p_\kappa(T, \mathbb{C})  \blue{\Big\}}.
			\\
			&\overset{(2)}{=} \sum_{T, \mathbb{C}} \ f(T, \mathbb{C}) \times S(T, \mathbb{C}) \times p(T, \mathbb{C}) =  \E\Big[f(T, \mathbb{C}) \times S(T, \mathbb{C})   \Big] 
			\\
			&\overset{(3)}{=}  \E\Big[  \Big(f(T, \mathbb{C})- \sum_{T, \mathbb{C}} \ f(T, \mathbb{C}) \times p(T, \mathbb{C}) \Big) \times S(T, \mathbb{C}) \Big] 
			\\
			&\overset{(4)}{=} \E\Big[  \Big(f(T, \mathbb{C}) - \psi(t) \Big) \times S(V) \Big] .
		\end{align*}
	}
	The first equality is term grouping, the second is by definition of marginal scores, the third and fourth equalities are implied by the fact that scores have expected value $0$. Therefore, the contribution of the last term is the following:
	
	{\scriptsize
		\begin{align*}
			& \sum_{V \setminus \{T, \mathbb{C}\}} \ Y \times  \prod_{M_i \in  \mathbb{M}} \ p(M_i \mid \tb_\G(M_i)) \big\vert_{T=t} \times  \prod_{L_i \in \mathbb{L} \setminus T} p(L_i \mid \tb_\G(L_i)) - \psi(t).  
		\end{align*}
	}%
	Putting all these together yields the final influence function. 
\end{proof}

%%%%%%%%%%%%%%%%%%%%%%%%%%%%%%%%%%%%%%%%%%%%%

\subsection*{Theorem~\ref{lem:dr} (Double robustness of augmented primal IPW) }

\begin{proof}
	We need to show that under correct specification of conditional densities in either $\{p(M_i \mid \tb_\G(M_i)), \forall M_i \in \mathbb{M} \}$ or $\{p(L_i \mid \tb_\G(L_i)), \forall L_i \in \mathbb{L} \},$ the influence function in Theorem~\ref{thm:IF_childless} remains to be mean zero.  We break this down into two scenarios. 
	
	\vspace{0.5cm}	
	\noindent \textbf{Scenario 1.} Assume models in $\mathbb{L}$ are correctly specified, and let $p^*(M_i \mid \tb_\G(M_i))$ denote the misspecified model for $p(M_i \mid \tb_\G(M_i)), \forall M_i \in \mathbb{M}.$ 
	We note that for any $L_j \in \mathbb{L} \setminus T,$ the following line in the IF evaluates to zero in expectation. 
	
	{\scriptsize
		\begin{align*}
			&\E\bigg[ \ \frac{\prod_{M_i \prec L_j} p^*(M_i \mid \tb_\G(M_i)) \vert_{T=t} }{\prod_{M_i \prec L_j} p^*(M_i \mid \tb_\G(M_i))} \bigg(\  \sum_{\succ L_j} Y \times \prod_{L_i \in \mathbb{L} \cap \{\succ L_j \}} p(L_i \mid \tb_\G(L_i)) \times  \prod_{M_i \in \mathbb{M} \cap \{\succ L_j\}} p^*(M_i \mid \tb_\G(M_i)) \vert_{T=t}  
			\\
			&\hspace{1cm} - \sum_{\succeq L_j} \ Y \times  \prod_{L_i \in \mathbb{L} \cap \{\succeq L_j \}} p(L_i \mid \tb_\G(L_i)) \times  \prod_{M_i \in \mathbb{M} \cap \{\succeq L_j\}} p^*(M_i \mid \tb_\G(M_i)) \vert_{T=t}    \bigg)\ \bigg]
			\\
			&\overset{(1)}{=} \sum_{\preceq L_j}  \frac{\prod_{M_i \prec L_j} p^*(M_i \mid \tb_\G(M_i)) \vert_{T=t} }{\prod_{M_i \prec L_j} p^*(M_i \mid \tb_\G(M_i))} 
			\times \prod_{V_i \prec L_j} \ p(V_i \mid \tb_\G(V_i)) \times p(L_j \times \tb_\G(L_j))
			\\
			&\hspace{1cm} \times \bigg(\  \sum_{\succ L_j} Y \times \prod_{L_i \in \mathbb{L} \cap \{\succ L_j\}} p(L_i \mid \tb_\G(L_i)) \times  \prod_{M_i \in \mathbb{M} \cap \{\succ L_j\}} p^*(M_i \mid \tb_\G(M_i)) \vert_{T=t}  
			\\
			&\hspace{2cm} - \sum_{\succeq L_j} \ Y \times  \prod_{L_i \in \mathbb{L} \cap \{\succeq L_j \}} p(L_i \mid \tb_\G(L_i)) \times  \prod_{M_i \in \mathbb{M} \cap \{\succ L_j\}} p^*(M_i \mid \tb_\G(M_i)) \vert_{T=t}    \bigg)
			\\
			&\overset{(2)}{=} \sum_{\prec L_j}  \frac{\prod_{M_i \prec L_j} p^*(M_i \mid \tb_\G(M_i)) \vert_{T=t} }{\prod_{M_i \prec L_j} p^*(M_i \mid \tb_\G(M_i))} 
			\times \prod_{V_i \prec L_j} \ p(V_i \mid \tb_\G(V_i)) \times \sum_{L_j} \ p(L_j \times \tb_\G(L_j))
			\\
			&\hspace{1cm} \times \bigg(\  \sum_{\succ L_j} Y \times \prod_{L_i \in \mathbb{L} \cap \{\succ L_j\}} p(L_i \mid \tb_\G(L_i)) \times  \prod_{M_i \in \mathbb{M} \cap \{\succ L_j\}} p^*(M_i \mid \tb_\G(M_i)) \vert_{T=t} 
			\\
			&\hspace{2cm} - \sum_{\succeq L_j} \ Y \times  \prod_{L_i \in \mathbb{L} \cap \{\succeq L_j\}} p(L_i \mid \tb_\G(L_i)) \times  \prod_{M_i \in \mathbb{M} \cap \{\succ L_j\}} p^*(M_i \mid \tb_\G(M_i)) \vert_{T=t}    \bigg) 
			\\
			&\overset{(3)}{=} \sum_{\prec L_i}  \frac{\prod_{M_i \prec L_j} p^*(M_i \mid \tb_\G(M_i)) \vert_{T=t} }{\prod_{M_i \prec L_j} p^*(M_i \mid \tb_\G(M_i))} 
			\times \prod_{V_i \prec L_j} \ p(V_i \mid \tb_\G(V_i)) 
			\\
			&\hspace{1cm} \times \bigg(\ \sum_{\succeq L_j} Y \times \prod_{L_i \in \mathbb{L} \cap \{\succeq L_j \}} p(L_i \mid \tb_\G(L_i)) \times  \prod_{M_i \in \mathbb{M} \cap \{\succ L_j\}} p^*(M_i \mid \tb_\G(M_i)) \vert_{T=t} 
			\\
			&\hspace{2cm} - \sum_{\succeq L_j} \ Y \times  \prod_{L_i \in \mathbb{L} \cap \{\succeq L_j \}} p(L_i \mid \tb_\G(L_i)) \times  \prod_{M_i \in \mathbb{M} \cap \{\succ L_j\}} p^*(M_i \mid \tb_\G(M_i)) \vert_{T=t}    \bigg) 
			\\
			&\overset{(4)}{=} 0. 
		\end{align*}
	}
	The first equality is by definition of joint expectation.
	The second equality is by the fact that terms associated with $\prec L_j$ are not functions of $L_j$. The third equality is by term grouping.
	
	\vspace{0.25cm}
	\noindent Moreover, for any $M_j, M_{j - 1} \in \mathbb{M},$ the following equality holds, 
	
	{\scriptsize
		\begin{align*}
			&\E\Bigg[  
			\frac{\I(T=t)}{\prod_{L_i \prec M_j} p(L_i \mid \tb_\G(L_i))}  
			\times \sum_{ \substack{T\cup \{\succeq M_j\}}} \ Y 
			\times 
			\prod_{L_i \in \mathbb{L} } p(L_i \mid \tb_\G(L_i)) 
			\times 
			\prod_{M_i \in \mathbb{M} \cap \{\succeq M_j\}} p^*(M_i \mid \tb_\G(M_i)) \ \vert_{T=t} 
			\Bigg] 
			\\
			&=\E\Bigg[
			\frac{\I(T=t)}{\prod_{L_i \prec M_{j-1}} p(L_i \mid \tb_\G(L_i))} \times \sum_{ \substack{T \cup \{\succ M_{j - 1}\}}} \ Y \times 
			\prod_{ L_i \in \ \mathbb{L} } p(L_i \mid \tb_\G(L_i)) \times 
			\prod_{M_i \in \mathbb{M} \cap \{\succ M_{j-1}\}} p^*(M_i \mid \tb_\G(M_i)) \ \vert_{T=t} 
			\Bigg],
		\end{align*}
	}%
	since the left hand side is equal to
	{\scriptsize
		\begin{align*}
			&\sum_{\prec M_i} p(\prec M_i) \times 
			\frac{\I(T=t)}{\prod_{L_i \prec M_j} p(L_i \mid \tb_\G(L_i))}  
			\\ 
			&\hspace{4cm} \times \bigg[ \sum_{ \substack{T\cup \{\succeq M_j\}}} \ 
			Y \times 
			\prod_{ L_i \in \ \mathbb{L} } p(L_i \mid \tb_\G(L_i)) 
			\times 
			\prod_{M_i \in \mathbb{M} \cap \{\succeq M_j\}} p^*(M_i \mid \tb_\G(M_i)) \ \vert_{T=t} \bigg] 
			\\
			&\overset{(1)}{=} \sum_{\preceq M_{j-1}} \ 
			p(\preceq M_{i-1}) \times \Bigg\{ 
			\sum_{M_{j-1} \prec L_k \prec M_j } \ 
			\frac{\I(T=t)}{\prod_{L_i \prec M_j} p(L_i \mid \tb_\G(L_i))}  
			\times 
			p(L_k \mid \tb_\G(L_k)) \\ 
			&\hspace{4cm} \times
			\bigg[ \sum_{ \substack{T\cup \{\succeq M_j\}}} \ 
			Y \times 
			\prod_{ L_i \in \ \mathbb{L} } p(L_i \mid \tb_\G(L_i)) 
			\times 
			\prod_{M_i \in \mathbb{M} \cap \{\succeq M_j\}} p^*(M_i \mid \tb_\G(M_i)) \ \vert_{T=t} \bigg] \Bigg\}
			\\
			&\overset{(2)}{=}  \sum_{\preceq M_{j-1}} \ 
			p(\preceq M_{j-1}) \times 
			\frac{\I(T=t)}{\prod_{L_i \prec M_{j-1}} p(L_i \mid \tb_\G(L_i))}  
			\\ 
			&\hspace{2.5cm} \times
			\sum_{M_{j-1} \prec L_k \prec M_j } \ 
			\Bigg\{ \sum_{ \substack{T\cup \{\succeq M_j\}}} \ 
			Y \times 
			\prod_{ L_i \in \ \mathbb{L} } p(L_i \mid \tb_\G(L_i)) 
			\times 
			\prod_{M_i  \in \mathbb{M} \cap \{\succeq M_j\}} p^*(M_i \mid \tb_\G(M_i)) \ \vert_{T=t}  \Bigg\}
			\\
			&\overset{(3)}{=}  \sum_{\preceq M_{j-1}} \ 
			p(\preceq M_{j-1}) \times  
			\frac{\I(T=t)}{\prod_{L_i \prec M_{j-1}} p(L_i \mid \tb_\G(L_i))}  
			\\ 
			&\hspace{2.5cm} \times
			\Bigg\{ \sum_{ \substack{T\cup \{\succ M_{j-1}\}}} \ 
			Y \times 
			\prod_{ L_i \in \ \mathbb{L} } p(L_i \mid \tb_\G(L_i)) 
			\times 
			\prod_{M_i \in \mathbb{M} \cap \{\succeq M_j\}} p^*(M_i \mid \tb_\G(M_i)) \ \vert_{T=t} \Bigg\}
			\\
			&\overset{(4)}{=} \E\Bigg[
			\frac{\I(T=t)}{\prod_{L_i \prec M_{j-1}} p(L_i \mid \tb_\G(L_i))} \ 
			\\
			&\hspace{3cm}  \times  \bigg\{ \sum_{ \substack{T \cup \{\succ M_{j - 1}\}}} \ Y 
			\times 
			\prod_{ L_i \in \ \mathbb{L} } p(L_i \mid \tb_\G(L_i)) 
			\times 
			\prod_{M_i \in \{\mathbb{M} \cap \succ M_{j-1}\}} p^*(M_i \mid \tb_\G(M_i)) \ \vert_{T=t} 
			\bigg\}\Bigg],
		\end{align*} 
	}%
	which is exactly the same as the right hand side. This leaves the IF with only two terms $\psi(t)$ and $\bprimal$ and according to Lemma~\ref{lem:primal_ipw}, $\E[\bprimal] = \psi(t)$,
	provided the models in $\mathbb{L}$ are correctly specified, which was assumed.  Therefore, $\E[U_{\psi_t}] = 0.$

	\vspace{0.5cm}	
	\noindent \textbf{Scenario 2.} Assume models in $\mathbb{M}$ are correctly specified, and let $p^*(L_i \mid \tb_\G(L_i))$ denote the misspecified model for $p(L_i \mid \tb_\G(L_i)), \forall L_i \in \mathbb{L}.$ 
	We note that for any $M_j \in \mathbb{M},$ the following line in the IF evaluates to zero. 
	
	{\scriptsize
		\begin{align*}
			&\E\bigg[\ \frac{\I(T=t)}{\prod_{L_i \prec M_j} p^*(L_i \mid \tb_\G(L_i))} \bigg(\  \sum_{ \substack{T \cup \{\succ M_j\}}} \ 
			Y \times \prod_{ L_i \in \ \mathbb{L} } \ p^*(L_i \mid \tb_\G(L_i)) 
			\times  \prod_{ M_i \in \mathbb{M} \cap \{\succ M_j\} } \ p(M_i \mid \tb_\G(M_i))\vert_{T=t}\ \nonumber \\
			&\hspace{4cm} - \ \sum_{ \substack{T\cup \{\succeq M_j\}}} \ Y \times  \prod_{L_i \in \ \mathbb{L} } p^*(L_i \mid \tb_\G(L_i)) 
			\times 
			\prod_{M_i \in \mathbb{M}  \cap \{\succeq M_j\}} p(M_i \mid \tb_\G(M_i))\vert_{T=t} \bigg) \bigg]
			\\
			&\overset{(1)}{=} \sum_{ \preceq M_j} \ 
			\frac{\I(T=t)}{\prod_{L_i \prec M_j} p^*(L_i \mid \tb_\G(L_i))}  \times \prod_{V_i \prec M_j} \ p(V_i \mid \tb_\G(V_i)) \times p(M_j \mid \tb_\G(M_j)) \\
			&\hspace{2cm} \times  \bigg(\  \sum_{ \substack{T \cup \{\succ M_j\}}} \ 
			Y \times \prod_{  L_i \in \ \mathbb{L} } \ p^*(L_i \mid \tb_\G(L_i)) 
			\times  \prod_{ M_i \in \mathbb{M} \cap \{\succ M_j\} } \ p(M_i \mid \tb_\G(M_i))\vert_{T=t}\ \nonumber \\
			&\hspace{2.5cm} - \ \sum_{ \substack{T\cup \{\succeq M_j\}}} \ Y \times  \prod_{L_i \in \ \mathbb{L} } p^*(L_i \mid \tb_\G(L_i)) 
			\times 
			\prod_{M_i \in \mathbb{M}  \cap \{\succeq M_j\}} p(M_i \mid \tb_\G(M_i))\vert_{T=t} \bigg)
			\\
			&\overset{(2)}{=} \sum_{\prec M_j} \ 
			\frac{\I(T=t)}{\prod_{L_i \prec M_j} p^*(L_i \mid \tb_\G(L_i))}  \times \prod_{V_i \prec M_j} \ p(V_i \mid \tb_\G(V_i)) \times  \sum_{M_j} \ p(M_j \mid \tb_\G(M_j))   \\
			&\hspace{1cm} \times  \bigg(\sum_{ \substack{T \cup \{\succ M_j\}}} \ 
			Y \times \prod_{  L_i \in \ \mathbb{L} } \ p^*(L_i \mid \tb_\G(L_i)) 
			\times  \prod_{ M_i \in \mathbb{M} \cap \{\succ M_j \} } \ p(M_i \mid \tb_\G(M_i))\vert_{T=t} \\
			&\hspace{1.5cm} - \ \sum_{ \substack{T\cup \{\succeq M_j\}}} \ Y \times  \prod_{L_i \in \ \mathbb{L} } p^*(L_i \mid \tb_\G(L_i)) 
			\times 
			\prod_{M_i \in \mathbb{M}  \cap \{\succeq M_j\}} p(M_i \mid \tb_\G(M_i))\vert_{T=t} \bigg)
			\\
			&\overset{(3)}{=} \sum_{\prec M_j} \ 
			\frac{\I(T=t)}{\prod_{L_i \prec M_j} p^*(L_i \mid \tb_\G(L_i))}  \times \prod_{V_i \prec M_j} \ p(V_i \mid \tb_\G(V_i))  \\
			&\hspace{2cm} \times  \bigg(\ \sum_{ \substack{T \cup \{\succeq M_j\}}} \ 
			Y \times \prod_{  L_i \in \ \mathbb{L} } \ p^*(L_i \mid \tb_\G(L_i)) 
			\times  \prod_{ M_i \in \mathbb{M} \cap \{\succeq M_j\} } \ p(M_i \mid \tb_\G(M_i))\vert_{T=t} \\
			&\hspace{2.5cm} - \ \sum_{ \substack{T\cup \{\succeq M_j\}}} \ Y \times  \prod_{L_i \in \ \mathbb{L} } p^*(L_i \mid \tb_\G(L_i)) 
			\times 
			\prod_{M_i \in \mathbb{M}  \cap \{\succeq M_j\}} p(M_i \mid \tb_\G(M_i))\vert_{T=t} \bigg)
			\\
			&\overset{(4)}{=}0.
		\end{align*} 
	}
	
	\noindent Moreover, for any $L_j, L_{j-1} \in \mathbb{L},$ the following equality holds,
	
	{\scriptsize
		\begin{align*}
			&\E\bigg[ \
			\frac{\prod_{M_i \prec L_j} p(M_i \mid \tb_\G(M_i)) \vert_{T=t} }{\prod_{M_i \prec L_j} p(M_i \mid \tb_\G(M_i))} \\
			&\hspace{2cm}  \times \sum_{\succeq L_j} \ 
			Y \times  
			\prod_{L_i \in \mathbb{L} \cap \{\succeq L_j\}} p^*(L_i \mid \tb_\G(L_i))
			\times  
			\prod_{M_i \in \mathbb{M} \cap \{\succeq L_j\}} p(M_i \mid \tb_\G(M_i))\ \vert_{T=t} 
			\bigg]
			\\
			&\E\bigg[ \
			\frac{\prod_{M_i \prec L_{j-1}} p(M_i \mid \tb_\G(M_i)) \vert_{T=t} }{\prod_{M_i \prec L_{j-1}} p(M_i \mid \tb_\G(M_i))} \\
			&\hspace{2cm}  \times \sum_{\succ L_{j-1}} \ 
			Y \times  
			\prod_{L_i \in \mathbb{L} \cap \{\succ L_{j-1}\}} p^*(L_i \mid \tb_\G(L_i))
			\times  
			\prod_{M_i \in \mathbb{M} \cap \{\succ L_{j-1}\}} p(M_i \mid \tb_\G(M_i))\ \vert_{T=t} 
			\bigg],
		\end{align*}
	}%
	since the left hand side is equal to
	
	{\scriptsize
		\begin{align*}
			&\sum_{\prec L_j} p(\prec L_j) \times 
			\frac{\prod_{M_i \prec L_j} p(M_i \mid \tb_\G(M_i)) \vert_{T=t} }{\prod_{M_i \prec L_j} p(M_i \mid \tb_\G(M_i))} \\
			&\hspace{3.5cm}
			\times  \sum_{\succeq L_j} \ 
			Y \times  
			\prod_{L_i \in \mathbb{L} \cap \{\succeq L_j\}} p^*(L_i \mid \tb_\G(L_i))
			\times  
			\prod_{M_i \in \mathbb{M} \cap \{\succeq L_j\}} p(M_i \mid \tb_\G(M_i))\ \vert_{T=t} 
			\\
			&\overset{(1)}{=}  \sum_{\preceq L_{j-1}} p(\preceq L_{j-1}) \times \Bigg\{\sum_{L_{j-1} \prec M_k \prec L_j} \frac{\prod_{M_i \prec L_j} p(M_i \mid \tb_\G(M_i)) \vert_{T=t} }{\prod_{M_i \prec L_j} p(M_i \mid \tb_\G(M_i))} 
			\times p(M_k \mid \tb_\G(M_k))     
			\\
			&\hspace{3.5cm} 
			\times  \sum_{\succeq L_j} \ 
			Y \times  
			\prod_{L_i \in \mathbb{L} \cap \{\succeq L_j\}} p^*(L_i \mid \tb_\G(L_i))
			\times  
			\prod_{M_i \in \mathbb{M} \cap \{\succeq L_j\}} p(M_i \mid \tb_\G(M_i))\ \vert_{T=t} 
			\Bigg\}
			\\
			&\overset{(2)}{=}  \sum_{\preceq L_{j-1}} p(\preceq L_{j-1}) \times \frac{\prod_{M_i \prec L_{j-1}} p(M_i \mid \tb_\G(M_i)) \vert_{T=t} }{\prod_{M_i \prec L_{j-1}} p(M_i \mid \tb_\G(M_i))} \times 
			\Bigg\{
			\sum_{L_{j-1} \prec M_k \prec L_j} p(M_k \mid \tb_\G(M_k))\vert_{T=t} 
			\\
			&\hspace{2cm} \times 
			\sum_{\succeq L_j} \ 
			Y \times  
			\prod_{L_i \in \mathbb{L} \cap \{\succeq L_j\}} p^*(L_i \mid \tb_\G(L_i))
			\times  
			\prod_{M_i \in \mathbb{M} \cap \{\succeq L_j\}} p(M_i \mid \tb_\G(M_i))\ \vert_{T=t} 
			\Bigg\}
			\Bigg\}
			\\
			&\overset{(3)}{=}  \sum_{\preceq L_{j-1}} p(\preceq L_{j-1}) \times \frac{\prod_{M_i \prec L_{j-1}} p(M_i \mid \tb_\G(M_i)) \vert_{T=t} }{\prod_{M_i \prec L_{j-1}} p(M_i \mid \tb_\G(M_i))} \times 
			\\
			&\hspace{1.5cm} \times  
			\Bigg\{ \sum_{L_{j-1} \prec M_k \prec L_j} 
			\bigg[  \sum_{\succeq L_j} \ 
			Y \times  
			\prod_{L_i \in \mathbb{L} \cap \{\succeq L_j\}} p^*(L_i \mid \tb_\G(L_i))
			\times  
			\prod_{M_i \in \mathbb{M} \cap \{\succeq L_j\}} p(M_i \mid \tb_\G(M_i))\ \vert_{T=t} 
			\bigg] \\
			&\hspace{8.2cm} \times p(M_k \mid \tb_\G(M_k))\vert_{T=t} 
			\Bigg\}
			\\
			&\overset{(4)}{=}  \sum_{\preceq L_{j-1}} p(\preceq L_{j-1}) \times \frac{\prod_{M_i \prec L_{j-1}} p(M_i \mid \tb_\G(M_i)) \vert_{T=t} }{\prod_{M_i \prec L_{j-1}} p(M_i \mid \tb_\G(M_i))} 
			\\
			&\hspace{2.5cm} \times
			\sum_{\succ L_{j-1}}  Y \times  
			\prod_{L_i \in \mathbb{L} \cap \{\succ L_{j-1}\}} p^*(L_i \mid \tb_\G(L_i))
			\times  
			\prod_{M_i \in \mathbb{M} \cap \{\succ L_{j-1}\}} p(M_i \mid \tb_\G(M_i))\ \vert_{T=t} 
			\Bigg\}
			\\
			&\overset{(5)}{=} \E\Bigg[ \
			\frac{\prod_{M_i \prec L_{j-1}} p(M_i \mid \tb_\G(M_i)) \vert_{T=t} }{\prod_{M_i \prec L_{j-1}} p(M_i \mid \tb_\G(M_i))} 
			\\
			&\hspace{3cm} \times
			\sum_{\succ L_{j-1}}  Y \times  
			\prod_{L_i \in \mathbb{L} \cap \{\succ L_{j-1}\}} p^*(L_i \mid \tb_\G(L_i))
			\times  
			\prod_{M_i \in \mathbb{M} \cap \{\succ L_{j-1}\}} p(M_i \mid \tb_\G(M_i))\ \vert_{T=t} 
			\Bigg],
		\end{align*}
	}%
	which is exactly the same as the right hand side. This leaves the IF with only two terms $\psi(t)$ and $\bdual$ and according to Lemma~\ref{lem:dual_ipw}, $\E[\bdual] = \psi(t).$ Therefore, $\E[U_{\psi_t}] = 0.$  
\end{proof}

%%%%%%%%%%%%%%%%%%%%%%%%%%%%%%%%%%%%%%%%%%%%%

\subsection*{Lemma~\ref{thm:reform_IF} (Reformulation of the IF for augmented primal IPW) }

\begin{proof}
	We prove this lemma by showing what happens to $V_i \in V,$ if $V_i$ is in $\mathbb{M},$ or $\mathbb{L},$ or $\mathbb{C}.$
	
	\vspace{0.5cm}
	\noindent $\circ$ For any $M_j \in \mathbb{M},$ we have,
	{\scriptsize
		\begin{align*}
			\E\Big[\bprimal \ \Big| \ \{\preceq M_j\} \Big] 
			&= \E\bigg[ 
			\frac{\I(T=t)}{\prod_{L_i \in \mathbb{L}} p(L_i \mid \tb_\G(L_i))} \times  \sum_{T} \ \prod_{L_i \in \mathbb{L}} \ p(L_i \mid \tb_\G(L_i)) \times Y  \ \bigg| \  \{\preceq M_i\} \bigg]
			\\
			&= \sum_{V_i \succ M_j} \  \prod_{V_i \succ M_j} \  p(V_i \mid \tb_\G(V_i))  \times \frac{\I(T=t)}{\prod_{L_i \in \mathbb{L}} p(L_i \mid \tb_\G(L_i))}  \times  \sum_{T} \ \prod_{L_i \in \mathbb{L}} \ p(L_i \mid \tb_\G(L_i)) \times Y 
			\\
			&= \frac{\I(T=t)}{\prod_{L_i \prec M_j} p(L_i \mid \tb_\G(L_i))}    
			\times \sum_{V_i \succ M_j} \  \sum_{T} \ \prod_{V_i \in \mathbb{L} \cup \{\succ M_j\}} \ p(V_i \mid \tb_\G(V_i))\vert_{T=t\text{ if } V_i \in \mathbb{M}}  \times Y \bigg\}
			\\
			&= \frac{\I(T=t)}{\prod_{L_i \prec M_j} p(L_i \mid \tb_\G(L_i))}    
			\times \sum_{T \cup \{\succ M_j\} } \ Y \times \prod_{V_i \in \mathbb{L} \cup \{\succ M_j\}} \ p(V_i \mid \tb_\G(V_i))\vert_{T=t\text{ if } V_i \in \mathbb{M}}  \bigg\}.
		\end{align*}
	}%
	Similarly, 
	{\scriptsize
		\begin{align*}
			\E\big[\bprimal \mid \{\prec M_j\} \big] 
			= \frac{\I(T=t)}{\prod_{L_i \prec M_j} p(L_i \mid \tb_\G(L_i))}    
			\times \sum_{T \cup \{\succeq M_j\} } \ Y \times \prod_{V_i \in \mathbb{L} \cup \{\succeq M_j\}} \ p(V_i \mid \tb_\G(V_i))\vert_{T=t\text{ if } V_i \in \mathbb{M}}  \bigg\}.
		\end{align*}
	}%
	Therefore, {\small $\E\big[\bprimal \mid \{\preceq M_j\} \big]  - \E\big[\bprimal \mid \prec \{M_j\} \big]$} is equivalent to $M_j$'s corresponding line in the IF. 
	
	\vspace{0.5cm}
	\noindent $\circ$ Now, for any $L_j \in \mathbb{L},$ we have, 
	{\scriptsize
		\begin{align*}
			\E\Big[\bdual \ \Big| \ \{\preceq L_j\} \Big] 
			&= \E\bigg[ 
			\frac{\prod_{M_i \in \mathbb{M}} \ p(M_i \mid \tb_\G(M_i))\vert_{T=t} }{\prod_{M_i \in \mathbb{M}} \ p(M_i \mid \tb_\G(M_i)) } \times Y \ \bigg| \ \{\preceq L_j\} 
			\Bigg]
			\\
			&= \sum_{V_i \succ L_j} \ \prod_{V_i \succ L_j} \ p(V_i \mid \tb_\G(V_i)) \times 	\frac{\prod_{M_i \in  \mathbb{M}} \ p(M_i \mid \tb_\G(M_i))\vert_{T=t} }{\prod_{M_i \in \mathbb{M}} \ p(M_i \mid \tb_\G(M_i)) } \times Y \
			\\
			&= 	\frac{\prod_{M_i  \prec L_j} \ p(M_i \mid \tb_\G(M_i))\vert_{T=t} }{\prod_{M_i \prec L_j} \ p(M_i \mid \tb_\G(M_i)) } \times \sum_{V_i \succ L_j} \ 
			Y \times  \prod_{V_i \succ L_j} \ p(V_i \mid \tb_\G(V_i))\vert_{T=t\text{ if } V_i \in \mathbb{M}}. 
		\end{align*}
	}%
	Similarly, 
	{\scriptsize
		\begin{align*}
			\E\Big[\bdual \ \Big| \ \{\prec L_j\} \big] =  \frac{\prod_{M_i  \prec L_j} \ p(M_i \mid \tb_\G(M_i))\vert_{T=t} }{\prod_{M_i \prec L_j} \ p(M_i \mid \tb_\G(M_i)) } \times \sum_{V_i \succeq L_j} \ 
			Y \times  \prod_{V_i \succeq L_j } \ p(V_i \mid \tb_\G(V_i))\vert_{T=t\text{ if } V_i \in \mathbb{M}} .
		\end{align*}
	}%
	Therefore, {\small $\E\big[\bdual \mid \{\preceq L_j\} \big]  - \E\big[\bdual\mid \{ \prec L_j\} \big]$} is equivalent to $L_j$'s corresponding line in the IF. 
	
	\vspace{0.5cm}
	\noindent $\circ$ For variables in $\mathbb{C},$ we have,
	
	{\scriptsize
		\begin{align*}
			\E[\bprimal \mid \mathbb{C}] 
			&= \E\bigg[ 
			\frac{\I(T=t)}{\prod_{L_i \in \mathbb{L}} p(L_i \mid \tb_\G(L_i))} \times  \sum_{T} \ \prod_{L_i \in \mathbb{L}} \ p(L_i \mid \tb_\G(L_i)) \times Y  \ \bigg| \  \mathbb{C} \bigg]
			\\
			&= \sum_{V \setminus \mathbb{C}} \ p(V \setminus \mathbb{C}) \times \frac{\I(T=t)}{\prod_{L_i \in \mathbb{L}} p(L_i \mid \tb_\G(L_i))} \times  \sum_{T} \ \prod_{L_i \in \mathbb{L}} \ p(L_i \mid \tb_\G(L_i)) \times Y 
			\\
			&= \sum_{V \setminus \mathbb{C}}\ \mathbb{I}(T=t) \times \prod_{M_i \in \mathbb{M}} \ p(M_i \mid \tb_\G(M_i)) \times  \sum_{T} \ \prod_{L_i \in \mathbb{L}} \ p(L_i \mid \tb_\G(L_i)) \times Y 
			\\
			&= \sum_{V \setminus \{T, \mathbb{C}\}} \ \prod_{M_i \in \mathbb{M}} \ p(M_i \mid \tb_\G(M_i))\Big\vert_{T =t} \times  \sum_{T} \ \prod_{L_i \in \mathbb{L}} \ p(L_i \mid \tb_\G(L_i)) \times Y,
		\end{align*}
	}%
	and based on Lemma~\ref{lem:primal_ipw}, ${\small \E[\bprimal] = \psi(t)}.$ Therefore, $\E[\bprimal \mid \mathbb{C}] - \E[\bprimal]$ corresponds to the last line in the IF. We can also run a similar argument for $\bdual$. According to Lemma~\ref{lem:dual_ipw}, ${\small \E[\bdual] = \psi(t)},$ and 
	
	{\small
		\begin{align*}
			\E[\bdual \mid \mathbb{C}] 
			&= \E\bigg[ \frac{\prod_{M_i \in \mathbb{M}} \ p(M_i \mid \tb_\G(M_i))\vert_{T=t} }{\prod_{M_i \in \mathbb{M}} \ p(M_i \mid \tb_\G(M_i)) } \times Y \ \bigg| \ \mathbb{C} \bigg]
			\\
			&= \sum_{V \setminus  \mathbb{C}} \ p(V \setminus \mathbb{C}) \times \frac{\prod_{M_i \in \mathbb{M}} \ p(M_i \mid \tb_\G(M_i))\vert_{T=t} }{\prod_{M_i \in \mathbb{M}} \ p(M_i \mid \tb_\G(M_i)) } \times Y
			\\
			&= \sum_{V \setminus  \mathbb{C}} \ \prod_{M_i \in \mathbb{M}} p(M_i \mid \tb_\G(M_i))\vert_{T=t} \times \prod_{L_i \in \mathbb{L}} \ p(L_i \mid \tb_\G(L_i)) \times Y
			\\
			&= \sum_{V \setminus \{T, \mathbb{C}\}} \ \prod_{M_i \in \mathbb{M}} \ p(M_i \mid \tb_\G(M_i))\Big\vert_{T =t} \times  \sum_{T} \ \prod_{L_i \in \mathbb{L}} \ p(L_i \mid \tb_\G(L_i)) \times Y
			\\
			&= \E[\bprimal \mid \mathbb{C}].
		\end{align*}
	}
\end{proof}

%%%%%%%%%%%%%%%%%%%%%%%%%%%%%%%%%%%%%%%%%%%%%

\subsection*{Lemma~\ref{thm:a-fix_ID} (Identifying functional when $T$ is fixable)}
\begin{proof}
	A set $Z$ satisfies the backdoor criterion (and so yields identification of the targe via the adjustment functional) with respect to a treatment $T$ and outcome $Y$ if (i) $Z$ does not contain any descendants of $T$; and (ii) $T$ and $Y$ are m-separated by $Z$  in a graph $\G_{\bar{T}}$ where all outgoing edges $T \diedgeright \circ$ from the treatment are removed \citep{pearl2009causality}. We now show that $\tb_\G(T)$ satisfies both criteria.
	By the pre-condition that $\dis_\G(T) \cap \de_\G(T) = \{T\},$ there exists a valid topological ordering on vertices $V$ such that $T$ appears last among the members of its district. Under such an ordering $\tb_\G(T) = \mb_\G(T).$ That is, under fixability $\tb_\G(T) = \dis_\G(T) \cup \pa_{\G}(\dis_\G(T)) \setminus T.$  To see (i) is satisfied we note that $\tb_\G(T)$ does not contain any descendants of $T.$ To see (ii) is satisfied, we first note that $Y$ is a non-descendant of $T$ in $\G_{\bar{T}},$ and the Markov blanket of $T$ in this graph remains the same and excludes the outcome $Y$ when $T$ is fixable. Further, from \cite{richardson2017nested} we know that the Markov blanket of a variable m-separates the variable from all its non-descendants (excluding the Markov blanket itself.) From this it follows that $T \ci Y \mid \tb_{\G}(T)$ in $\G_{\bar{T}}.$
\end{proof}

%%%%%%%%%%%%%%%%%%%%%%%%%%%%%%%%%%%%%%%%%%%%%

\subsection*{Theorem~\ref{thm:eff-APIPW} (Efficient augmented primal IPW in mb-shielded ADMGs) }

\begin{proof}
	Consider the reformulated IF in Lemma~\ref{thm:reform_IF}. In order to get the efficient IF, we project the reformulated IF onto the tangent space $\Lambda^*$ given by Lemma~\ref{lem:Lambda}. We first note that we can rewrite the term $\sum_{\mathbb{C}} \E[\beta_{\text{primal/dual}} \mid \mathbb{C}] - \psi(t)$ in the reformulated IF as ${\small \sum_{C_i \in \mathbb{C}} \E[\beta_{\text{primal/dual}} \mid \{\preceq C_i\}] -  \E[\beta_{\text{primal/dual}} \mid \{\prec C_i\}],}$ where $\beta_{\text{primal/dual}}$ means that we can use either $\bprimal$ or $\bdual$ for the $\mathbb{C}$ term. We have,  	
	
	{\small
		\begin{align*}
			\pi[U^{\text{reform}}_{\psi_t} \mid \Lambda^*] 
			&= 
			\sum_{M_i \in \mathbb{M}} \pi\Big[
			\E[\bprimal \mid \{\preceq M_i\}] - \E[\bprimal \mid \{\prec M_i\}] \Big| \ \Lambda^*\Big] \\
			&\hspace{0.25cm} + \sum_{L_i \in \mathbb{L}}  \pi\Big[
			\E[\bdual \mid \{\preceq L_i\}] - \E[\bdual \mid \{\prec L_i\}] \ \Big| \ \Lambda^*
			\Big] \\
			&\hspace{0.25cm} + \sum_{C_i \in \mathbb{C}}  \pi\Big[
			\E[\beta_{\text{primal/dual}} \mid \preceq C_i] - \E[\beta_{\text{primal/dual}} \mid \prec C_i] \ \Big| \ \Lambda^*
			\Big].
		\end{align*}
	}%
	Let $\beta$ be either $\bprimal$ or $\bdual$ or $\beta_{\text{primal/dual}}.$ 
	Note that $\Big\{\E\big[\beta \mid \{\preceq V_i\}\big] - \E\big[\bprimal \mid \{\prec V_i\}\big]\Big\}$ lives in $\Lambda_{V_i},$ and $\Lambda_{V_i} \ci \Lambda^*\setminus \Lambda^*_{V_i}.$ Therefore, their projection onto $\Lambda^*\setminus \Lambda^*_{V_i}$ is zero. We have, 
	
	{\small
		\begin{align*}
			\pi\Big[
			&\E\big[\beta \mid \{\preceq V_i\}\big] - \E\big[\beta \mid \{\prec V_i\}\big] \Big| \ \Lambda^*_{V_i} \Big] \\
			&\hspace{0.25cm} = \E\Big[ \E[\beta \mid \{\preceq V_i\}] - \E[\beta \mid \{\prec V_i\}] \ \Big| \ V_i, \tb_\G(V_i) \Big] - \E\Big[ \E[\beta \mid \{\preceq V_i\}] - \E[\beta \mid \{\prec V_i\}] \ \Big| \ \tb_\G(V_i) \Big] \\
			&\hspace{0.25cm} = \E\big[\beta \ \big| \ V_i, \tb_\G(V_i) \big] - \E\Big[ \E\big[\beta \ \big| \prec V_i \big] \ \Big| \ V_i, \tb_\G(V_i) \Big]  - \E\big[\beta \ \big| \ \tb_\G(V_i) \big] + \E\big[\beta \ \big| \ \tb_\G(V_i) \big] \\
			&\hspace{0.25cm} =  \E\big[\beta \ \big| \ V_i, \tb_\G(V_i) \big] - \E\Big[ \E\big[\beta \ \big| \prec V_i \big] \ \Big| \ V_i, \tb_\G(V_i) \Big] \\
			&\hspace{0.25cm} =  \E\big[\beta \ \big| \ V_i, \tb_\G(V_i) \big] -  \E\big[\beta \ \big| \  \tb_\G(V_i) \big]. 
		\end{align*}
	}%
	
	\vspace{0.3cm} 
	\noindent Therefore, the efficient IF is as follows. 
	{\small
		\begin{align*}
			\pi[U^{\text{reform}}_{\psi_t} \mid \Lambda^*] 
			&= \sum_{M_i \in \mathbb{M}} 
			\E\big[\bprimal \mid M_i, \tb_\G(M_i)\big] - \E\big[\bprimal \mid \tb_\G(M_i)\big] \\
			&\hspace{0.25cm} + \sum_{L_i \in \mathbb{L}} 
			\E\big[\bdual \mid L_i, \tb_\G(L_i)\big] - \E\big[\bdual \mid \tb_\G(L_i)\big] \\
			&\hspace{0.25cm}  + \sum_{C_i \in \mathbb{C}} 
			\E\big[\beta_{\text{primal/dual}} \mid C_i, \tb_\G(C_i)\big] - 	\E\big[\beta_{\text{primal/dual}} \mid \tb_\G(C_i)\big].
		\end{align*}
	}%
\end{proof}

%%%%%%%%%%%%%%%%%%%%%%%%%%%%%%%%%%%%%%%%%%%%%

\subsection*{Lemma~\ref{thm:a-fix_EIF} (Efficient augmented IPW in mb-shielded ADMGs) }

\begin{proof}
	An mb-shielded ADMG is Markov equivalent to a DAG $\G^d$, which can be constructed as follows. Under the topological order $\tau$ fixed on the original ADMG $\G, V_i \diedgeright V_j$ exists in $\G^d$ if $V_i$ and $V_j$ are adjacent in $\G$ and $V_i \prec_\tau V_j.$ $\G^d$ is a DAG because we only allow for directed edges and there is no directed cycle as we follow a valid topological order in $\G.$ Further, $\tb_\G(V_i) = \pa_{\G^d}(V_i), \forall V_i \in V.$ Therefore, the identifying functional for the target parameter is the same in both $\G$ and $\G^d,$ that is $\E[\E[Y \mid T = t, \tb_\G(T)] = \E[\E[Y \mid T = t, \pa_{\G^d}(T)]]$.
	
	We know for the instrumental variables in 
	\[
	Z = \{Z_i \in V \mid Z_i \ci Y \mid \tb_\G(Z_i) \text{ in } \G_{V\setminus T} \text { and } Z_i \not\ci T \mid \tb_\G(Z_i)\},
	\]
	there always exists a set $F \in V$ that d-separates  $Z_i \in Z$ from $Y$ given $F, T$ \citep{van2015efficiently, rotnitzky2019efficient}.
	Showing that the equation
	{\small
		\begin{align*}
			\E\Big[\frac{\mathbb{I}(T = t)}{p(T \mid \tb_\G(T))} \times Y  \ \Big| \ Z_i, \tb_\G(Z_i) \Big] 
			&= \E\Big[\frac{\mathbb{I}(T = t)}{p(T \mid \tb_\G(T))} \times Y  \ \Big| \tb_\G(Z_i) \Big]. 
		\end{align*}
	}%
	holds then simply follows from the argument outlined in Proposition 3 of \cite{rotnitzky2019efficient}. 
	
	\noindent Finally, given $\Lambda^*$ in Lemma~\ref{lem:Lambda}, the efficient IF is as follows. Let $V^* = V \setminus (T \cup Z \cup D),$ 
	{\small
		\begin{align*}
			U^{\text{eff}}_{\psi_t} \
			&=  \pi\Big[ \frac{\mathbb{I}(T = t)}{p(T \mid \tb_\G(T))} \times Y  \ \Big| \ \Lambda^*\setminus \Lambda^*_{T} \Big] \\
			&= \sum_{V_i \in V^*} \ \E\Big[  \frac{\mathbb{I}(T = t)}{p(T \mid \tb_\G(T))} \times Y \ \Big| \  V_i, \tb_\G(V_i)   \Big] -  \E\Big[  \frac{\mathbb{I}(T = t)}{p(T \mid \tb_\G(T))} \times Y \ \Big| \  \tb_\G(V_i)  \Big]. 
		\end{align*}
	}%	
\end{proof}

%%%%%%%%%%%%%%%%%%%%%%%%%%%%%%%%%%%%%%%%%%%%%

\subsection*{Theorem~\ref{thm:nested_ipw} (Soundness and completeness of Algorithm 2)}

\begin{proof}
	Soundness of the algorithm implies that when our algorithm succeeds, the subsequent identifying functional for $\psi(t)$ is correct. Completeness implies, that when the algorithm fails, the target parameter $\psi(t)$ is not identifiable within the model.
	
	\subsubsection*{Soundness}
	We first prove soundness of the algorithm. That is, when Algorithm~\ref{alg:simplify} does not fail, $\psi(t)$ is indeed equal to $\psi(t)_{\text{nested}}.$ The algorithm does not fail when all districts $D \in {\cal D}^*$ are intrinsic in $\G.$ Note that ${\cal D}^*$ is a subset of the districts in $\G_{Y^*}.$ However, by construction of ${\cal D}^*,$ the remaining districts in $\G_{Y^*}$ are those that do not have any overlap with $D_T.$ We now show that such districts are always intrinsic in $\G.$
	
	Consider a district $D \in {\cal D}(\G_{Y^*})$ such that $D \cap D_T = \emptyset.$ The district $D$ forms a subset of a larger district in $\G,$ say $D' \in {\cal D}(\G).$ Due to results in \citep{tian2002general}, we know that $D'$ is always intrinsic. If $D = D'$ then the result immediately follows. Otherwise, In the CADMG $\phi_{V\setminus D'}(\G),$ there exists at least one vertex $D_i$ in $D'$ not in $Y^*,$ that has no children. This is because all directed paths from $D_i$ to vertices in $Y^*$ must go through $T$ and since $T$ is not in $D',$ all incoming edges to $T$ have been deleted. The only other way $D_i$ may not be childless is if there existed a cycle in $\G,$ which is a contradiction. Thus, such a vertex $D_i$ is always fixable and furthermore, fixing it corresponds to the marginalization operation  $\sum_{D_i} q_{D'}(D' \mid \pa_{\G}(D'))$ \citep{richardson2017nested}. Once $D_i$ is fixed, another vertex $D_j$ that is in $D'$ but not in $Y^*$ becomes childless. Applying this argument inductively, we see that all $D_i \in D'$ such that $D_i \not\in Y^*$ are fixable through marginalization under a reverse topological order. Hence for districts $D$ in $\G_{Y^*}$ that do not overlap with $D_T,$  the set $D = D' \setminus \{D_i \in D' \mid D_i \not\in Y^*\}$ is always intrinsic. Thus, Algorithm~\ref{alg:simplify} succeeds when all districts in $\G_{Y^*}$ are intrinsic.
	
	\subsubsection*{Soundness}
	
	We now show that under this condition, $\psi(t)_{\text{nested}} \equiv \E_{p^\dagger}\big[\frac{\I(T=t)}{p(T \mid \tb_\G(T))} \times Y \big] = \psi(t)$.  By definition, we have
	{\small 
		\begin{align*}
			\psi(t)_{\text{nested}} = \sum_{V} p(V) \times \prod_{D^* \in {\cal D^*}} \frac{q_{D^*}(D^* \mid \pa_{\G}(D^*))}{\prod_{D_i^* \in D^*}p(D_i^* \mid \tb_{\G}(D_i^*))} \times \frac{\I(T=t)}{p(T \mid \tb_\G(T))} \times Y.
		\end{align*}
	}%
	The districts of $\G$ can be partitioned into three sets. ${\cal D}_T$ is the district in $\G$ that contains $T$ (with all elements in ${\cal D}^*$, if any, subsets of $D_T$).  ${\cal D}'$ is the set of districts in $\G$, excluding $D_T$, that overlap with $Y^*$. 
	${\cal D}^z$ is the set of districts in $\G$, excluding $D_T$, that do not overlap with $Y^*$. 
	The observed distribution $p(V)$ then district factorizes as,
	
	{\small 
		\begin{align*}
			p(V) = \prod_{D^z \in {\cal D}^z} q_{D^z}(D^z \mid \pa_\G(D^z)) \times \prod_{D' \in {\cal D}'} q_{D'}(D' \mid \pa_\G(D')) \times q_{D_T}(D_T | \pa_\G(D_T)).
			%\prod_{D'' \in {\cal D}''} q_{D''}(D'' \mid \pa_\G(D''))
		\end{align*}
	}%
	By results in \cite{tian2002general}, $q_{D_T}(D_T \mid \pa_\G(D_T))$ is identified as $\prod_{D_i \in D_T} p(D_i \mid \tb_{\G}(D_i))$ (for any topological ordering).  Since every element in ${\cal D}^*$ is a subset of $D_T$, and since vertices in $D_T \setminus \bigcup_{D^* \in {\cal D}^*}$ precede vertices $D_T \cap \bigcup_{D^* \in {\cal D}^*} = D_T \cap Y^*$ in the ordering, we have
	
	{\small 
		\begin{align*}
			\psi(t)_{\text{nested}} = \sum_V &\prod_{D^z \in {\cal D}^z} q_{D^z}(D^z \mid \pa_\G(D^z)) \times \prod_{D' \in {\cal D}'} q_{D'}(D' \mid \pa_\G(D')) \times \prod_{D^* \in {\cal D^*}} q_{D^*}(D^* \mid \pa_{\G}(D^*)) \\
			&\times %\prod_{D'' \in {\cal D}''} \sum_{D'' \cap Y^*} q_{D''\setminus Y^*}(D'' \mid \pa_\G(D''))
			\sum_{D_T \cap Y^*} q_{D_T}(D_T | \pa_\G(D_T)) \times \frac{\I(T=t)}{p(T \mid \tb_\G(T))} \times Y.
		\end{align*}
	}%
	Since $T$ is the last element in the ordering in $D_T \setminus Y^*$, we further have:
	{\small 
		\begin{align*}
			\psi(t)_{\text{nested}} = \sum_{Y^*} \sum_{V \setminus Y^*} &\prod_{D^z \in {\cal D}^z} q_{D^z}(D^z \mid \pa_\G(D^z)) \times \prod_{D' \in {\cal D}'} q_{D'}(D' \mid \pa_\G(D')) \times \prod_{D^* \in {\cal D^*}} q_{D^*}(D^* \mid \pa_{\G}(D^*)) \\
			&\times %\prod_{D'' \in {\cal D}''} \sum_{D'' \cap Y^*} q_{D''\setminus Y^*}(D'' \mid \pa_\G(D''))
			\sum_{(D_T \cap Y^*) \cup \{ T \}} q_{D_T}(D_T | \pa_\G(D_T)) \times \I(T=t) \times Y.
		\end{align*}
	}%	
	
	Consider applying marginalization of elements in $V \setminus Y^*$ to $\psi(t)_{\text{nested}}$ above in the reverse topological ordering on $V \setminus Y^*$.
	Districts in $\G$ partition $V$ and so, by definition of ${\cal D}^*, {\cal D}'$ and $D_T$, elements in ${\cal D}^z \cup \{ D' \setminus Y^* : D' \in {\cal D}' \} \cup \{ D_T \setminus (Y^* \cup \{ T \}) \}$ partition $V \setminus Y^*$.  This partition, and the fact that marginalizations are processed in reverse topological order, means that at every stage, the variable to be summed occurs in precisely one place in the expression.  This implies that the result of the overall summation of $V \setminus Y^*$ yields:
	
	{\small 
		\begin{align*}
			\psi(t)_{\text{nested}} = \sum_{Y^*} & \prod_{D' \in {\cal D}'} \sum_{D' \setminus Y^*} q_{D'}(D' \mid \pa_\G(D')) \times \prod_{D^* \in {\cal D^*}} q_{D^*}(D^* \mid \pa_{\G}(D^*)) \times \I(T=t) \times Y \\
		\end{align*}
	}%	
	By definition, $q_{D^*}(D^* \mid \pa_\G(D^*)) \equiv \phi_{V \setminus D^*}(p(V); \G(V))$.  Since every $D'$ in ${\cal D}'$ is a top level district in $\G$, there exists a valid fixing sequence on $V \setminus D'$.  Further, in the CADMG $\phi_{V \setminus D'}(\G(V))$, any element in $D' \setminus Y^*$ cannot be an ancestor of an element in $D' \cap Y^*$ (if a directed path not through $T$ existed from an element $V_i$ in $D'$ to an element in $D' \cap Y^*$, then $V_i$ must itself be in $D' \cap Y^*$, while a directed path from $V_i$ to $D' \cap Y^*$ through $T$ disappears in $\phi_{V \setminus D'}(\G(V))$ since $T$ is outside $D'$.  Consequently fixing elements $D' \setminus Y^*$ in reverse topological order in $\phi_{V \setminus D'}(\G(V))$ and $\phi_{V \setminus D'}(p(V), \G(V))$ is equivalent to marginalizing those variables.  As a result, for every $D' \in {\cal D}'$, $\sum_{D' \setminus Y^*} q_{D'}(D' \mid \pa_\G(D')) = \phi_{V \setminus (D' \cap Y^*)}(p(V); \G(V))$. 
	Our conclusion follows:
	{\small 
		\begin{align*}
			\psi(t)_{\text{nested}} = \sum_{Y^*}\prod_{D \in {\cal D}(\G_{Y^*})} \phi_{V\setminus D} (p(V); \G) \times Y \bigg\vert_{T = t} = \psi(t).
		\end{align*}
	}%
	
	\subsubsection*{Completeness}
	Follows trivially as we have shown the failure condition of Algorithm~\ref{alg:simplify} to be equivalent to the failure condition of the identification algorithm in \cite{richardson2017nested} which is known to be sound and complete.
\end{proof}

%%%%%%%%%%%%%%%%%%%%%%%%%%%%%%%%%%%%%%%%%%%%%

\subsection*{Lemma~\ref{lem:p-commute} (Commutativity of p-fixing)}

\begin{proof}
	Consider a valid p-fixing sequence $(S_1,\dots,S_p)$ for the set $S.$ That the kernel $\primal_{(S_1, \ldots, S_p)}(p(V);\G)$ evaluated at any $s_1, \ldots, s_p$ is equal to $p(V \setminus \{ S_1, \ldots, S_p\} \mid \text{do}(s_1, \ldots, s_p))$ follows by an inductive application of Theorem 3 in \cite{tian2002general}. That is, for any two valid p-fixing sequences $\sigma_S^1$ and $\sigma_S^2$ defined on $S$ we have that $\primal_{\sigma_S^1}(p(V);\G)= \primal_{\sigma_S^2}(p(V);\G) = p(V \setminus S \mid \doo(S=s)).$
\end{proof}

%%%%%%%%%%%%%%%%%%%%%%%%%%%%%%%%%%%%%%%%%%%%%

\subsection*{Corollary~\ref{lem:p-fix_sequence_id} (Identification via a sequence of p-fixing)}

\begin{proof}
	Given any valid p-fixing sequence $(Z_1,\dots,Z_p, T)$ the kernel $\primal_{(Z_1, \ldots, Z_p, T)}(p(V);\G)$ evaluated at any $z_1, \ldots, z_p, t$ is equal to $p(V \setminus \{ Z_1, \ldots, Z_p, T \} \mid \text{do}(z_1, \ldots, z_p, t)).$  Then
	$p(Y \mid \text{do}(z_1, \ldots, z_p, t)) = \sum_{V \setminus \{ Z_1, \ldots, Z_p, T, Y \}} p(V \setminus \{ Z_1, \ldots, Z_p, T \} \mid \text{do}(z_1, \ldots, z_p, t))$.
	Since all vertices $Z_1, \ldots, Z_p$ have no directed paths to $Y$ except through $T$ in the original graph, the corresponding exclusion restrictions in the causal model imply 
	$p(Y \mid \text{do}(z_1, \ldots, z_p, t)) = p(Y \mid \text{do}(t))$.
\end{proof}

\end{appendices}

%ilya: commented this out to get rid of a blank page after supplement.
%\newpage
\bibliography{references}

\end{document}